\pdfoutput=1

\documentclass[11pt]{article}

\usepackage[preprint]{acl}

\usepackage{times}
\usepackage{latexsym}

\usepackage{tempora}

\usepackage[T2A,T1]{fontenc}
\usepackage[utf8]{inputenc}
\usepackage[russian,english]{babel}

\usepackage[utf8]{inputenc}

\usepackage{microtype}
\usepackage{booktabs}
\usepackage{amsmath}

\usepackage{inconsolata}

\usepackage{amsmath} 
\usepackage{amssymb}
\usepackage{array} 
\usepackage{multirow}
\usepackage{microtype}
\usepackage{inconsolata}

\usepackage{times}
\usepackage{latexsym}
\usepackage{adjustbox} 
\usepackage{graphicx}
\usepackage{listings}
\usepackage{booktabs}
\usepackage{hyperref}
\usepackage{subfigure}
\usepackage[most]{tcolorbox}

\usepackage{url}

\usepackage{xcolor}
\usepackage{enumitem}

\usepackage{colortbl}

\title{M-ABSA: A Multilingual Dataset for Aspect-Based Sentiment Analysis}

\author{
  \textbf{Chengyan Wu\textsuperscript{$\ast$1,2}}, 
\textbf{Bolei Ma\textsuperscript{$\ast$3,4}}, 
\textbf{Yihong Liu\textsuperscript{3,4}}, 
  \textbf{Zheyu Zhang\textsuperscript{5}}, 
  \textbf{Ningyuan Deng\textsuperscript{6}}, 
\\
\textbf{Yanshu Li\textsuperscript{7}}, 
\textbf{Baolan Chen\textsuperscript{8}}, 
\textbf{Yi Zhang\textsuperscript{3,9}}, 
\textbf{Yun Xue\textsuperscript{$\ddagger$1,2}}, 
\textbf{Barbara Plank\textsuperscript{3,4}} 
\vspace{4.5pt}
\\
{\small \textsuperscript{1}Guangdong Provincial Key Laboratory of Quantum Engineering and Quantum Materials} \\
{\small \textsuperscript{2}School of Electronic Science and Engineering (School of Microelectronics), 
South China Normal University} \\
{\small \textsuperscript{3}LMU Munich \textsuperscript{4}Munich Center for Machine Learning} 
{\small \textsuperscript{5}Technical University of Munich}\\
{\small \textsuperscript{6}Renmin University of China} 
{\small \textsuperscript{7}Brown University} 
{\small \textsuperscript{8}ShanghaiTech University} 
{\small \textsuperscript{9}FAU Erlangen-Nuremberg}
\vspace{3.5pt}
\\
\small{$^\ast$Equal contributions. $^\ddagger$Corresponding author.}
\vspace{3pt}
\\
\small{
\texttt{\{chengyan.wu, xueyun\}@m.scnu.edu.cn, bolei.ma@lmu.de}
}
}

\begin{document}
\maketitle
\begin{abstract}
Aspect-based sentiment analysis (ABSA) is a crucial task in information extraction and sentiment analysis, aiming to identify aspects with associated sentiment elements in text. However, existing ABSA datasets are predominantly English-centric, limiting the scope for multilingual evaluation and research. To bridge this gap, we present \textbf{M-ABSA}, a comprehensive dataset spanning 7 domains and 21 languages, making it the most extensive multilingual parallel dataset for ABSA to date. Our primary focus is on triplet extraction, which involves identifying aspect terms, aspect categories, and sentiment polarities. The dataset is constructed through an automatic translation process with human review to ensure quality. We perform extensive experiments using various baselines to assess performance and compatibility on M-ABSA. Our empirical findings highlight that the dataset enables diverse evaluation tasks, such as multilingual and multi-domain transfer learning, and large language model evaluation, underscoring its inclusivity and its potential to drive advancements in multilingual ABSA research. \footnote{We release our resources at \url{https://huggingface.co/datasets/Multilingual-NLP/M-ABSA} and \url{https://github.com/swaggy66/M-ABSA}.} 
\end{abstract}

\section{Introduction}

\begin{figure}[t]
\centering
  \includegraphics[width=1\columnwidth]{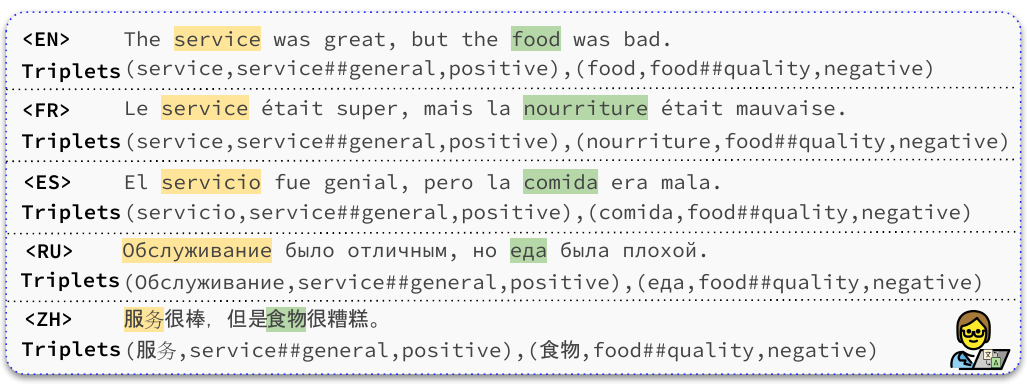}
  \caption{Example parallel sentences from the multilingual ABSA dataset. The sentiment triplet includes three elements: aspect, category, and sentiment.}
  \label{tag:introduction}
\end{figure}

Aspect-based sentiment analysis (ABSA) is a key task in fine-grained sentiment analysis, focusing on identifying opinions and sentiments associated with specific aspects. In recent years, this task has attracted considerable attention \cite{Zhang_Liu_2017, Peng_Xu_Bing_Huang_Lu_Si_2020}. Since the complexity of sentiment variations across different aspects, several studies \cite{wan2020target,zhang-etal-2021-towards-generative} have proposed using joint extraction of triplets from a sentence to tackle the challenges of fine-grained sentiment analysis, which includes three elements: aspect term, aspect category, and sentiment polarity. Given a simple example sentence in Figure \ref{tag:introduction}, ``\emph{The service was great, but the food was bad.}'', the corresponding elements are (``\emph{service}'', ``\emph{service\#\#general}'', ``\emph{positive}'') and (``\emph{food}'', ``\emph{food\#\#quality}'', ``\emph{negative}''), respectively.

In general, most existing studies focus primarily on 
monolingual datasets. For example, the English datasets from the SemEval workshops \cite{pontiki-etal-2014-semeval, pontiki-etal-2015-semeval, pontiki-etal-2016-semeval}, such as the Restaurant and Laptop datasets, have been extensively explored. 
However, this narrow focus overlooks the need for multilingual sentiment analysis in the real world. 
Recently, there have been efforts to extend ABSA datasets to multiple languages.
For instance, the SemEval workshop actually provides multilingual versions, while the dataset for each language differs in content.
\citet{zhang-etal-2021-cross} construct a multilingual dataset by automatically translating the SemEval-2016 dataset \cite{pontiki-etal-2016-semeval}, covering five languages for evaluation.
However, there is no assessment of the translation quality, and most importantly -- 
the number of languages in this dataset is limited, preventing researchers from conducting a strictly controlled evaluation of the effectiveness of cross-lingual transfer.
Moreover, this translated dataset only includes aspect terms and sentiment polarities, lacking the joint detection of aspect categories, which is crucial for ABSA, thereby limiting the scope of the evaluation. 
Therefore, a high-quality multilingual and parallel ABSA dataset is missing in the community.

To bridge this gap, this paper presents the \textbf{M-ABSA} dataset, which includes 21 languages and 7 distinct domains, making it the first comprehensive multilingual parallel ABSA dataset. 
Specifically, we use existing high-quality English datasets from multiple domains and construct a dataset (by manually annotating an English corpus) from another domain. 
These datasets are then automatically translated into 20 languages, followed by an efficient automatic data quality verification and manual inspection if necessary. 
To further investigate the quality of M-ABSA and unveil its possible usage in multilingual ABSA research,
we conduct evaluations on M-ABSA under various settings, including zero-shot cross-lingual transfer, cross-domain transfer, and zero-shot prompting with large language models (LLMs). Our contributions are as follows:
\begin{itemize}[leftmargin=*]
\item  We present a high-quality \textbf{multilingual parallel dataset for ABSA}, covering \textbf{21 typologically diverse languages} and \textbf{7 domains},
ensuring its applicability in multilingual ABSA tasks for \textbf{triplet extraction}.
\item We provide a feasible method of extending monolingual datasets to multiple languages with high quality by \textbf{automatic translation, quality evaluation, and manual inspection}.
\item We investigate the robustness and applicability of M-ABSA by a comprehensive evaluation, including \textbf{cross-lingual transfer, cross-domain transfer, and zero-shot prompting with LLMs}. Our results highlight the potential of M-ABSA in future multilingual ABSA research. 
\end{itemize}

\section{Background and Related Work}

\subsection{UABSA and TASD Tasks}
In the current work, we mainly cover two types of mainstream ABSA tasks, from UABSA to TASD. 

\textbf{Unified ABSA (UABSA)} is a basic form of ABSA tasks. It extracts aspect terms and predicts their sentiment polarities \cite{li2019unified, Chen_Qian_2020, zhang-etal-2021-towards-generative}. It can also be formulated as an (\textit{aspect term, sentiment polarity}) pair extraction problem \cite{zhang-etal-2021-towards-generative}. For the example ``\textit{The service was great, but the food was bad.}'', it aims to extract two pairs: (\textit{service, positive}) and (\textit{food, negative}). 

\textbf{Target-Aspect-Sentiment Detection (TASD)} is an extended task for UABSA with an additional aspect category, which belongs to a pre-defined category set. It aims to detect all (\textit{aspect term, aspect category, sentiment polarity}) triplets for a given sentence \cite{wan2020target, zhang-etal-2021-towards-generative}. For the same example sentence in the previous paragraph, it should extract the triplets: (\textit{service, service\#\#general, positive}) and (\textit{food, food\#\#quality, negative}), where the ``\textit{service\#\#general}'' and ``\textit{food\#\#quality}'' correspond to the categories of the respective aspect terms.

\subsection{Current Multilingual ABSA Datasets} 
Research in multilingual ABSA has been constrained by limited datasets, with most focusing on English. SemEval workshops introduced early datasets for restaurant and laptop reviews \citep{pontiki-etal-2014-semeval, pontiki-etal-2015-semeval}, later expanding to Chinese, Turkish, and Spanish \citep{pontiki-etal-2016-semeval}. Other efforts, like GermEval-2017 for German \citep{wojatzki2017germeval}, ABSITA-2020 for Italian \citep{de2020ate}, SemEval-2023 for African languages \citep{muhammad2023semeval}, and Polish-ASTE \cite{lango-etal-2024-polish}, added linguistic diversity but often focused on sentence-level analysis. Recent datasets like ROAST extended coverage to two other low-resource languages, Hindi and Telugu, and incorporated review-level ABSA analysis across multiple domains \citep{chebolu2024roast}. Despite these advances, existing datasets lack broad domain coverage, linguistic variety, and parallelism essential for robust multilingual evaluation. Also, current datasets limit in standardization and diversity to make it possible for broader model robustness checks \cite{chebolu-etal-2023-review}. There is a new multi-domain dataset \cite{Cai2025}, but the language is English. Therefore, broadening the variety of ABSA datasets is crucial for advancing both research and applications.

\begin{figure*}[htbp] 
\centering 
\includegraphics[width=1\linewidth]{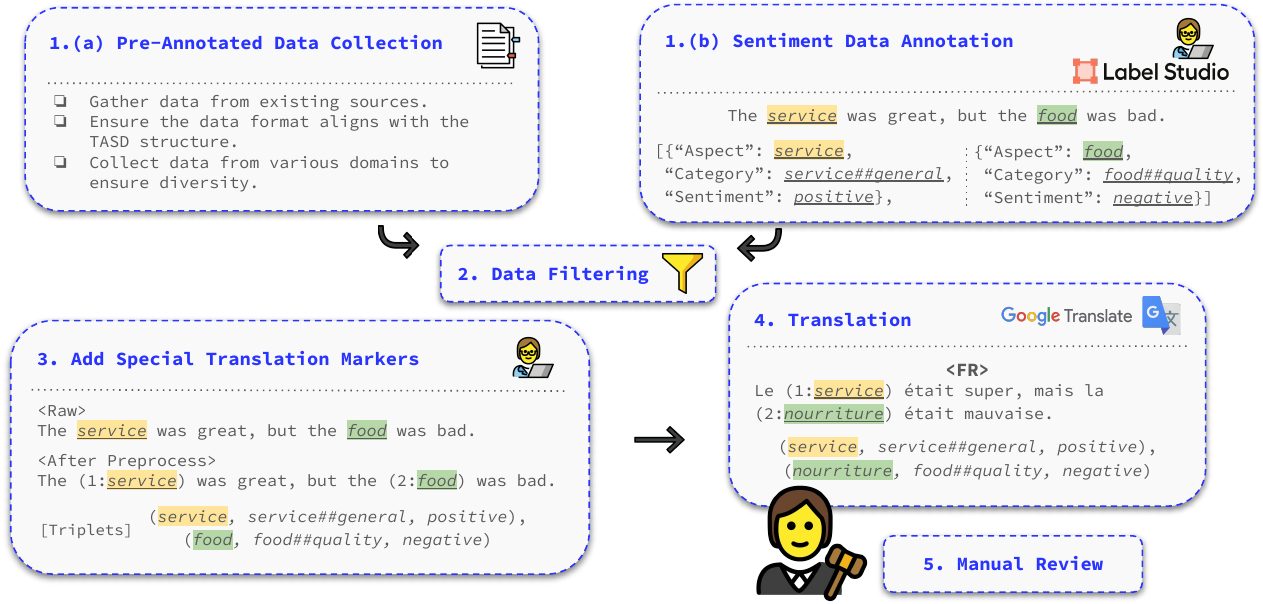} 
\caption{The construction process of the M-ABSA dataset.} 
\label{fig:construction_process} 
\end{figure*}

\subsection{Methods for Multilingual ABSA}
Early multilingual ABSA methods rely on supervised learning with annotated data, using translation systems and alignment algorithms for cross-lingual label projection \citep{lin2014cross, klinger2015instance, lambert2015aspect, barnes2016exploring}. However, these methods struggle with scalability and translation quality \citep{zhou2015clopinionminer}. Subsequent works leverage cross-lingual word embeddings trained on parallel corpora to align representations across languages, enabling language-agnostic ABSA \citep{barnes2016exploring, wang2018transition}. Multilingual pre-trained language models
further improved performance through fine-tuning and data augmentation \citep{xu2019bert, phan2021exploring}. More recently, large language models have been explored for improving training data \cite{MAI2024112289} and evaluating zero-shot multilingual ABSA \citep{wu2024evaluating}. Although these models offer scalability, they still fall short of fine-tuned models in low-resource scenarios. In this work, we evaluate basic baselines on our curated multilingual dataset, establishing a new benchmark to support future ABSA research.

\section{Dataset Construction Process}
This section outlines the process for constructing the M-ABSA dataset. Given the abundance of existing ABSA datasets in English and the scarcity in other languages, our primary goal is to collect English datasets from various domains and translate them into target languages to create a high-quality multilingual parallel ABSA dataset. Figure \ref{fig:construction_process} illustrates the overall process.

The process begins with the collection of existing English datasets, which can follow two approaches: a) using pre-annotated datasets for ABSA triplet extraction tasks, or b) sourcing sentiment datasets that lack triplet-specific annotations and manually annotating them to fit the triplet format. For pre-annotated datasets, we evaluate label completeness and extract subsets suitable for our purposes. To enhance domain diversity, we may need to create new datasets by identifying suitable data sources for sentiment analysis and recruiting annotators to label them according to a predefined triplet schema, as in the second case. All collected datasets are balanced in size. Details of this data collection phase are provided in \S\ref{sec:datasets}.

Once the datasets are ready, we automatically translate them into target languages, focusing on both sentence-level translations and entity-level translations. This approach aligns with previous relation extraction studies, such as those by \citet{bassignana-plank-2022-crossre} and \citet{bassignana-etal-2023-multi}, and text generation tasks, such as \citet{chen-etal-2022-mtg}. Following machine translation, human reviewers check for errors and omissions. The quality of the translated datasets is then assessed. Further details on the translation and evaluation are outlined in \S\ref{sec:translation}.

\section{Datasets}
\label{sec:datasets}
\subsection{Collecting Existing Datasets}
\label{sec:collection}
We first collect six existing annotated English ABSA datasets from multiple domains. Each dataset is introduced in the following.

\textbf{Hotel.} This dataset is based on TripAdvisor review data from \citet{yin-etal-2017-document}, featuring over 100K hotel reviews. \citet{chebolu-etal-2024-oats} extended the dataset for the opinion aspect target sentiment (OATS) \footnote{The OATS refers to the quadruple extraction task that includes the aspect term, aspect category, sentiment polarity and the opinion words. Moreover, it includes both the sentence-level quadruples and the review-level tuples \cite{chebolu-etal-2024-oats}.} quadruple extraction task by annotating a subset of sentences. We select 2,147 annotated sentences from this subset.

\textbf{Food.} This dataset includes approximately 500K Amazon fine food reviews curated from a Kaggle competition. \citet{chebolu-etal-2024-oats} leveraged this dataset for the OATS task. We select 2,136 annotated sentences from this collection.

\textbf{Coursera.} This dataset contains around 100K reviews from Coursera, focusing on course quality, content, comprehensiveness, etc, sourced from a Kaggle competition.
\citet{chebolu-etal-2024-oats} also employed this dataset for the OATS task, from which we select 2,156 annotated sentences.

\textbf{Phone.} This dataset, created by \citet{zhou-etal-2023-unified-one}, focuses on the OATS task. 
It includes reviews from various e-commerce platforms, collected in mid-2021, covering 12 cellphone brands. We select a subset of 2,109 sentences.

\textbf{Laptop.} 
This dataset features reviews from the Amazon platform between 2017 and 2018, covering ten types of laptops across six brands \citep{cai-etal-2021-aspect}. We select 2,122 sentences out of 4,076 review sentences.

\textbf{Restaurant.} This dataset contains customer reviews for restaurants. It is one of the domain datasets from the SemEval 2016 \cite{pontiki-etal-2016-semeval} and is constructed for TASD tasks with the aspect-based sentiment triplets. Although the dataset provides reviews for 5 languages including English, reviews are not parallel.
We select 2,124 reviews from English to create our parallel dataset.

\subsection{Constructing a New Dataset}
\label{sec:annotation}

In addition to the six existing annotated datasets discussed in \S\ref{sec:collection}, we develop a new dataset from a different domain to enhance the dataset's diversity.

\textbf{Sight.} The Sight dataset, introduced by \citet{wang-etal-2023-sight}, consists of 15,784 comments on online math lectures from the Massachusetts Institute of Technology OpenCourseWare (MIT OCW) YouTube channel. These comments pertain to specific math lectures.
While the original dataset provides sentence-level sentiment annotations, we extend it by annotating sentiment triplets. We select 1,986 sentences for this purpose.

\textbf{}
Six annotators are invited for data annotation using Label Studio.\footnote{\url{https://labelstud.io}}
Following the annotation guidelines from the SemEval 2016 Task 5 \cite{pontiki-etal-2016-semeval}, we provide a comprehensive guide document to introduce them to the topics, dataset, tool, and specific annotation requirements. We divide all the data into six subsets, with three annotators involved in each subset. 
Two annotators independently annotate a subset and another annotator resolves disagreements between the two annotators.
Additional annotation details can be found in \S\ref{sec:annotation_process}.

Following \citet{kim2018feels,mayhew2023universal}, we report the inter-annotator agreement with  F1 score for the annotated aspect terms and sentiment triplets.
For F1, 
we use the results of the 
first annotator as the predictions and the results of the second annotator as the reference. The F1 score for aspect terms and sentiment triplets are 82.57\% and 80.70\%, respectively, reflecting a high level of agreement between two annotators \cite{kim2018feels, zhou-etal-2023-unified-one}.

\subsection{Dataset Statistics and Characteristics}
By grouping the sentences from each domain, we present our newly collected multi-domain English dataset for ASBA. 
Table \ref{tab:en_data_statistic} shows the key statistics across the seven domains. 
We ensure that each domain contained approximately 2,000 sentences to exclude the confounding factor of data size for cross-domain transfer (\S\ref{sec:cross_domain}). The aspect counts vary across domains, with the Phone domain having the highest number. 
Notably, all datasets except the Phone domain include many implicit aspects -- sentiment entities that are not explicitly mentioned in the sentence. Such implicit aspects are very frequent because many sentences contain indefinite pronouns like ``it'', which are marked as ``NULL'' and labeled with their corresponding categories and sentiments \cite{chebolu-etal-2023-review}.
The outlier -- the Phone dataset -- simply did not annotate such implicit aspects.  \S\ref{sec:detailed_aspect_categories} presents the detailed aspect categories in each domain.

\begin{table}[htbp]
\setlength\tabcolsep{2pt}
\renewcommand\arraystretch{0.85}
\tiny
\centering
\begin{tabular}{rr|rrrr|rrrr|r}
\toprule
& & \multicolumn{4}{|c|}{Sentences} & \multicolumn{4}{|c|}{Aspects} & \multicolumn{1}{|c}{Cat.} \\
\midrule
& Dataset & Train & Dev & Test & All & Train & Dev & Test & All & All  \\
\midrule
& Cours.    & 1278 & 312 & 566 & 2156 & 1198 (423) & 372 (113) & 744 (263) & 2314 (799) & 30 \\
& Food        & 1278 & 336 & 522 & 2136 & 1075 (698) & 338 (209) & 742 (414) & 2155 (1321) & 13 \\
& Hotel       & 1255 & 308 & 584 & 2147 & 1631 (381) & 354 (74)  & 975 (261) & 2960 (716) & 37 \\
& Laptop      & 1264 & 326 & 532 & 2122 & 1758 (392) & 440 (88)  & 757 (147) & 2955 (627) & 114 \\
& Phone       & 1279 & 307 & 523 & 2109 & 2883 (0)   & 710 (0)   & 1217 (0)  & 4810 (0)   & 88 \\
& Res.  & 1264 & 316 & 544 & 2124 & 1989 (446) & 507 (104) & 799 (179) & 3295 (729) & 12 \\
& Sight       & 1181 & 281 & 524 & 1986 & 1378 (203) & 346 (46)  & 836 (130) & 2560 (379) & 5 \\
\bottomrule
\end{tabular}
\caption{Statistics of the English ABSA dataset before translation. In the aspects column, X (Y) represents the total number of aspects X, with Y denoting the number of implicit aspects.}
\label{tab:en_data_statistic}
\end{table}

\section{Annotation Projection and Evaluation}
\label{sec:translation}

\subsection{ Projection with Translation}
We propose an effective yet straightforward method to project span-based annotations, specifically for aspect entities in ABSA tasks, to extensive languages.
We use the Google Translate API\footnote{\url{https://cloud.google.com/translate}} 
to translate the English data to multiple languages while preserving the label classes including aspect categories and their associated sentiments in English.

\textbf{Selecting Languages.}  We include 20 typologically diverse languages: Arabic (ar), Chinese (zh), Croatian (hr), Danish (da), Dutch (nl), French (fr), German (de), Hindi (hi), Indonesian (id), Japanese (ja), Korean (ko), Portuguese (pt), Russian (ru), Slovak (sk), Spanish (es), Swahili (sw), Swedish (sv), Thai (th), Turkish (tr), Vietnamese (vi). 
These languages are selected to ensure good coverage of languages from different language families, including Indo-European, Sino-Tibetan, Afro-Asiatic, and Austroasiatic, among others.

\textbf{Preserved Translation.} 
Similar to \citet{zhang-etal-2021-cross}, we add special aspect markers to surround the tokens (aspects) in a sentence before translation. Through extensive validation, we select ``( )'' and ensure these markers are preserved after translation while tokens inside the markers can be correctly translated into the target language. In addition, within each surrounding marker, we add a number to denote its sequential order of appearance in the sentence, as shown in Figure \ref{fig:construction_process}. This setup allows us to easily project the labels to the target-language aspects after translation, without losing track of the correct order of aspects in the target language (the order of sentence components can vary in the target language compared with English).

\textbf{Manual Review and Revision.} 
After annotation projection to the 20 languages, the resulting multilingual dataset undergoes manual review to identify two primary error types: (1) \textit{non-translations}: the original English aspect terms remain untranslated, and (2) \textit{omissions}: aspect terms are missing in the translated text. 
Table \ref{tab:translaton-quantify-errors} presents the error statistics across languages, showing that these errors constitute only a small fraction of the dataset, ensuring overall translation quality. 
For each identified error, the annotators then manually check and update the respective data based on Google Translate. 
Details of the proofreading process are provided in \S\ref{sec:manual_proofreading}.

\begin{table}[htbp]
\centering
\footnotesize
\setlength\tabcolsep{5pt}
\begin{tabular}{cccc}
\toprule
Dataset & non-translations (\%) & omission (\%) & ALL\\
\midrule
Coursera & 6234 (13.47\%) & 393 (0.85\%) & 46280 \\
Food & 7646 (17.75\%) & 369 (0.86\%) & 43100 \\
Hotel & 5600 (\ 9.46\%) & 456 (0.77\%) & 59200 \\
Laptop & 9714 (16.44\%) & 375 (0.63\%) & 59100 \\
Phone & 5899 (\ 6.13\%) & 437 (0.45\%) & 96200 \\
Restaurant & 8152 (12.37\%) & 283 (0.43\%) & 65900 \\
Sight & 9800 (19.14\%) & 398 (0.78\%) & 51200  \\
\bottomrule
\end{tabular}
\caption{Statistics of \textit{non-translations} and \textit{omissions} in each dataset after translating English to all target languages. We report the sum over all target languages and also the percentage over the count of all aspects (ALL).}
\label{tab:translaton-quantify-errors}
\end{table}

\subsection{Automatic Evaluation of Translation Quality}
\label{tab:main translation quality}
We evaluate the translation quality for all 20 languages. Specifically, we randomly sample 100, 20, and 30 data items from the train, validation, and test set from each data domain for each language. We then concatenate the sampled items from all domains. This results in 1,050 items for each language (150 $\times$ 7 = 1,050). Finally, we evaluate the translation quality for each language individually using the sampled items as a proxy from two perspectives: \textbf{consistency} and \textbf{faithfulness}, as discussed below.\footnote{The details of the translation quality evaluation and per-domain evaluation (in comparison with a random-paired baseline) are presented in \S\ref{sec:translation quality details}}

\textbf{Consistency.} Since we insert special aspect markers into the English data, this might influence the translation. Therefore, we want to ensure the translation is consistent with or without the special aspect markers. 
To do this, we translate the same 1,050 items from the English data to all target languages, without using the special aspect markers. Based on the new translations ($\text{translation}_{\text{w/o marker}}$) and the sampled data ($\text{translation}_{\text{w/ marker}}$) for each language, we consider two metrics. 1) \textbf{Aspect accuracy (Acc)} measures the percentage of all aspects in the 1,050 items also presenting in $\text{translation}_{\text{w/o marker}}$. 2) \textbf{chrF++} \citep{popovic-2017-chrf} measures the character-based translation quality using $\text{translation}_{\text{w/o marker}}$ as references and $\text{translation}_{\text{w/ marker}}$ as hypotheses. 

\textbf{Faithfulness.} We also want to ensure our translations convey the same meanings as the source data in English. 
To do this, we back-translate the 1,050 items from each language to English. 
Then, we compare the back-translations with the original English data. 
Similar back-translation-based approaches have also been used to evaluate the translation quality or filter unreliable translations \citep{sobrevilla-cabezudo-etal-2019-back,sekizawa-etal-2023-constructing}.
Our evaluation includes three metrics. 1) \textbf{BERTScore} \citep{zhang2020bertscore} measures how semantically similar the back-translations and the original English data are in the representation space. 2) \textbf{SBERT} \citep{reimers-gurevych-2019-sentence} also measures the semantic similarity as BERTScore. 3) \textbf{BLEU} \citep{papineni-etal-2002-bleu} measures the $n$-gram translation quality using the original English data as references and back-translations as hypotheses. 

\textbf{} 
The results are shown in Table \ref{tab:translaton-quality}. We see that both Acc and chrF++ are high across languages, indicating good consistency. However, Acc is not perfect, suggesting adding special aspect markers will influence the translation of aspects to some degree. Similarly, high BERTScore, SBERT, and BLEU show that our datasets of languages other than English are faithful in keeping the same meanings as the original English. In summary, good consistency and faithfulness demonstrate the effectiveness of our method and guarantee the quality of our multilingual parallel dataset M-ABSA.  We show additional details of translation quality evaluation in single domains and cross-domains in \S\ref{sec:details_quality}.

\begin{table}[htbp]
\centering
\setlength\tabcolsep{4pt}
\footnotesize
\begin{tabular}{cccccc}
\toprule
Lang. & Acc & chrF++ & BERTScore & SBERT & BLEU \\
\midrule
ar & 68.44 & 86.58 & 95.66 & 90.81 & 53.97 \\
da & 77.82 & 93.69 & 96.88 & 95.30 & 68.51 \\
de & 81.77 & 91.16 & 95.93 & 92.27 & 54.30 \\
es & 76.92 & 92.51 & 96.04 & 92.39 & 58.26 \\
fr & 74.50 & 92.63 & 96.22 & 92.66 & 59.39 \\
hi & 79.50 & 89.36 & 95.89 & 92.40 & 56.00 \\
hr & 66.37 & 89.98 & 96.15 & 93.29 & 59.95 \\
id & 67.87 & 90.09 & 95.72 & 90.96 & 52.43 \\
ja & 80.25 & 78.59 & 94.52 & 88.89 & 39.16 \\
ko & 76.37 & 80.97 & 94.21 & 86.85 & 37.57 \\
nl & 75.40 & 91.21 & 96.05 & 92.53 & 57.90 \\
pt & 78.44 & 93.20 & 96.26 & 92.91 & 60.37 \\
ru & 66.19 & 89.73 & 95.39 & 90.39 & 50.09 \\
sk & 62.31 & 89.54 & 96.14 & 93.54 & 58.76 \\
sv & 78.96 & 92.93 & 96.83 & 94.66 & 67.04 \\
sw & 60.34 & 88.28 & 95.79 & 90.91 & 58.18 \\
th & 74.48 & 81.57 & 94.39 & 87.01 & 39.00 \\
tr & 68.04 & 89.21 & 95.23 & 89.97 & 48.53 \\
vi & 70.72 & 91.08 & 95.43 & 89.79 & 50.22 \\
zh & 81.24 & 77.20 & 94.91 & 89.16 & 43.96 \\
\midrule
Avg. & 73.47 & 88.94 & 95.70 & 91.41 & 53.84 \\
\bottomrule
\end{tabular}
\caption{Translation quality evaluation using different automatic metrics.}
\label{tab:translaton-quality}
\end{table}

\subsection{Human Evaluation of Translation Quality}
\label{tab:human translation quality}

In addition to the automatic evaluation, we randomly sampled 70 sentences in total (10 sentences per domain) for each of eight representative target languages (ar, de, es, hi, ja, ru, th, zh), covering four language families and seven scripts.
For each language, we recruited three native speakers (with high proficiency in English) via Prolific\footnote{\url{https://www.prolific.com/}} and compensated them at an hourly rate of £9.

Each annotator received the English source sentences (with the special aspect markers removed) and the translations. They rated the system output on four task‑agnostic dimensions widely used in MT evaluation:
\textbf{Grammar}, \textbf{Fluency}, \textbf{Adequacy}, and \textbf{Code-Switching}, using a 5-scale rating from 1 (poor) to 5 (excellent), based on previous MT evaluation works \cite{chen-etal-2022-mtg}. Additionally, given that aspect terms were also translated, we introduced a fifth evaluation dimension: \textbf{Aspect Term Translation}, also rated on a 5-point scale. Further details and examples of the human evaluation process are provided in \S\ref{sec:human-eval-appendix}.

We report the results of the human evaluation as average ratings for each language and average across all languages in Table \ref{tab:huaman_translaton-quality}. Overall, we observe strong human evaluation performance across all five dimensions, with particularly high ratings in \textbf{Aspect Term Translation} and \textbf{Code-Switching}. These results confirm that our translations are not only grammatically and semantically sound but also preserve critical task-specific information of the aspect terms, thereby supporting that M-ABSA is a high-quality benchmark for multilingual ABSA.

\begin{table}[htbp]
\centering
\setlength\tabcolsep{3pt}
\footnotesize
\begin{tabular}{cccccc}
\toprule
Lang. & Gram. & Flu. & Adeq. & C.-Switch. & Asp. Term \\
\midrule
ar  &4.04 &3.98&4.33&4.72&4.63 \\
de  &4.30 &4.22&4.68&4.98&4.82\\
es  &4.03&3.66&4.36&4.63&4.77 \\
hi  &3.31 &2.84&3.16&3.12&2.97 \\
ja  &3.09 &2.93&4.01&3.77&3.76 \\
ru  &3.77 &3.59&4.35&4.73&4.64 \\
th  &3.35&2.81&3.30&4.56&4.07 \\
zh  &4.09 &3.85&4.23&4.58&4.47 \\
\midrule
Avg. &3.75 &3.61&4.05&4.39&4.27 \\
\bottomrule
\end{tabular}
\caption{Human evaluation scores (1–5, 1 is worst and 5 is best) averaged over eight languages.}
\label{tab:huaman_translaton-quality}
\end{table}

\begin{table*}[ht]
	\centering  
	\subfigure[\textbf{TASD} Results]{  
		\begin{minipage}{.485\linewidth}
			\centering 
\setlength\tabcolsep{3pt}
\scriptsize
\begin{tabular}{c|ccccccc|c}
\toprule 
Lang. & Coursera & Food & Hotel & Laptop & Phone & Res. & Sight & Avg. \\
\midrule
ar & \cellcolor{purple!11.95} 11.95 & \cellcolor{purple!17.21} 17.21 & \cellcolor{purple!18.08} 18.08 & \cellcolor{purple!18.73} 18.73 & \cellcolor{purple!16.97} 16.97 & \cellcolor{purple!29.88} 29.88 & \cellcolor{purple!13.39} 13.39 & \cellcolor{purple!18.03} 18.03 \\ 
da & \cellcolor{purple!23.18} 23.18 & \cellcolor{purple!27.81} 27.81 & \cellcolor{purple!18.41} 18.41 & \cellcolor{purple!30.84} 30.84 & \cellcolor{purple!21.27} 21.27 & \cellcolor{purple!52.20} 52.20 & \cellcolor{purple!29.93} 29.93 & \cellcolor{purple!29.52} 29.52 \\ 
de & \cellcolor{purple!28.74} 28.74 & \cellcolor{purple!34.33} 34.33 & \cellcolor{purple!32.03} 32.03 & \cellcolor{purple!27.77} 27.77 & \cellcolor{purple!27.31} 27.31 & \cellcolor{purple!57.20} 57.20 & \cellcolor{purple!28.52} 28.52 & \cellcolor{purple!33.99} 33.99 \\ 
es & \cellcolor{purple!23.71} 23.71 & \cellcolor{purple!23.83} 23.83 & \cellcolor{purple!36.77} 36.77 & \cellcolor{purple!22.46} 22.46 & \cellcolor{purple!20.55} 20.55 & \cellcolor{purple!50.44} 50.44 & \cellcolor{purple!23.56} 23.56 & \cellcolor{purple!28.33} 28.33 \\ 
fr & \cellcolor{purple!29.82} 29.82 & \cellcolor{purple!26.83} 26.83 & \cellcolor{purple!19.49} 19.49 & \cellcolor{purple!24.91} 24.91 & \cellcolor{purple!24.37} 24.37 & \cellcolor{purple!52.32} 52.32 & \cellcolor{purple!23.50} 23.50 & \cellcolor{purple!28.89} 28.89 \\ 
hi & \cellcolor{purple!13.42} 13.42 & \cellcolor{purple!22.36} 22.36 & \cellcolor{purple!22.19} 22.19 & \cellcolor{purple!23.41 } 23.41 & \cellcolor{purple!21.48} 21.48 & \cellcolor{purple!35.19} 35.19 & \cellcolor{purple!10.38} 10.38 & \cellcolor{purple!21.06} 21.06 \\ 
hr & \cellcolor{purple!11.26} 11.26 & \cellcolor{purple!11.39} 11.39 & \cellcolor{purple!31.99} 31.99 & \cellcolor{purple!17.48} 17.48 & \cellcolor{purple!16.00} 16.00 & \cellcolor{purple!31.59 } 31.59 & \cellcolor{purple!15.74} 15.74 & \cellcolor{purple!19.49 } 19.49 \\ 
id & \cellcolor{purple!24.85} 24.85 & \cellcolor{purple!22.92} 22.92 & \cellcolor{purple!31.01} 31.01 & \cellcolor{purple!30.27} 30.27 & \cellcolor{purple!21.65} 21.65 & \cellcolor{purple!41.80} 41.80 & \cellcolor{purple!26.84} 26.84 & \cellcolor{purple!28.33} 28.33 \\ 
ja & \cellcolor{purple!16.58} 16.58 & \cellcolor{purple!26.26} 26.26 & \cellcolor{purple!25.30} 25.30 & \cellcolor{purple!22.22} 22.22 & \cellcolor{purple!26.72} 26.72 & \cellcolor{purple!41.06 } 41.06 & \cellcolor{purple!20.87} 20.87 & \cellcolor{purple!25.86} 25.86 \\ 
ko & \cellcolor{purple!11.11} 11.11 & \cellcolor{purple!22.96} 22.96 & \cellcolor{purple!13.36} 13.36 & \cellcolor{purple!18.01} 18.01 & \cellcolor{purple!16.25} 16.25 & \cellcolor{purple!27.46} 27.46 & \cellcolor{purple!17.89} 17.89 & \cellcolor{purple!18.86} 18.86 \\ 
nl & \cellcolor{purple!27.86} 27.86 & \cellcolor{purple!25.27} 25.27 & \cellcolor{purple!26.30} 26.30 & \cellcolor{purple!33.54} 33.54 & \cellcolor{purple!30.20} 30.20 & \cellcolor{purple!47.59} 47.59 & \cellcolor{purple!27.20} 27.20 & \cellcolor{purple!31.85} 31.85 \\ 
pt & \cellcolor{purple!30.57} 30.57 & \cellcolor{purple!22.83} 22.83 & \cellcolor{purple!30.51} 30.51 & \cellcolor{purple!26.63} 26.63 & \cellcolor{purple!22.69} 22.69 & \cellcolor{purple!49.50} 49.50 & \cellcolor{purple!26.78} 26.78 & \cellcolor{purple!29.79} 29.79 \\ 
ru & \cellcolor{purple!19.38} 19.38 & \cellcolor{purple!23.36} 23.36 & \cellcolor{purple!20.57} 20.57 & \cellcolor{purple!18.37} 18.37 & \cellcolor{purple!20.06 } 20.06 & \cellcolor{purple!39.89} 39.89 & \cellcolor{purple!16.44} 16.44 & \cellcolor{purple!22.58} 22.58 \\ 
sk & \cellcolor{purple!21.74} 21.74 & \cellcolor{purple!21.17} 21.17 & \cellcolor{purple!23.49} 23.49 & \cellcolor{purple!30.65} 30.65 & \cellcolor{purple!25.54} 25.54 & \cellcolor{purple!46.75} 46.75 & \cellcolor{purple!20.51} 20.51 & \cellcolor{purple!27.84} 27.84 \\ 
sv & \cellcolor{purple!27.56} 27.56 & \cellcolor{purple!27.87} 27.87 & \cellcolor{purple!22.12} 22.12 & \cellcolor{purple!27.76} 27.76 & \cellcolor{purple!24.48} 24.48 & \cellcolor{purple!53.30} 53.30 & \cellcolor{purple!27.51} 27.51 & \cellcolor{purple!30.51} 30.51 \\ 
sw & \cellcolor{purple!10.07} 10.07 & \cellcolor{purple!11.06} 11.06 & \cellcolor{purple!10.35} 10.35 & \cellcolor{purple!16.65 } 16.65 & \cellcolor{purple!13.66} 13.66 & \cellcolor{purple!27.61} 27.61 & \cellcolor{purple!17.34} 17.34 & \cellcolor{purple!15.96} 15.96 \\ 
th & \cellcolor{purple!17.71} 17.71 & \cellcolor{purple!25.35} 25.35 & \cellcolor{purple!20.90} 20.90 & \cellcolor{purple!22.56} 22.56 & \cellcolor{purple!23.39} 23.39 & \cellcolor{purple!43.33} 43.33 & \cellcolor{purple!23.02} 23.02 & \cellcolor{purple!25.32} 25.32 \\ 
tr & \cellcolor{purple!17.62} 17.62 & \cellcolor{purple!16.77} 16.77 & \cellcolor{purple!17.71} 17.71 & \cellcolor{purple!20.88} 20.88 & \cellcolor{purple!18.55} 18.55 & \cellcolor{purple!37.65} 37.65 & \cellcolor{purple!19.39} 19.39 & \cellcolor{purple!21.37} 21.37 \\ 
vi & \cellcolor{purple!10.19} 10.19 & \cellcolor{purple!19.83} 19.83 & \cellcolor{purple!15.69} 15.69 & \cellcolor{purple!19.59} 19.59 & \cellcolor{purple!15.86} 15.86 & \cellcolor{purple!33.06} 33.06 & \cellcolor{purple!24.84} 24.84 & \cellcolor{purple!19.87} 19.87 \\ 
zh & \cellcolor{purple!18.17} 18.17 & \cellcolor{purple!24.98} 24.98 & \cellcolor{purple!24.45} 24.45 & \cellcolor{purple!22.55} 22.55 & \cellcolor{purple!26.57} 26.57 & \cellcolor{purple!47.24} 47.24 & \cellcolor{purple!24.18} 24.18 & \cellcolor{purple!26.59 } 26.59 \\
\midrule
en & \cellcolor{purple!48.36} 48.36 & \cellcolor{purple!48.87} 48.87 & \cellcolor{purple!40.69} 40.69 & \cellcolor{purple!43.54} 43.54 & \cellcolor{purple!47.24} 47.24 & \cellcolor{purple!66.34} 66.34 & \cellcolor{purple!36.72} 36.72 & \cellcolor{purple!47.68} 47.68 \\
\bottomrule
\end{tabular}
\label{tab:main_result_tasd}
		\end{minipage}
	}
  \subfigure[\textbf{UABSA} Results]{ 
		\begin{minipage}{.485\linewidth}
			\centering 
\setlength\tabcolsep{3pt}
\scriptsize
\begin{tabular}{c|ccccccc|c}
\toprule 
Lang. & Coursera & Food & Hotel & Laptop & Phone & Res. & Sight & Avg. \\
\midrule
ar & \cellcolor{purple!26.22} 26.22 & \cellcolor{purple!24.64} 24.64 & \cellcolor{purple!27.50} 27.50 & \cellcolor{purple!38.54} 38.54 & \cellcolor{purple!27.14} 27.14 & \cellcolor{purple!34.84} 34.84 & \cellcolor{purple!18.97} 18.97 & \cellcolor{purple!28.41} 28.41 \\ 
da & \cellcolor{purple!41.73} 41.73 & \cellcolor{purple!50.88} 50.88 & \cellcolor{purple!43.92} 43.92 & \cellcolor{purple!62.42} 62.42 & \cellcolor{purple!29.67} 29.67 & \cellcolor{purple!60.90} 60.90 & \cellcolor{purple!30.25} 30.25 & \cellcolor{purple!45.68} 45.68 \\ 
de & \cellcolor{purple!52.67} 52.67 & \cellcolor{purple!61.12} 61.12 & \cellcolor{purple!45.54} 45.54 & \cellcolor{purple!54.50} 54.50 & \cellcolor{purple!42.83} 42.83 & \cellcolor{purple!67.63} 67.63 & \cellcolor{purple!32.52} 32.52 & \cellcolor{purple!50.55} 50.55 \\ 
es & \cellcolor{purple!48.96} 48.96 & \cellcolor{purple!43.65} 43.65 & \cellcolor{purple!41.03} 41.03 & \cellcolor{purple!44.89} 44.89 & \cellcolor{purple!28.47} 28.47 & \cellcolor{purple!57.74} 57.74 & \cellcolor{purple!29.02} 29.02 & \cellcolor{purple!42.68} 42.68 \\ 
fr & \cellcolor{purple!48.92} 48.92 & \cellcolor{purple!41.43} 41.43 & \cellcolor{purple!34.08} 34.08 & \cellcolor{purple!48.65} 48.65 & \cellcolor{purple!33.98} 33.98 & \cellcolor{purple!54.59} 54.59 & \cellcolor{purple!26.96} 26.96 & \cellcolor{purple!41.94} 41.94 \\ 
hi & \cellcolor{purple!31.21} 31.21 & \cellcolor{purple!37.29} 37.29 & \cellcolor{purple!32.45} 32.45 & \cellcolor{purple!42.50} 42.50 & \cellcolor{purple!35.91} 35.91 & \cellcolor{purple!40.68} 40.68 & \cellcolor{purple!18.91} 18.91 & \cellcolor{purple!34.42} 34.42 \\ 
hr & \cellcolor{purple!29.15} 29.15 & \cellcolor{purple!19.46} 19.46 & \cellcolor{purple!35.25} 35.25 & \cellcolor{purple!39.18} 39.18 & \cellcolor{purple!24.08} 24.08 & \cellcolor{purple!53.31} 53.31 & \cellcolor{purple!24.50} 24.50 & \cellcolor{purple!32.13} 32.13 \\ 
id & \cellcolor{purple!46.39} 46.39 & \cellcolor{purple!47.91} 47.91 & \cellcolor{purple!39.51} 39.51 & \cellcolor{purple!60.73} 60.73 & \cellcolor{purple!32.85} 32.85 & \cellcolor{purple!50.07} 50.07 & \cellcolor{purple!31.10} 31.10 & \cellcolor{purple!44.37} 44.37 \\ 
ja & \cellcolor{purple!31.56} 31.56 & \cellcolor{purple!47.14} 47.14 & \cellcolor{purple!35.93} 35.93 & \cellcolor{purple!44.93} 44.93 & \cellcolor{purple!39.40} 39.40 & \cellcolor{purple!48.25} 48.25 & \cellcolor{purple!26.96} 26.96 & \cellcolor{purple!39.45} 39.45 \\ 
ko & \cellcolor{purple!25.62} 25.62 & \cellcolor{purple!31.00} 31.00 & \cellcolor{purple!31.89} 31.89 & \cellcolor{purple!41.60} 41.60 & \cellcolor{purple!22.29} 22.29 & \cellcolor{purple!32.37} 32.37 & \cellcolor{purple!24.50} 24.50 & \cellcolor{purple!29.89} 29.89 \\ 
nl & \cellcolor{purple!53.03} 53.03 & \cellcolor{purple!52.03} 52.03 & \cellcolor{purple!49.35} 49.35 & \cellcolor{purple!64.71} 64.71 & \cellcolor{purple!38.57} 38.57 & \cellcolor{purple!55.57} 55.57 & \cellcolor{purple!32.07} 32.07 & \cellcolor{purple!49.76} 49.76 \\ 
pt & \cellcolor{purple!50.34} 50.34 & \cellcolor{purple!35.35} 35.35 & \cellcolor{purple!48.51} 48.51 & \cellcolor{purple!61.12} 61.12 & \cellcolor{purple!30.74} 30.74 & \cellcolor{purple!63.40} 63.40 & \cellcolor{purple!29.75} 29.75 & \cellcolor{purple!45.89} 45.89 \\ 
ru & \cellcolor{purple!34.67} 34.67 & \cellcolor{purple!43.45} 43.45 & \cellcolor{purple!38.47} 38.47 & \cellcolor{purple!42.69} 42.69 & \cellcolor{purple!25.71} 25.71 & \cellcolor{purple!47.58} 47.58 & \cellcolor{purple!19.19} 19.19 & \cellcolor{purple!36.25} 36.25 \\ 
sk & \cellcolor{purple!42.27} 42.27 & \cellcolor{purple!35.95} 35.95 & \cellcolor{purple!39.48} 39.48 & \cellcolor{purple!56.45} 56.45 & \cellcolor{purple!34.80} 34.80 & \cellcolor{purple!61.90} 61.90 & \cellcolor{purple!27.49} 27.49 & \cellcolor{purple!42.62} 42.62 \\ 
sv & \cellcolor{purple!47.23} 47.23 & \cellcolor{purple!42.58} 42.58 & \cellcolor{purple!36.09} 36.09 & \cellcolor{purple!55.02} 55.02 & \cellcolor{purple!37.90} 37.90 & \cellcolor{purple!61.44} 61.44 & \cellcolor{purple!29.71} 29.71 & \cellcolor{purple!44.14} 44.14 \\ 
sw & \cellcolor{purple!25.61} 25.61 & \cellcolor{purple!19.83} 19.83 & \cellcolor{purple!22.79} 22.79 & \cellcolor{purple!34.34} 34.34 & \cellcolor{purple!24.90} 24.90 & \cellcolor{purple!46.10} 46.10 & \cellcolor{purple!25.95} 25.95 & \cellcolor{purple!28.93} 28.93 \\ 
th & \cellcolor{purple!35.11} 35.11 & \cellcolor{purple!32.81} 32.81 & \cellcolor{purple!37.20} 37.20 & \cellcolor{purple!44.00} 44.00 & \cellcolor{purple!34.51} 34.51 & \cellcolor{purple!49.30} 49.30 & \cellcolor{purple!29.09} 29.09 & \cellcolor{purple!37.57} 37.57 \\ 
tr & \cellcolor{purple!38.62} 38.62 & \cellcolor{purple!35.75} 35.75 & \cellcolor{purple!46.87} 46.87 & \cellcolor{purple!53.76} 53.76 & \cellcolor{purple!26.44} 26.44 & \cellcolor{purple!50.32} 50.32 & \cellcolor{purple!23.61} 23.61 & \cellcolor{purple!39.77} 39.77 \\ 
vi & \cellcolor{purple!28.38} 28.38 & \cellcolor{purple!31.82} 31.82 & \cellcolor{purple!25.44} 25.44 & \cellcolor{purple!40.76} 40.76 & \cellcolor{purple!23.61} 23.61 & \cellcolor{purple!44.28} 44.28 & \cellcolor{purple!31.12} 31.12 & \cellcolor{purple!32.63} 32.63 \\ 
zh & \cellcolor{purple!49.30} 49.30 & \cellcolor{purple!59.36} 59.36 & \cellcolor{purple!39.73} 39.73 & \cellcolor{purple!53.25} 53.25 & \cellcolor{purple!41.85} 41.85 & \cellcolor{purple!50.47} 50.47 & \cellcolor{purple!31.22} 31.22 & \cellcolor{purple!46.17} 46.17 \\
\midrule
en & \cellcolor{purple!55} 62.83 & \cellcolor{purple!60} 72.75 & \cellcolor{purple!55} 68.01 & \cellcolor{purple!60} 69.69 & \cellcolor{purple!60} 73.36 & \cellcolor{purple!60} 70.76 & \cellcolor{purple!40} 48.36 & \cellcolor{purple!55} 66.68 \\
\bottomrule
\end{tabular}

\label{tab:main_result_uabsa}
  \end{minipage}
	}
\vspace{-10pt}
\caption{Main results with mT5 on \textbf{TASD} and \textbf{UABSA} tasks in English-centric zero-shot transfer fashion.
} 
\label{tab:main_result_all}
\end{table*}

\begin{figure*}[ht]
\centering
\subfigure[Zero-shot performance using different source languages]{ 
\begin{minipage}{.762\linewidth}
\centering
\includegraphics[width=\linewidth]{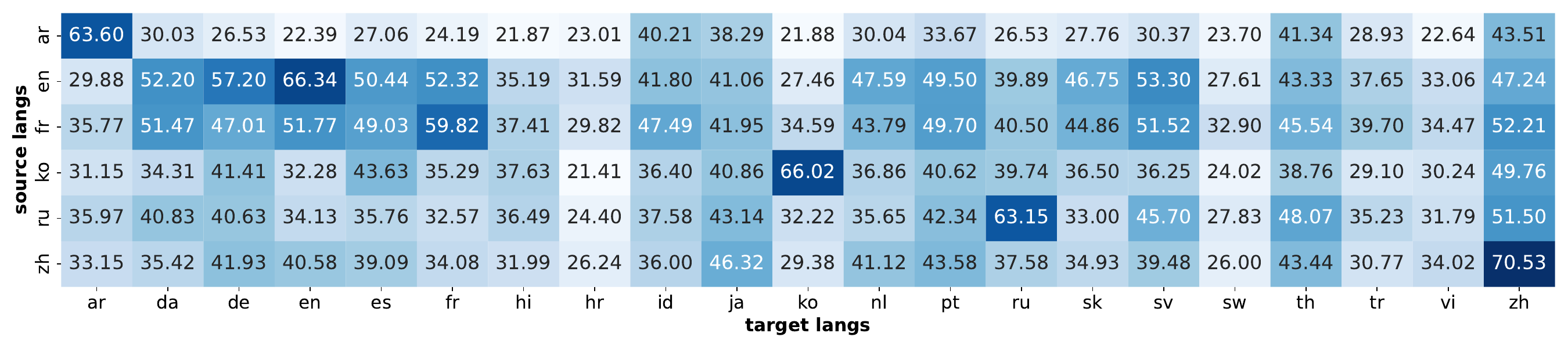}
\label{fig:non-en2target}
\end{minipage}}
\hfill
\subfigure[T-SNE of 6 langs]{ \begin{minipage}{.22\linewidth}
\centering
\includegraphics[width=\linewidth]{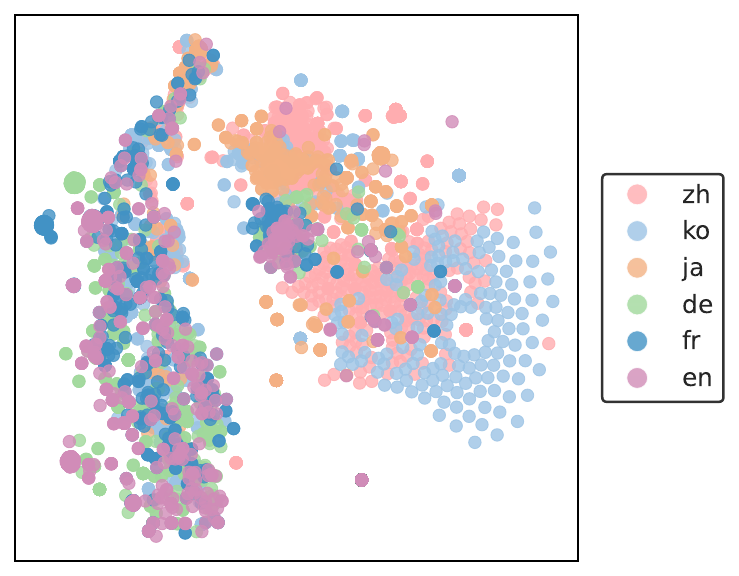}
\label{fig:tsne_res}
\end{minipage}}
\caption{Zero-shot fine-tuning for \textbf{TASD} task on Resteraunt domain. (a): Cross-lingual performance; (b): T-SNE visualizations of the selected source languages in the test set of Restaurant domain.}
\label{fig:combined}
\end{figure*}

\section{Experiments and Results}

\subsection{Experimental Setups}

\paragraph{Task Setups.}
As the main goal of the work is to provide a multilingual dataset with aspect triplets (aspect term, category, sentiment), we conduct the main experiments on the TASD task (triplet extraction). Also, our triplets contain subsets of sentiment pairs (aspect term, sentiment). Therefore, we also experiment on the UABSA (pairwise extraction) task and compare the results with TASD.

\paragraph{Model.}
We use a generation model to address the implicit aspects in the dataset, due to the compatibility of our dataset for generative tasks. Specifically, we finetune mT5-base \cite{DBLP:conf/naacl/XueCRKASBR21} on the M-ABSA dataset. The mT5-base is pretrained on a span-corruption variant of the masked language modeling objective, covering 101 languages, including all 21 languages in our dataset.

\paragraph{Evaluation Metric.}
We adopt Micro-F1 scores as the main evaluation metrics for all tasks. A prediction is correct if and only if all its predicted sentiment elements in the pair or triplet are correct. All the experimental results are reported using the average of 5 random seeds.

\subsection{Cross-Lingual Transfer Results}

\textbf{Main Results.}
Table \ref{tab:main_result_all} shows our main results of zero-shot cross-lingual transfer (i.e., training on English, inference on all target languages) by fine-tuning the mT5 model on the TASD and UABSA tasks. We observe the following phenomenon: 1) Compared to triplet extraction (cf.\ Table \ref{tab:main_result_tasd}), pairwise extraction (cf.\ Table \ref{tab:main_result_uabsa}) achieves better cross-lingual transfer performance across all domains. This indicates that introducing complex categories\footnote{For instance, as shown in Appendix \ref{sec:detailed_aspect_categories}, the Laptop dataset is divided into 114 fine-grained labels, representing the largest number of categories.} as sentiment elements presents a greater challenge for ABSA, as also shown by \citet{zhang-etal-2021-towards-generative}. 
2) On average, the performance of English exceeds that of the other languages.  
German (de) and Dutch (nl) achieve the highest scores of all languages other than English, with averages of 33.99\% and 31.85\% on the TASD task respectively. It is worth noting that German, Dutch, and English all belong to the West Germanic group of languages, which share structural and vocabulary similarities. In contrast, Swahili (sw), a Bantu language with relatively limited linguistic resources and distinct grammatical structures compared to more widely studied languages, emerges as the most challenging language, with an average performance of 15.96\%. This suggests the model has lower transferability when handling languages with fewer resources, particularly those from non-Indo-European language families.

\paragraph{Impact of Source Language.} We conduct additional experiments using five typologically diverse non-English source languages (zh, ko, ar, ru, fr), each with a different script, to examine their impact on TASD performance on the Restaurant dataset. In some cases, selecting a non-English source improves performance compared to English. As shown in Figure \ref{fig:non-en2target}, cross-lingual transfer benefits when the target language is semantically close to the source (e.g., Chinese-to-Japanese outperforms English-to-Japanese), likely due to biases in cross-lingual models, which are predominantly trained on English. Using non-English sources helps mitigate this bias. However, Arabic presents a challenge, possibly due to its significant linguistic and typological differences from the other languages.

We use the t-SNE algorithm to visualize aspect term representations in 2-dimensional Euclidean space for six languages from the Restaurant dataset in Figure \ref{fig:tsne_res}, revealing two clusters: (1) Chinese, Japanese, and Korean and (2) French, English, and German. These clusters align well with typological similarities in their respective language families. This visualization also aligns with the empirical results from \ref{fig:non-en2target} that transferring from linguistically similar languages can enhance performance. For instance, Chinese-to-Japanese and Korean yield improvements of 5.26\% and 1.92\% compared to Chinese-to-English, respectively. These findings highlight the importance of source language similarity in cross-lingual transfer.

In general, the results show that our M-ABSA dataset is a suitable multilingual dataset for evaluating the cross-lingual transfer abilities of models.

\subsection{LLMs Zero-Shot Results}
Additionally, we conduct extensive evaluations with the following open-weight LLMs on the dataset for the UABSA and TASD task, in a zero-shot prompting setting: Llama-3.1 8B \cite{llama31modelcard}, Mistral 7B \cite{jiang2023mistral7b}, Gemma-2 9B \cite{gemmateam2024gemma2improvingopen}, Qwen-2.5 7B \cite{qwen2.5}. 
To test the cross-lingual generative ability of multilingual pre-trained models without direct cross-lingual training data, we evaluate the zero-shot cross-lingual ABSA performance of LLMs based on prompt engineering. We show our prompts used in \S\ref{sec:prompts}.

\begin{figure}[htbp]
    \centering
    \begin{minipage}{0.46\textwidth}
        \centering
        \includegraphics[width=\textwidth]{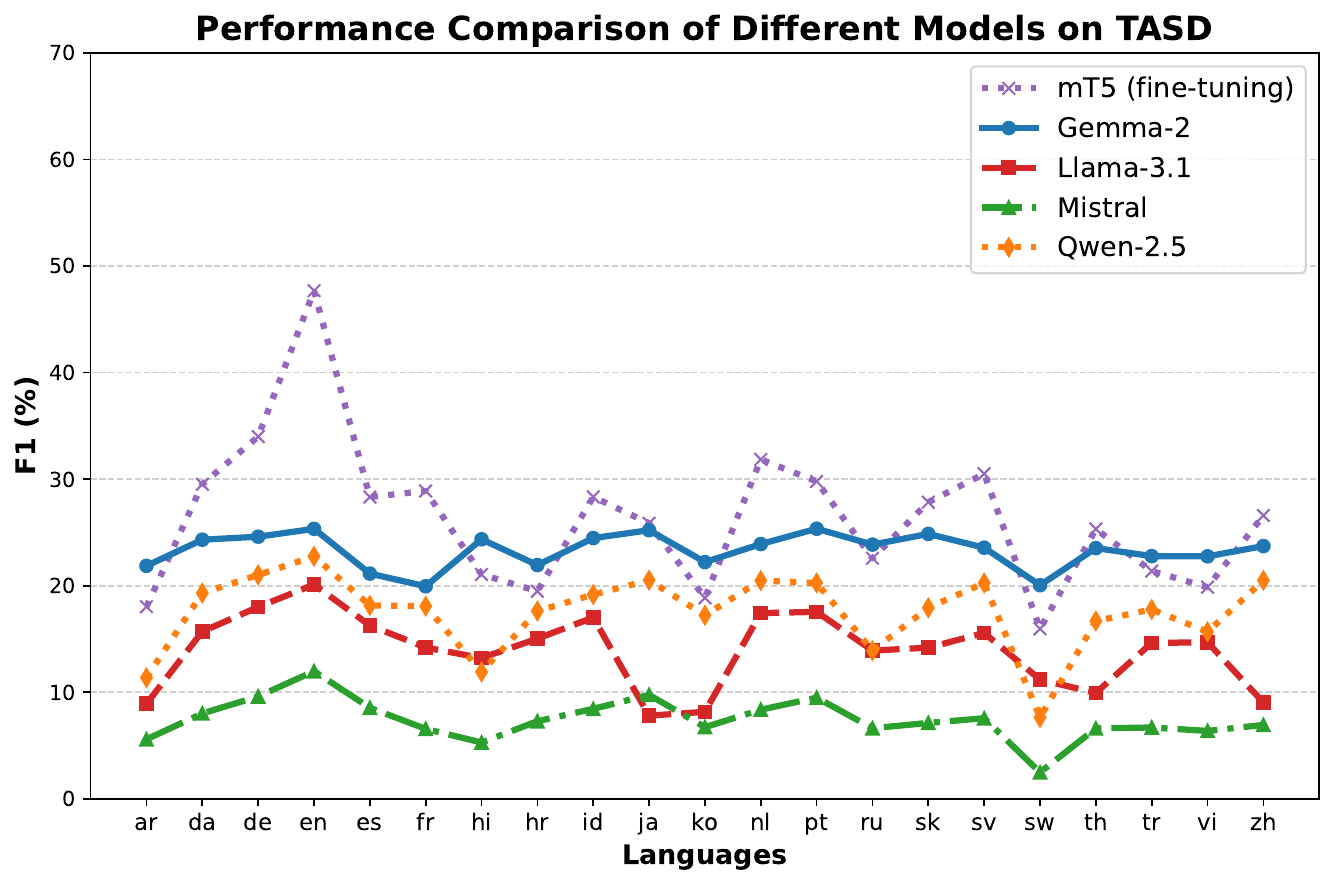}
    \end{minipage}
    \\
    \vspace{-3pt}
    \begin{minipage}{0.46\textwidth}
        \centering
        \includegraphics[width=\textwidth]{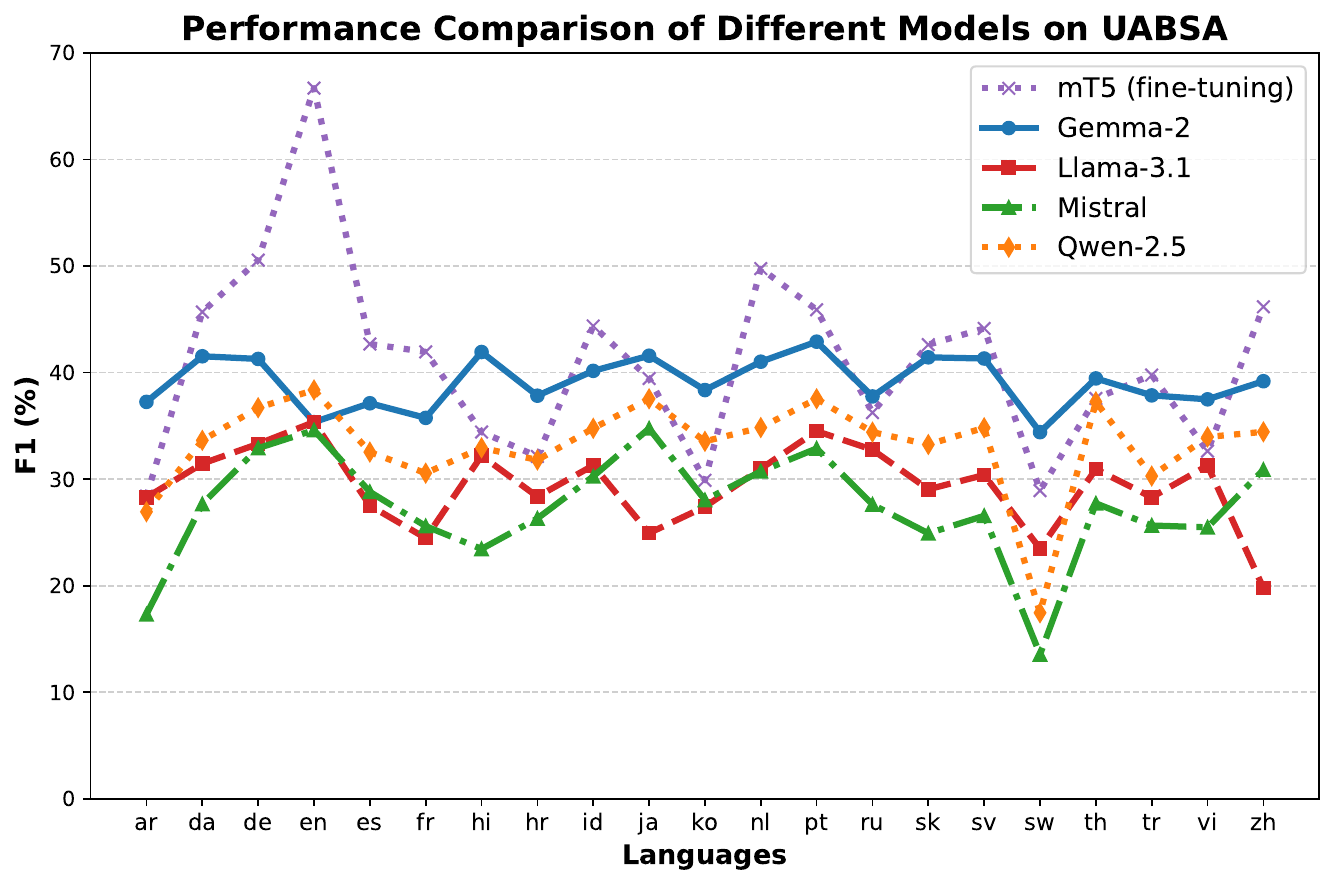}
    \end{minipage}
\caption{LLM zero-shot results vs mT5 fine-tuning results on TASD and UABSA. The results are averaged across 7 domains.}
\label{fig:llm}
\end{figure}

Figure \ref{fig:llm} shows the zero-shot inference results of the LLMs. We also include the previous mT5 results (fine-tuning on English, inference on target languages) as a comparison. We observe some fluctuations, but the best-performing LLM, Gemma-2, can achieve performance comparable to the zero-shot fine-tuning results of mT5. When it comes to the task type, we notice that the performance on TASD is much lower than on UABSA, as also observed in the fine-tuning results (cf. Table \ref{tab:main_result_all}). This further shows the challenge of the TASD task with triplet extraction and provides a potential for further finer-grained methods to achieve this task.

\subsection{Cross-Domain Results}\label{sec:cross_domain}
As M-ABSA contains multiple domains, it is interesting to investigate how a model performs when it is fine-tuned in one domain and tested in other domains. 
As the aspect categories of different domains are different, we only evaluate the UABSA task with the aspect term and sentiment tuples. We conduct cross-domain UABSA experiments on all seven domains in all languages with the results shown in Figure \ref{fig:cross-domain-all} in the Appendix. Figure \ref{fig:domain} shows an example of the results on the English dataset. 

\begin{figure}[htbp]
    \centering
        \includegraphics[width=1\linewidth]{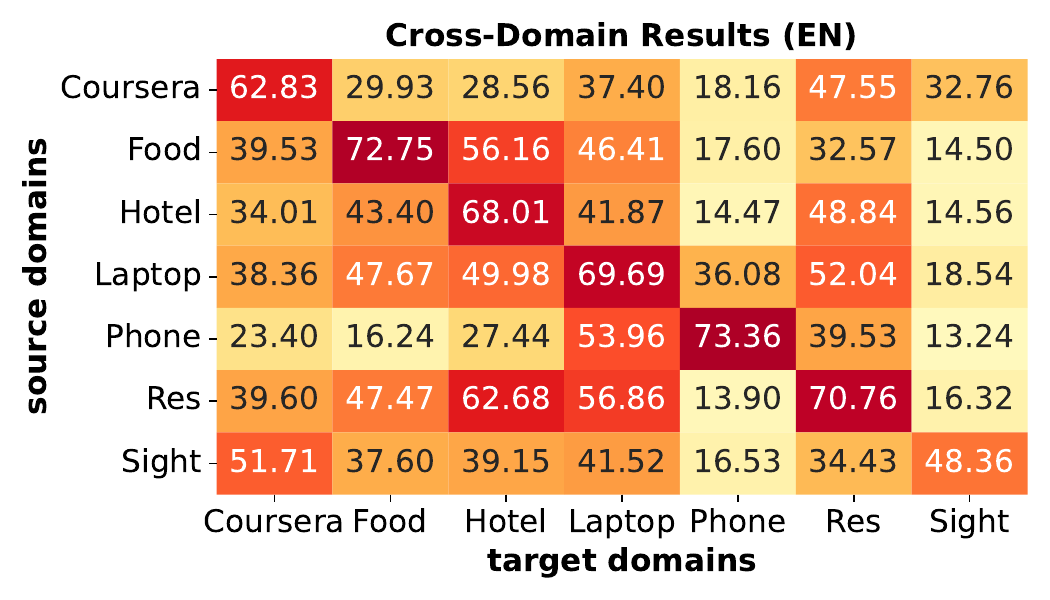}
\caption{An example of F1 scores of single-source cross-domain UABSA on M-ABSA (EN). Full results on all languages are shown in Figure \ref{fig:cross-domain-all} of Appendix.}
\label{fig:domain}
\end{figure}

We observe that transfer between similar domains exhibits positive transfer characteristics. For example, F1 scores from Restaurant to Hotel are second only to in-domain results, whereas transfer from Phone to Hotel yields the lowest performance. This can be explained by the fact that Restaurant and Hotel domains are both service-related and share common features. This suggests that additional data from similar domains can help mitigate data scarcity in the domain of interest.

\section{Conclusion}
We present M-ABSA, the most diverse multilingual parallel ABSA dataset to date, which includes 20 typologically different languages in addition to English and covers 7 different domains. 
Furthermore, we provide an efficient span-based annotation projection method together with minimal human review and revision, which can be easily adapted to work that extends existing monolingual datasets to multiple languages with high quality. Lastly, we conduct extensive experiments using M-ABSA including cross-lingual transfer, cross-domain transfer, and zero-shot prompting with LLMs. The results highlight the potential of our dataset for future multilingual ABSA research.

\section{Limitations}
We acknowledge that this work still has the following limitations:

\paragraph{Aspect-Based Sentiment Representation.} The triplet extraction task TASD extracts aspect, category, and sentiment triplets from reviews, but incorporating the opinion element from ABSA could provide a more comprehensive understanding. However, defining the opinion element across domains is challenging, as it often consists of multiple nouns or complex phrases that complicate machine translation. Future work could explore integrating opinion elements to enhance both linguistic richness and translation accuracy.

\paragraph{Specialized Cross-Lingual ABSA Models.} Our evaluation confirms that M-ABSA poses significant challenges, but we primarily rely on existing pipelines rather than designing a task-specific cross-lingual ABSA model. Future work should develop methods, such as recent knowledge and retrievel augmented approaches \cite{math10224273,zhang-etal-2024-mar}, tailored to these token-level entity recognition tasks’ linguistic diversity and structural complexities.

\paragraph{Translation Challenges for Long Phrases.} Certain languages struggle with accurately translating long opinion phrases, leading to potential semantic shifts. Addressing these translation inconsistencies, particularly for languages with distinct morphosyntactic structures, is an open challenge for improving multilingual ABSA.

\section{Ethical Considerations}
Since we have introduced a new multilingual triplet extraction ABSA dataset, we address some potential ethical considerations in this section.

\paragraph{Dataset Source.}
We select the English datasets Coursera \cite{chebolu-etal-2024-oats}, Food \cite{chebolu-etal-2024-oats}, Hotel \cite{chebolu-etal-2024-oats}, Laptop \cite{cai-etal-2021-aspect}, Phone \cite{zhou-etal-2023-unified-one}, Restaurant \cite{pontiki-etal-2016-semeval}, and Sight \cite{wang-etal-2023-sight}, and extend the multilingual sentiment analysis dataset using machine translation and manual verification. 
We ensure that this new dataset is intended solely for research purposes and should not be used for commercial purposes. Furthermore, the dataset construction strictly adheres to the intellectual property and privacy protection requirements of the original authors and is freely available for download from their official website.

\paragraph{Data Annotation.}  Before the annotation process, we fully informed the annotators about the nature and goals of the task and obtained their informed consent. 
All annotators are project partners and voluntarily joined as contributors. Additionally, all annotators provided explicit consent for the use of the collected data. 
During human evaluation of translation quality, we recruited external human participants from Prolific. All human evaluators were paid properly at an hourly rate of £9. 

\paragraph{Risk Concerns.} Constructing multilingual datasets using advanced machine translation engines does not raise any ethical concerns, as the process involves the automatic translation of publicly available textual data without any human involvement in altering the original meaning. Machine translation tools, such as those provided by commercial services (e.g., Google Translate), operate on publicly accessible models trained on large corpora, ensuring that the translation process is fair and does not introduce any bias or manipulation. Moreover, the dataset does not contain personally identifiable information or sensitive data, and the translation process does not involve any potentially ethically risky manual annotations.

\paragraph{Use of AI Assistants.} The authors acknowledge the use of ChatGPT solely for correcting grammatical errors, enhancing the coherence of the final manuscript, and providing assistance with coding.

\section*{Acknowledgments}
We thank the members of the MaiNLP lab from LMU Munich for their constructive feedback. 
CW and YX are supported in part by the Guangdong Basic and Applied Basic Research Foundation under Grant 2023A1515011370, the National Natural Science Foundation of China (32371114), the Characteristic Innovation Projects of Guangdong Colleges and Universities (No. 2018KTSCX049), and the Guangdong Provincial Key Laboratory (No. 2023B1212060076). 
BP is supported by ERC Consolidator Grant DIALECT (101043235).

\bibliography{custom}

\appendix

\section{Annotation Process} 
\label{sec:annotation_process}

As we describe in \S\ref{sec:collection}, for the Sight dataset, we need to manually annotate the aspect terms. The entire dataset construction process consists of the following steps: 

\paragraph{Data Cleaning.} Sentences with fewer than 6 valid tokens or consisting solely of symbols are filtered out. Duplicate sentences are then removed.

\paragraph{Sentence Language Purity Check.} We use the LangID\footnote{\url{https://github.com/cisnlp/GlotLID}} \cite{kargaran-etal-2023-glotlid} toolkit to detect and remove non-English sentences. Specifically, the tool detects the top two candidate languages, and we calculate the probability difference between the two. If the difference exceeds 0.6, the language with the highest probability is assigned as the sentence's language. We then randomly sample approximately 2,000 sentences from the original dataset to construct the entire dataset.

\paragraph{Topic Modeling.} For constructing the aspect categories, we apply LDA-based topic models to generate the clusters for the categories based on the input sentences, with manual checks on the generated topics, following this topic curation process \cite{resnik2024topic}. Specifically, we initialize a granularity of 10 topics and then manually code the 10 topics, which are then reviewed by another annotator with domain experts.

\paragraph{Aspect Categories Coding.} Once we get the topics, we input the sentence sets from each of the five categories (with the top 100 sentences per topic) into GPT-3.5, using 50-shot samples for each input, to let it code the topics. By using prompt engineering, we generate the aspect category mentioned in each sentence and classified them. Finally, we perform manual analysis and unify the categories into five distinct classes. For specific prompts, refer to Figure \ref{fig:llm_classification}.

\paragraph{Multi-round Annotation.} As described in Section \ref{sec:annotation}, 6 annotators are invited for annotations and follow strict quality control procedures to ensure the quality of the annotations. Each sentence is annotated by three annotators: Annotator A, Annotator B, and Reviewer C. Annotators A and B check and modify each other's annotations, while Reviewer C resolves any disagreements between the annotations of A and B. Initially, Annotator A labels the entire sub-dataset, after which Annotator B reviews and makes corrections, followed by Reviewer C, who checks and balances the annotations of both A and B. During the annotation process, any newly emerged discrepancies are resolved through consultation with our NLP expert (Reviewer C). These experts are added to the annotation team for future reviews. 

\paragraph{Annotation Consistency.} As detailed in Section \ref{sec:annotation}, for the annotated dataset obtained from Step 6, we use the F1 score to evaluate the consistency of the annotations throughout the entire process.

\begin{figure*}[htbp] 
\begin{tcolorbox}
\small{
\textbf{[Background]}: 
These comments are collected from the Massachusetts Institute of
Technology OpenCourseWare (MIT OCW) YouTube channel and aim to gather
feedback from online students about the lecture content of a mathematics course. The
course includes transcriptions of the math lectures, and we have categorized the
comments into five themes using the LDA topic model, ranking the sentences based
on their relevance to each theme from highest to lowest. \\

\vspace{1pt}
\textbf{[Task]}: 
Below are representative examples for each theme, please classify the
comments based on these examples. The comments mention the teaching methods
employed by the instructor, which include, but are not limited to, the use of examples,
applications, problem-solving, proofs, visualizations, explanations, and analogies.\\

\vspace{1pt}
\textbf{[Case]}: 
<Sentence>
}
\end{tcolorbox}
\caption{The sentence classification process with GPT-3.5.} 
\label{fig:llm_classification} 
\end{figure*}

\section{Annotation Guidelines} 
\label{sec:annotation_guideline}

Following the annotation guidelines of SemEval 2016 \cite{pontiki-etal-2016-semeval}, we have developed the annotation guidelines for the three fundamental sentiment elements of TASD and their corresponding outcomes for the Sight dataset. The annotators are project partners and experts in NLP. They are required to mark the texts according to the following guideline. 

\paragraph{Aspect Categories.} The aspect categories are defined and coded according to the step demonstrated in \S\ref{sec:annotation_process}.

\begin{itemize}[leftmargin=*]
\item Teaching\_Setup: Comments describing or mentioning the teaching setup of the lecture. The teaching setup includes aspects related to the blackboard, chalk, microphone or audio, volume, and camera or camera-related aspects (e.g., angle).

\item  Course\_General\_Feedback: Comments describing or mentioning overall feedback on the course, lecture, or video (experience, opinions).

\item Instructor: Comments expressing evaluations of the instructor (speaker).

\item Mathematical\_Related\_Concept: Comments describing or mentioning personal feelings or evaluations related to examples, concepts, explanations, or proofs within mathematical subjects.

\item Other: Aspects with sentiment that do not fall under the categories listed above.

\end{itemize}
\paragraph{Aspect Terms.}
The aspect can be a specific entity, a common noun, or a multi-word term, indicating the opinion target in a sentence. Moreover, to provide more fine-grained information, we include three additional rules:
\begin{itemize}[leftmargin=*]
\item Top-priority in labeling fine-grained aspects. For mathematical concept elements, such as in the example ``Why should \(a_1\) and \(a_2\) both be perpendicular to the vector \(b\)?'', each element—like \(a_1\), \(a_2\), and vector \(b\)—should be annotated individually, rather than annotating them as a whole. For mathematical equations, such as in the example ``Can it be solved when you put \( y = a \cos(\theta) \)?'', the entire equation should be annotated as a whole, as in the case of \( y = a \cos(\theta) \).

\item Priority order of temporal logic. For sentences where emotional changes occur based on temporal order, we assign the emotion according to the most recent time point.

\end{itemize}
\paragraph{Sentiment Polarity.}
The sentiment polarity belongs to one of the sentiment label sets: \{POS, NEU, NEG\}, which stand for positive, neutral, and negative, respectively.
\begin{itemize}[leftmargin=*]
\item NEU: Indicates a slight positive or slightly negative emotional tone towards a specific aspect, rather than describing objective facts. For example, a comment expressing personal confusion about mathematical concepts or reasoning processes.

\item NEG: Indicates a strong negative emotional tone towards a specific aspect. For example, a comment expressing dissatisfaction with the course experience.

\item POS: Indicates a strong positive emotional tone towards a specific aspect. For example, a comment expressing admiration for the speaker.

\end{itemize}

\section{Manual Proofreading}
\label{sec:manual_proofreading}
We first perform translation by inserting special markers into the sentence, and then correct two types of errors: aspect translation errors and aspect omissions. Specifically, we use the original sentence without special markers for translation, and then identify the corresponding aspects to replace them at the appropriate positions.

\begin{figure*}[htbp] 
\centering 
\includegraphics[width=0.95\linewidth]{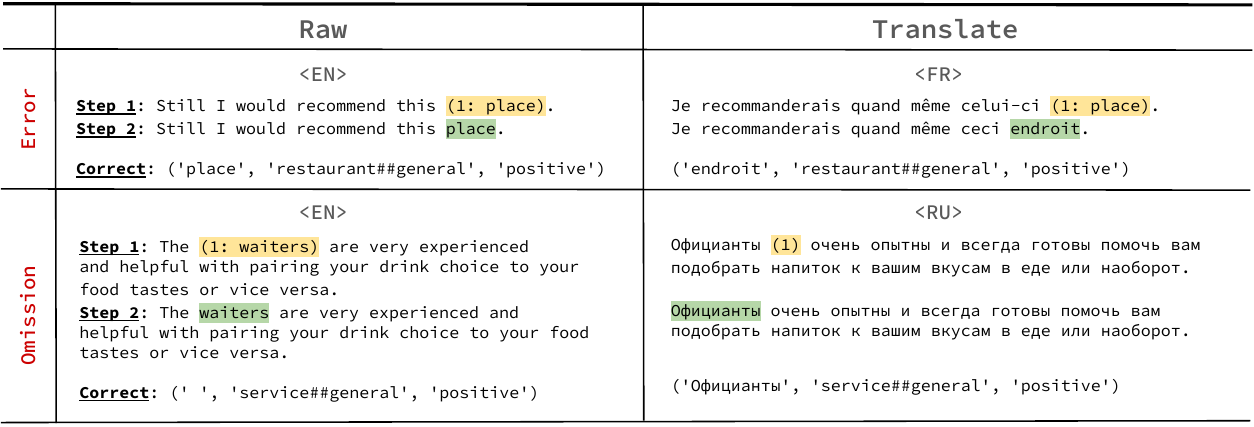} 
\caption{The correcting process of the current multilingual ABSA dataset.} 
\label{fig:correct_process} 
\end{figure*}

Figure \ref{fig:correct_process} presents two error correction examples. In the case of translation errors (where the aspect term remains unchanged before and after translation), as shown in Step 1, the term ``place'' is not correctly translated to its French counterpart ``endroit'' due to special markers. In Step 2, we retranslate the original sentence without special markers, and by consulting the Wikipedia knowledge base, we replace the misaligned ``place'' with the correct ``endroit''.

For omission translations, as shown in Step 1, the aspect term ``waiters'' is misaligned during the translation into Russian, due to special markers causing the term to fall outside the marker range, leading to the loss of alignment information. In Step 2, we repeat the same process and replace the missing aspect term with the correct ``\foreignlanguage{russian}{Официанты}''.

\section{Experiment Details}
\label{sec:experiment_details}
During the experiments, we used the \texttt{transformers} \cite{wolf-etal-2020-transformers} and \texttt{pytorch} \cite{NEURIPS2019_9015} library for training the models. 
Figure \ref{tab:hyperparameters} presents the hyperparameter settings of the mT5-base model used in the experiments. For all datasets, except for the Hotel dataset where training is set to 5 epochs, the number of epochs is set to 30. The learning rate and batch size are set to 3e-4 and 16, respectively, with a maximum of 2500 training steps, and the best model is selected based on the performance during the final 500 steps. Additionally, the model's dropout rate, Adam epsilon, and warmup factor are set to 0.1, 1e-6, and 0.1, respectively.

\begin{table}[htbp]
\small
\centering
\renewcommand\arraystretch{1}
\begin{tabular}{lc}
\toprule
\textbf{Parameter}          & \textbf{Value}     \\ \midrule
Epoch                       & [5, 30]           \\ 
Batch size                  & 16            \\ 
Learning rate       & 3e-4              \\ 
Hidden size                 & 768                \\ 
Dropout rate                & 0.1                \\ 
Max steps                   & [2000, 2500]        \\ 
Adam epsilon                & 1e-6               \\ 
Warm factor                 & 0.1                \\ \bottomrule
\end{tabular}
\caption{Hyper-parameter settings for the mT5-base model.}
\label{tab:hyperparameters}
\end{table}

\section{Prompts for LLMs}
\label{sec:prompts}
We show both our prompts for the UABSA and TASD tasks in Figure \ref{fig:absa-prompt}. The prompts are developed based on previous work \cite{wu2024evaluating} on evaluating LLMs for UABSA tasks.

\section{Details of Automatic Translation Quality Evaluation}
\label{sec:details_quality}
\label{sec:translation quality details}
To measure the chrF++ and BLEU scores, we use the \texttt{sacrebleu} package.\footnote{\url{https://github.com/mjpost/sacrebleu}}
To measure the BERTScore, we use the default model for English in \texttt{bert\_score} package, i.e., \texttt{roberta-large}.\footnote{\url{https://huggingface.co/FacebookAI/roberta-large}}
To measure the SBERT scores, we use the \texttt{sentence-transformer} package and the \texttt{all-MiniLM-L6-v2} model.\footnote{\url{https://github.com/UKPLab/sentence-transformers}}\footnote{\url{https://huggingface.co/sentence-transformers/all-MiniLM-L6-v2}} 

We also present the evaluation of translation quality in each domain separately as opposed to the aggregated evaluation conducted in \S\ref{tab:main translation quality}.
In addition to our actual dataset, we also show a naive random baseline -- constructed by (1) randomly pairing up the new translation ($\text{translation}_{\text{w/o marker}}$) and the original translation ($\text{translation}_{\text{w/ marker}}$) in the target language for measuring \emph{consistency} (Acc and chrF++) and
(2) randomly pairing up back-translation and original English data for measuring \emph{faithfulness} (BERTScore, SBERT, and BLEU). Table \ref{tab:baseline_evaluation-results_food}-\ref{tab:baseline_evaluation-results_phone} presents the results of each domain.

We observe that the scores of our actual dataset are much higher than the random baseline for most metrics, indicating good translation quality across domains. One exception is the BERTScore, where the random baselines also obtain relatively high scores. This is because the BERT model tends to assign high similarity even to the random word/sentence pairs \citep{ethayarajh-2019-contextual,zhao-etal-2021-inducing,liu-etal-2025-transliterations}. Its counterpart, SBERT, is further fine-tuned in a contrastive way so that it can better differentiate matched pairs from random pairs. Therefore, we observe very low SBERT scores for the random baseline but high scores for our dataset, suggesting the translation in different languages is faithful in keeping the meaning of the original English data. To sum up, the translation quality is good across languages and domains.

\section{Human Evaluation of Translation Quality}
\label{sec:human-eval-appendix}

To assess translation quality beyond automatic metrics, we conducted a human evaluation of the sampled translations along five dimensions:

\begin{itemize}[leftmargin=*]
\item \textbf{Grammar:} Is the sentence grammatically well-formed?
\item \textbf{Fluency:} Does the sentence sound natural and idiomatic in the target language?
\item \textbf{Adequacy:} Does the translation preserve the meaning of the original English sentence?
\item \textbf{Code-Switching:} Does the translation avoid unnecessary mixing of English or other foreign words?
\item \textbf{Aspect Term Translation:} Is the aspect term (originally marked in brackets) correctly and accurately translated?
\end{itemize}

Each dimension was rated on a 5-point Likert scale, from 1 (poor) to 5 (excellent).

Human annotators were recruited via Prolific\footnote{\url{https://www.prolific.com/}}. To ensure quality, all annotators were required to be native speakers of the target language and have high proficiency in English. Annotation instructions were provided in English. We included explanations for each evaluation dimension.

\begin{figure}[htbp]
\centering
\includegraphics[width=1\columnwidth]{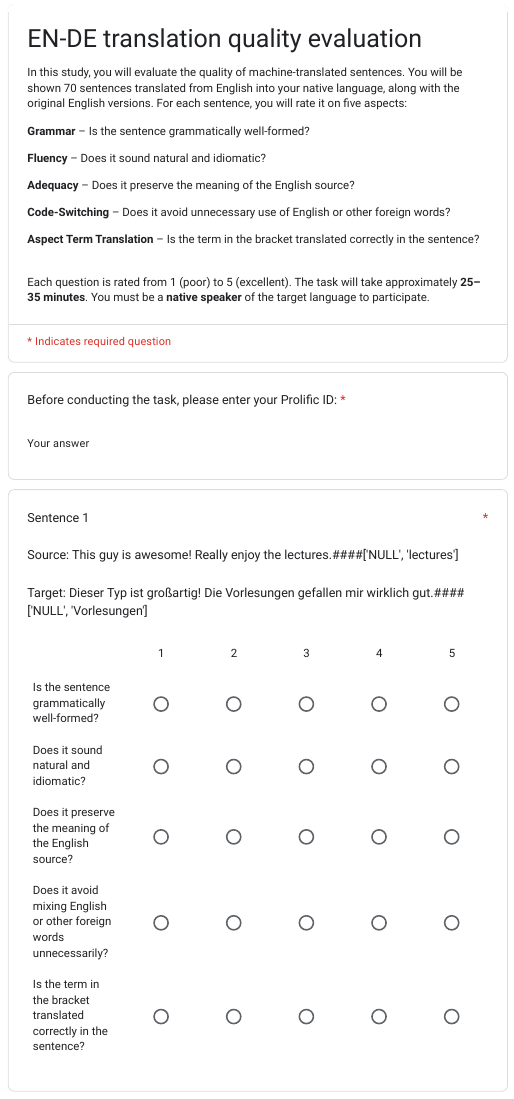}
\caption{Screenshot of the annotation interface showing instructions and an example evaluation.}
\label{fig:annotation-eval}
\end{figure}

Figure~\ref{fig:annotation-eval} shows a screenshot of the annotation interface, including the evaluation instructions and the rating interface for a sample sentence.

\section{Detailed Aspect Categories}
\label{sec:detailed_aspect_categories}
We present in this section the detailed aspect categories for each subset of the dataset, in addition to the ones we self-curate and present in the previous section (\S\ref{sec:annotation_process}).

\begin{tcolorbox}
    [colframe=gray!100, sharp corners, leftrule={3pt}, rightrule={0pt}, toprule={0pt}, bottomrule={0pt}, left={2pt}, right={2pt}, top={3pt}, bottom={3pt}, breakable]
    \small
\textbf{Coursera} =  ['assignments comprehensiveness', 'assignments quality', 'assignments quantity', 'assignments relatability', 'assignments workload', 'course comprehensiveness', 'course general', 'course quality', 'course relatability', 'course value', 'course workload', 'faculty comprehensiveness', 'faculty general', 'faculty relatability', 'faculty response', 'faculty value', 'grades general', 'material comprehensiveness', 'material quality', 'material quantity', 'material relatability', 'material workload', 'polarity negative', 'polarity neutral', 'polarity positive', 'presentation comprehensiveness', 'presentation quality', 'presentation quantity', 'presentation relatability', 'presentation workload']
\end{tcolorbox}

\begin{tcolorbox}
    [colframe=gray!100, sharp corners, leftrule={3pt}, rightrule={0pt}, toprule={0pt}, bottomrule={0pt}, left={2pt}, right={2pt}, top={3pt}, bottom={3pt}, breakable]
    \small
\textbf{Food} =  ['amazon availability', 'amazon prices', 'food general', 'food prices', 'food quality', 'food recommendation', 'food style\_options', 'shipment delivery', 'shipment prices', 'shipment quality']
\end{tcolorbox}

\begin{tcolorbox}
    [colframe=gray!100, sharp corners, leftrule={3pt}, rightrule={0pt}, toprule={0pt}, bottomrule={0pt}, left={2pt}, right={2pt}, top={3pt}, bottom={3pt}, breakable]
    \small
\textbf{Hotel} =  ['facilities cleanliness', 'facilities comfort', 'facilities design\_features', 'facilities general', 'facilities miscellaneous', 'facilities prices', 'facilities quality', 'food\_drinks miscellaneous', 'food\_drinks prices', 'food\_drinks quality', 'food\_drinks style\_options', 'hotel cleanliness', 'hotel comfort', 'hotel design\_features', 'hotel general', 'hotel miscellaneous', 'hotel prices', 'hotel quality', 'location general', 'polarity positive', 'room\_amenities cleanliness', 'room\_amenities comfort', 'room\_amenities design\_features', 'room\_amenities general', 'room\_amenities prices', 'room\_amenities quality', 'rooms cleanliness', 'rooms comfort', 'rooms design\_features', 'rooms general', 'rooms miscellaneous', 'rooms prices', 'rooms quality', 'service general']
\end{tcolorbox}

\begin{tcolorbox}
    [colframe=gray!100, sharp corners, leftrule={3pt}, rightrule={0pt}, toprule={0pt}, bottomrule={0pt}, left={2pt}, right={2pt}, top={3pt}, bottom={3pt}, breakable]
    \small
\textbf{Laptop} = ['battery\#design\_features', 'battery\#general', 'battery\#operation\_performance', 'battery\#quality', 'company\#design\_features', 'company\#general', 'company\#operation\_performance', 'company\#price', 'company\#quality', 'cpu\#design\_features', 'cpu\#general', 'cpu\#operation\_performance', 'cpu\#price', 'display\#design\_features', 'display\#general', 'display\#operation\_performance', 'display\#price', 'display\#quality', 'display\#usability', 'fans\&cooling\#general', 'fans\&cooling\#operation\_performance', 'fans\&cooling\#quality', 'graphics\#design\_features', 'graphics\#general', 'graphics\#operation\_performance', 'graphics\#usability', 'hardware\#design\_features', 'hardware\#general', 'hardware\#operation\_performance', 'hardware\#quality', 'hardware\#usability', 'hard\_disc\#design\_features', 'hard\_disc\#general', 'hard\_disc\#miscellaneous', 'hard\_disc\#operation\_performance', 'hard\_disc\#price', 'hard\_disc\#quality', 'hard\_disc\#usability', 'keyboard\#design\_features', 'keyboard\#general', 'keyboard\#miscellaneous', 'keyboard\#operation\_performance', 'keyboard\#portability', 'keyboard\#price', 'keyboard\#quality', 'keyboard\#usability', 'laptop\#connectivity', 'laptop\#design\_features', 'laptop\#general', 'laptop\#miscellaneous', 'laptop\#operation\_performance', 'laptop\#portability', 'laptop\#price', 'laptop\#quality', 'laptop\#usability', 'memory\#design\_features', 'memory\#general', 'memory\#operation\_performance', 'memory\#quality', 'memory\#usability', 'motherboard\#quality', 'mouse\#design\_features', 'mouse\#general', 'mouse\#usability', 'multimedia\_devices\#connectivity', 'multimedia\_devices\#design\_features', 'multimedia\_devices\#general', 'multimedia\_devices\#operation\_performance', 'multimedia\_devices\#quality', 'optical\_drives\#general', 'optical\_drives\#usability', 'os\#design\_features', 'os\#general', 'os\#miscellaneous', 'os\#operation\_performance', 'os\#price', 'os\#quality', 'os\#usability', 'out\_of\_scope\#design\_features', 'out\_of\_scope\#general', 'out\_of\_scope\#operation\_performance', 'out\_of\_scope\#usability', 'ports\#connectivity', 'ports\#design\_features', 'ports\#general', 'ports\#operation\_performance', 'ports\#portability', 'ports\#quality', 'ports\#usability', 'power\_supply\#design\_features', 'power\_supply\#general', 'power\_supply\#operation\_performance', 'power\_supply\#quality', 'shipping\#general', 'shipping\#operation\_performance', 'shipping\#quality', 'software\#design\_features', 'software\#general', 'software\#operation\_performance', 'software\#portability', 'software\#price', 'software\#quality', 'software\#usability', 'support\#general', 'support\#operation\_performance', 'support\#price', 'support\#quality', 'warranty\#general']
\end{tcolorbox}

\begin{tcolorbox}
    [colframe=gray!100, sharp corners, leftrule={3pt}, rightrule={0pt}, toprule={0pt}, bottomrule={0pt}, left={2pt}, right={2pt}, top={3pt}, bottom={3pt}, breakable]
    \small
\textbf{Phone} =  ['After-sales Service\#Exchange/Warranty/Return', 'Appearance Design\#Aesthetics General', 'Appearance Design\#Color', 'Appearance Design\#Exterior Design Material', 'Appearance Design\#Fuselage Size', 'Appearance Design\#Grip Feeling', 'Appearance Design\#Thickness', 'Appearance Design\#Weight', 'Appearance Design\#Workmanship and Texture', 'Audio/Sound\#Tone quality', 'Audio/Sound\#Volume and Speaker', 'Battery/Longevity\#Battery Capacity', 'Battery/Longevity\#Battery Life', 'Battery/Longevity\#Charging Method', 'Battery/Longevity\#Charging Speed', 'Battery/Longevity\#General', 'Battery/Longevity\#Power Consumption Speed', 'Battery/Longevity\#Standby Time', 'Branding/Marketing\#Promotional Giveaways', 'Buyer Attitude\#Loyalty', 'Buyer Attitude\#Recommendable', 'Buyer Attitude\#Repurchase and Churn Tendency', 'Buyer Attitude\#Shopping Experiences', 'Buyer Attitude\#Shopping Willingness', 'Camera\#Fill light', 'Camera\#Front Camera', 'Camera\#General', 'Camera\#Rear Camera', 'Ease of Use\#Audience Groups', 'Ease of Use\#Easy to Use', 'Intelligent Assistant\#Intelligent Assistant General', 'Key Design\#General', 'Logistics\#Lost and Damaged', 'Logistics\#Shipping Fee', 'Logistics\#Speed', 'Logistics\#general', 'Overall\#Overall', 'Performance\#General', 'Performance\#Heat Generation', 'Performance\#Running Speed', 'Price\#Price', 'Price\#Value for Money', 'Product Accessories\#Cell Phone Film', 'Product Accessories\#Charger', 'Product Accessories\#Charging Cable', 'Product Accessories\#Headphones', 'Product Accessories\#Phone Cases', 'Product Configuration\#CPU', 'Product Configuration\#Memory', 'Product Configuration\#Operating Memory', 'Product Packaging\#Completeness of Accessories', 'Product Packaging\#General', 'Product Packaging\#Instruction Manual', 'Product Packaging\#Packaging Grade', 'Product Packaging\#Packaging Materials', 'Product Quality\#Cleanliness', 'Product Quality\#Dustproof', 'Product Quality\#Fall Protection', 'Product Quality\#General', 'Product Quality\#Genuine Product', 'Product Quality\#Water Resistant', 'Screen\#Clarity', 'Screen\#General', 'Screen\#Size', 'Security\#Screen Unlock', 'Seller Service\#Attitude', 'Seller Service\#Inventory', 'Seller Service\#Seller Expertise', 'Seller Service\#Shipping', 'Seller Service\#Timeliness of Seller Service', 'Shooting Functions\#General', 'Shooting Functions\#Pixel', 'Signal\#Call Quality', 'Signal\#Signal General', 'Signal\#Signal of Mobile Network', 'Signal\#Wifi Signal', 'Smart Connect\#Bluetooth Connection', 'Smart Connect\#Positioning and GPS', 'System\#Application', 'System\#Lock Screen Design', 'System\#NFC', 'System\#Operation Smoothness', 'System\#Software Compatibility', 'System\#System General', 'System\#System Upgrade', 'System\#UI Interface Aesthetics']
\end{tcolorbox}

\begin{tcolorbox}
    [colframe=gray!100, sharp corners, leftrule={3pt}, rightrule={0pt}, toprule={0pt}, bottomrule={0pt}, left={2pt}, right={2pt}, top={3pt}, bottom={3pt}, breakable]
    \small
\textbf{Restaurant} =  ['ambience general', 'drinks prices', 'drinks quality', 'drinks style\_options', 'food general', 'food prices', 'food quality', 'food style\_options', 'location general', 'restaurant general', 'restaurant miscellaneous', 'restaurant prices', 'service general']
\end{tcolorbox}

\begin{tcolorbox}
    [colframe=gray!100, sharp corners, leftrule={3pt}, rightrule={0pt}, toprule={0pt}, bottomrule={0pt}, left={2pt}, right={2pt}, top={3pt}, bottom={3pt}, breakable]
    \small
\textbf{Sight} = ['Course\_General\_Feedback', 'Instructor', 'Mathematical\_Related\_Concept', 'Other', 'Teaching\_Setup']
\end{tcolorbox}

\section{Additional Experimental Results}
\label{sec:additional_experimental_results}

\begin{figure*}[htbp]
\centering
  \begin{minipage}{0.3\textwidth}
 \centering
 \includegraphics[width=\linewidth]{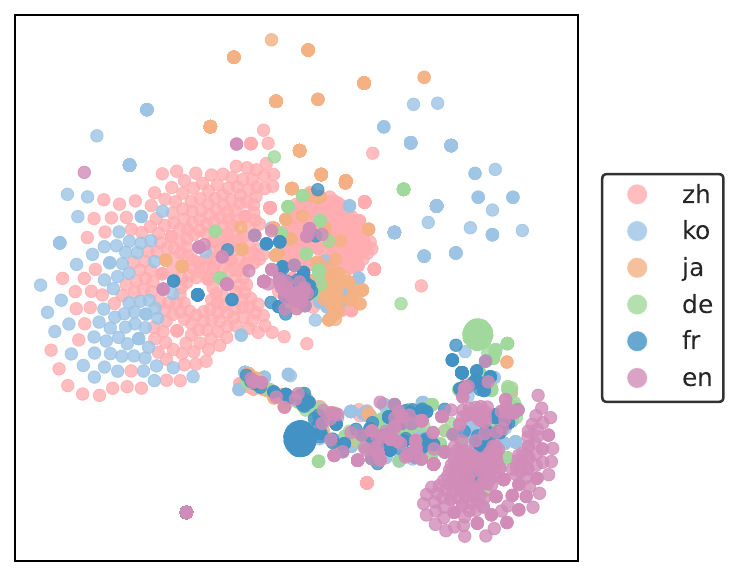}
  \end{minipage}%
  \begin{minipage}{0.3\textwidth}
 \centering
 \includegraphics[width=\linewidth]{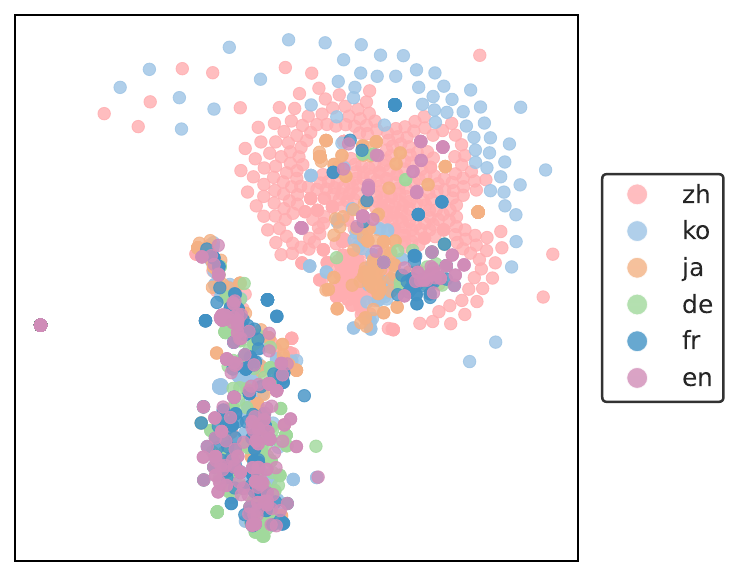}
  \end{minipage}%
  \begin{minipage}{0.3\textwidth}
 \centering
 \includegraphics[width=\linewidth]{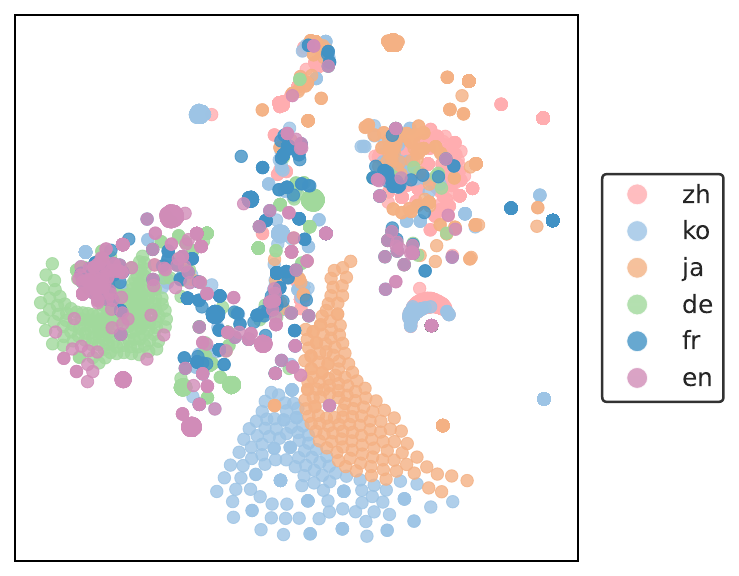}
  \end{minipage}

  \vspace{0.5cm}  

  \begin{minipage}{0.3\textwidth}
 \centering
 \includegraphics[width=\linewidth]{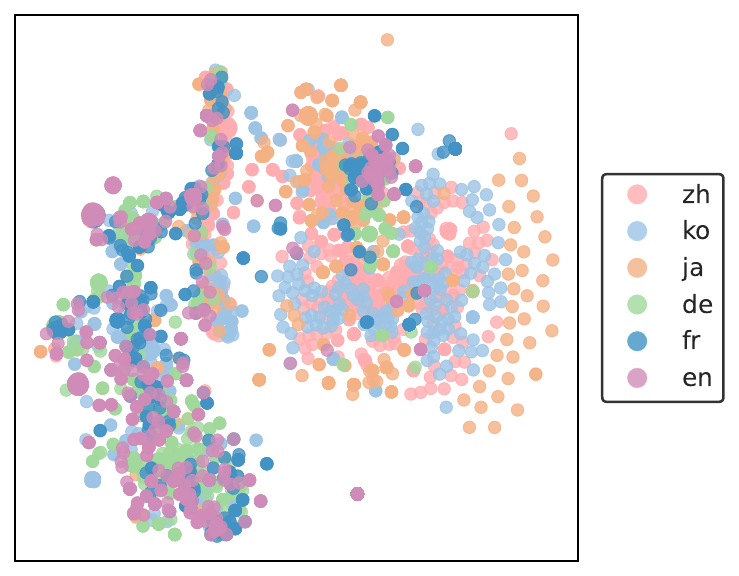}
  \end{minipage}%
  \begin{minipage}{0.3\textwidth}
 \centering
 \includegraphics[width=\linewidth]{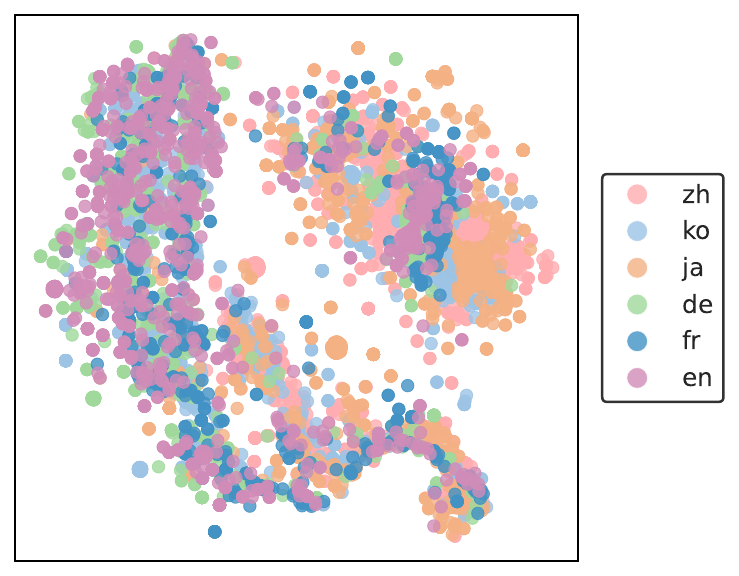}
  \end{minipage}%
  \begin{minipage}{0.3\textwidth}
 \centering
 \includegraphics[width=\linewidth]{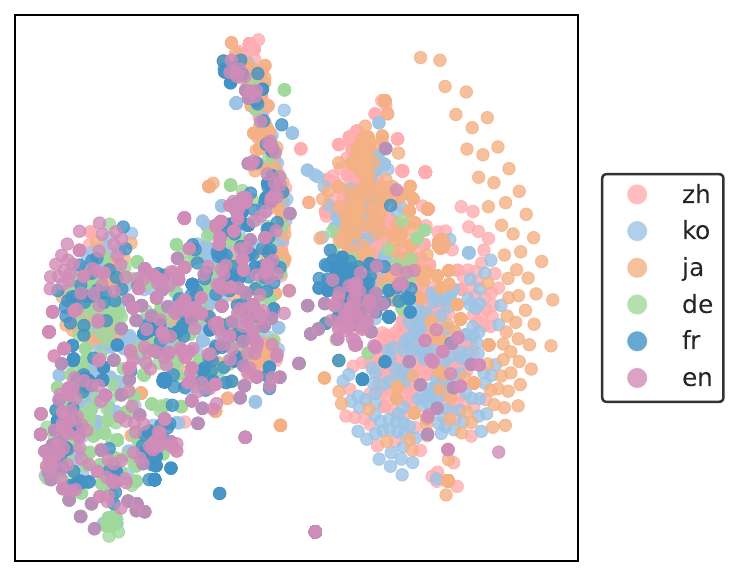}
  \end{minipage}
  
  \caption{T-SNE Visualizations of the Test Set for M-ABSA under all other domains. Top: Coursera, Food, Hotel; Bottom: Laptop, Phone, Sight.}
  \label{fig:test-visual-6}
\end{figure*}

Figure \ref{fig:test-visual-6} shows the T-SNE visualization of the test set for M-ABSA under all other domains apart from the restaurant we showed in the main paper. 

In Table \ref{tab:non-en-7}-\ref{tab:non-en-14}, we present cross-lingual results obtained with non-English source languages.

In Table \ref{tab:main_result_gemma}-\ref{tab:main_result_qwen}, we present the detailed results of LLM evaluation.

\begin{figure*}[htbp]
\centering
  \begin{minipage}{0.3\textwidth}
 \centering
 \includegraphics[width=\linewidth]{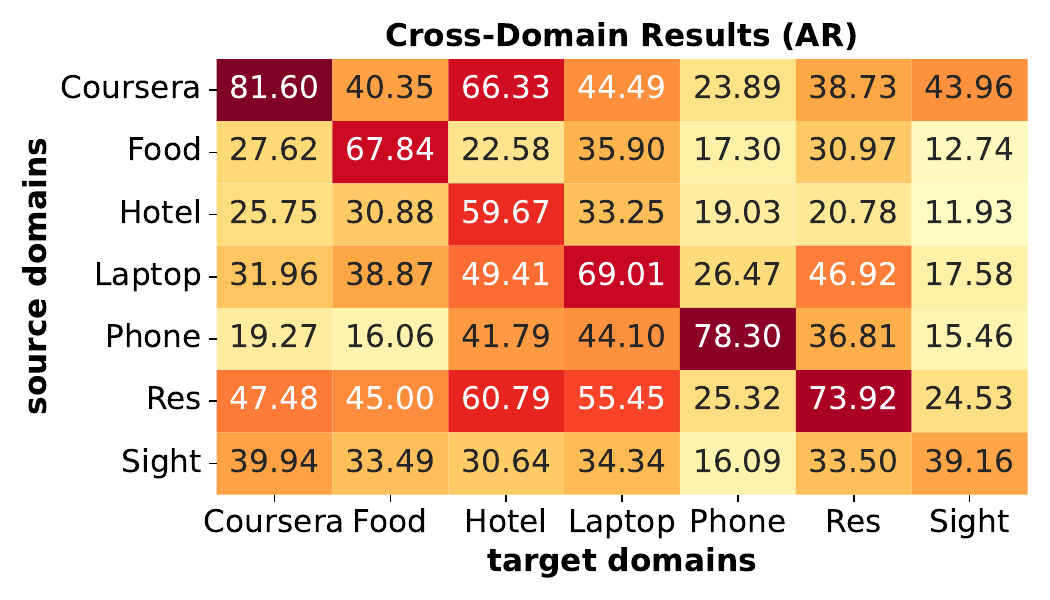}
  \end{minipage}%
  \begin{minipage}{0.3\textwidth}
 \centering
 \includegraphics[width=\linewidth]{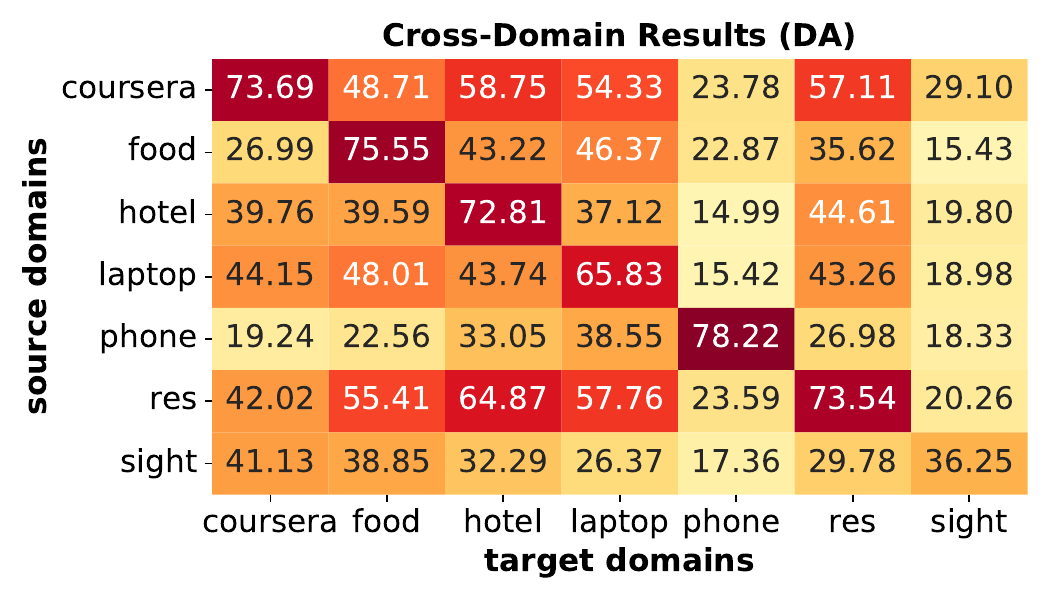}
  \end{minipage}%
  \begin{minipage}{0.3\textwidth}
 \centering
 \includegraphics[width=\linewidth]{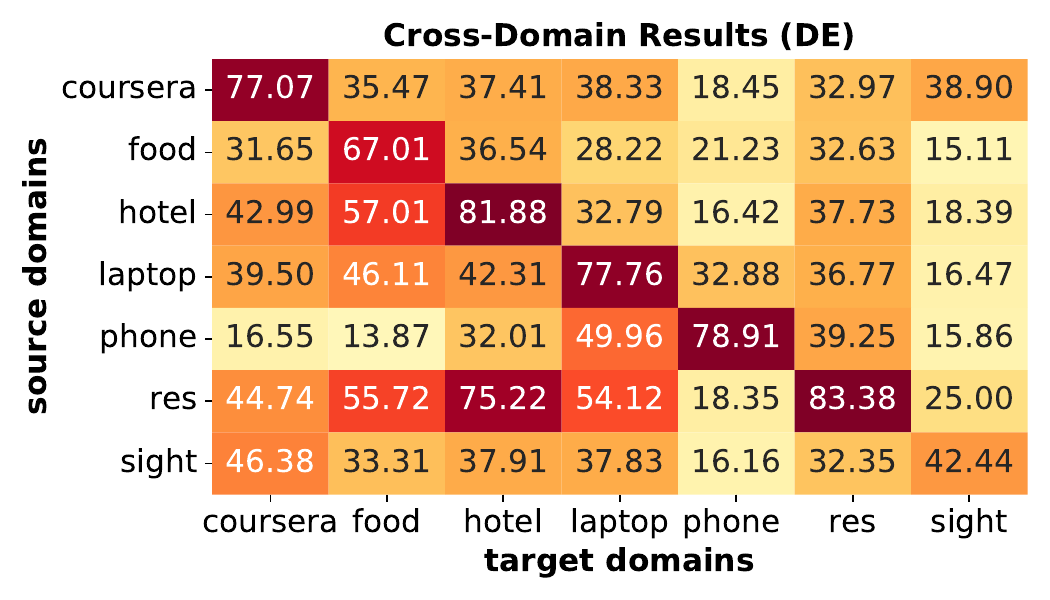}
  \end{minipage}

  \vspace{0.5cm}  

  \begin{minipage}{0.3\textwidth}
 \centering
 \includegraphics[width=\linewidth]{figs/cross-domain/domain_en.pdf}
  \end{minipage}%
  \begin{minipage}{0.3\textwidth}
 \centering
 \includegraphics[width=\linewidth]{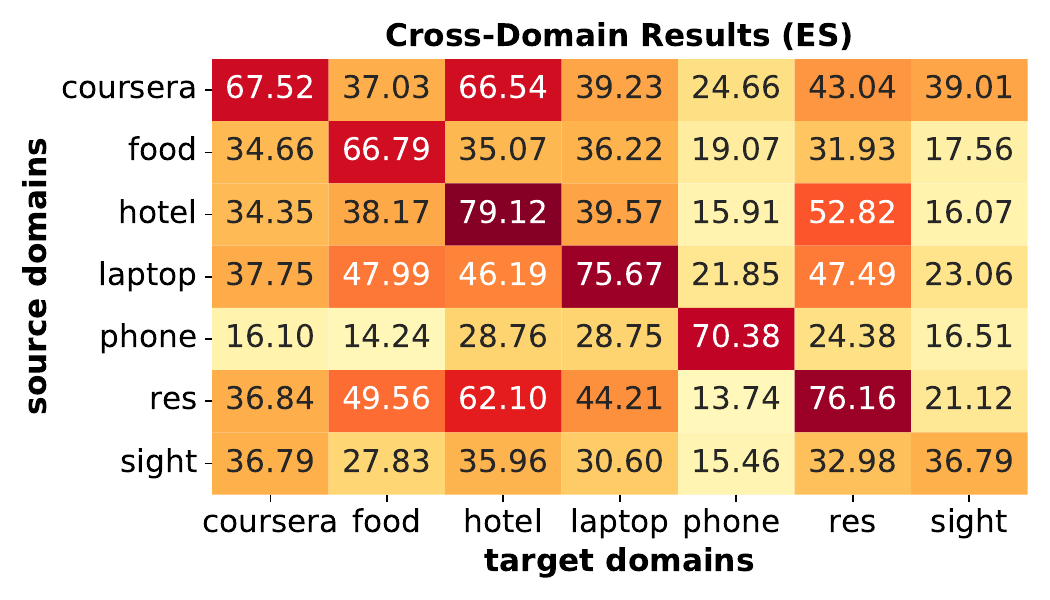}
  \end{minipage}%
  \begin{minipage}{0.3\textwidth}
 \centering
 \includegraphics[width=\linewidth]{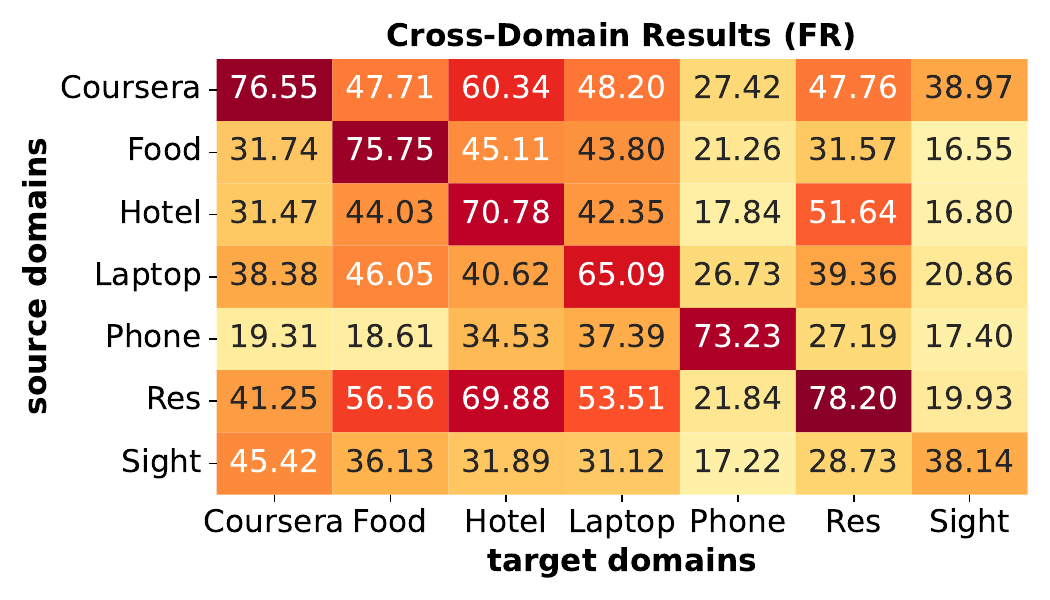}
  \end{minipage}

  \vspace{0.5cm}  

  \begin{minipage}{0.3\textwidth}
 \centering
 \includegraphics[width=\linewidth]{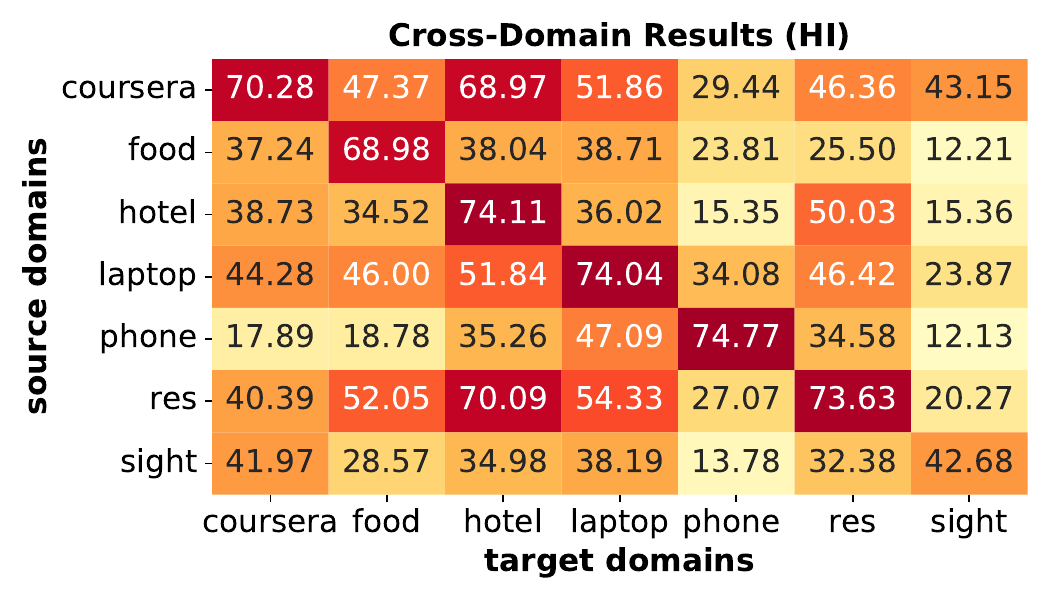}
  \end{minipage}%
  \begin{minipage}{0.3\textwidth}
 \centering
 \includegraphics[width=\linewidth]{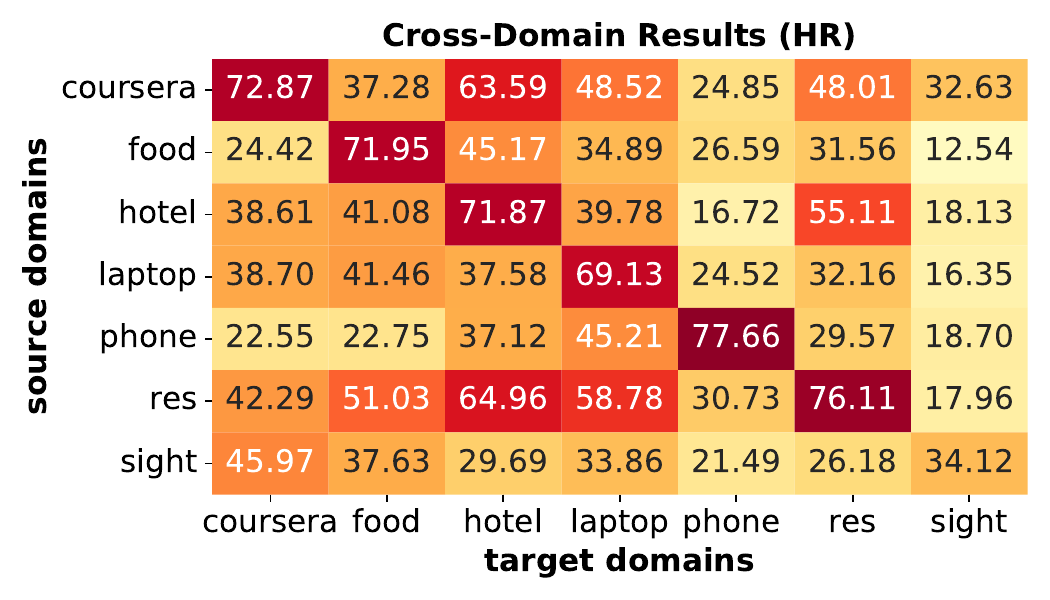}
  \end{minipage}%
  \begin{minipage}{0.3\textwidth}
 \centering
 \includegraphics[width=\linewidth]{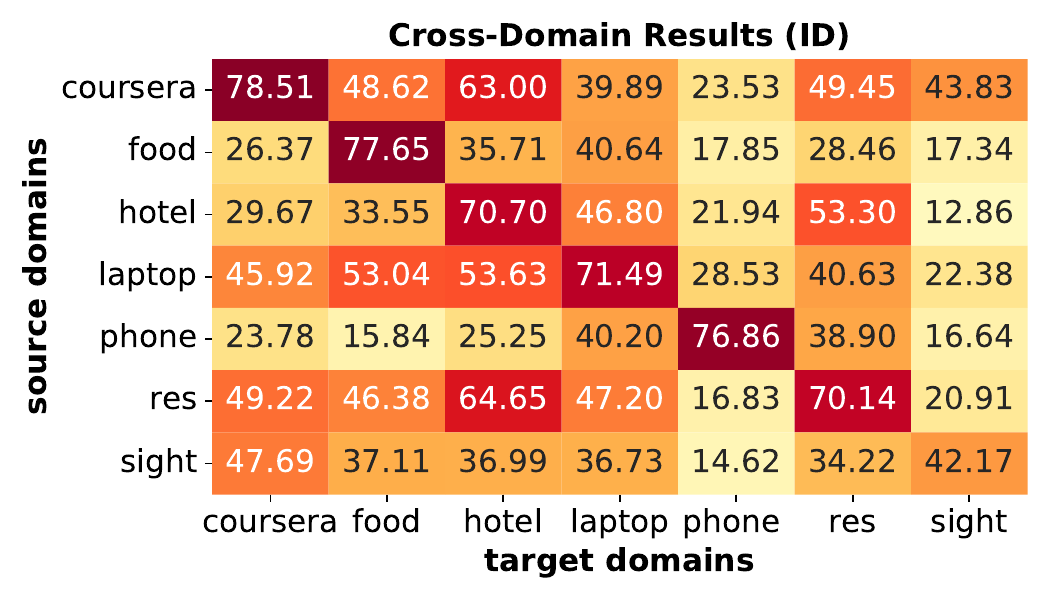}
  \end{minipage}

  \vspace{0.5cm}  

  \begin{minipage}{0.3\textwidth}
 \centering
 \includegraphics[width=\linewidth]{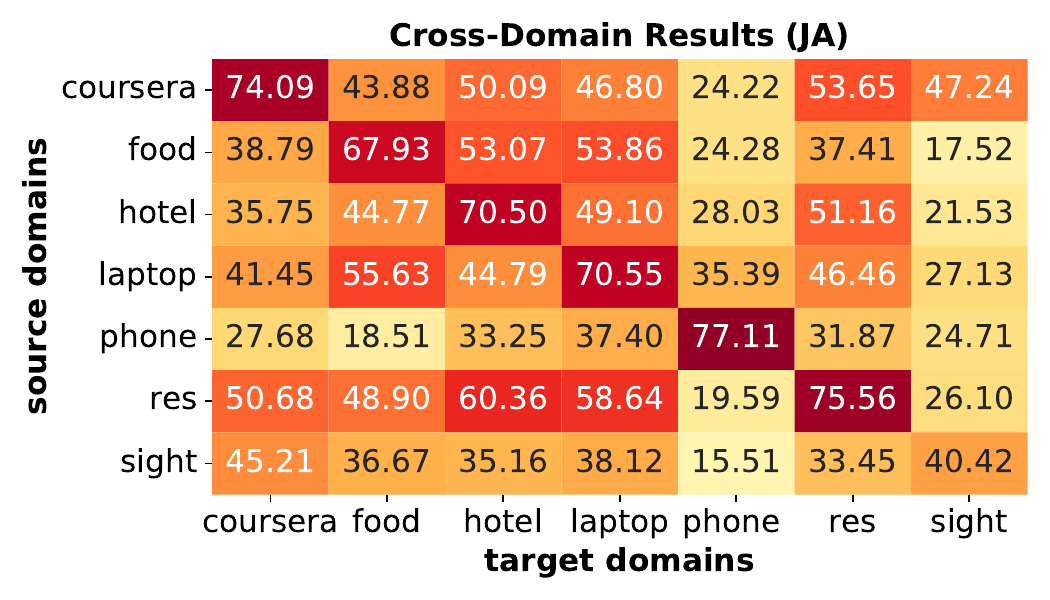}
  \end{minipage}%
  \begin{minipage}{0.3\textwidth}
 \centering
 \includegraphics[width=\linewidth]{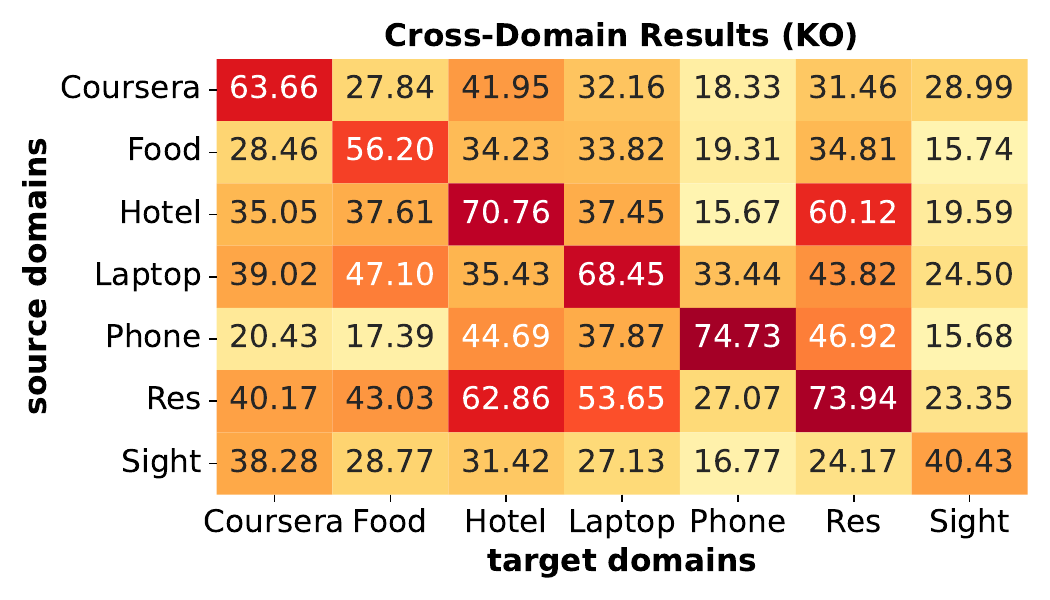}
  \end{minipage}%
  \begin{minipage}{0.3\textwidth}
 \centering
 \includegraphics[width=\linewidth]{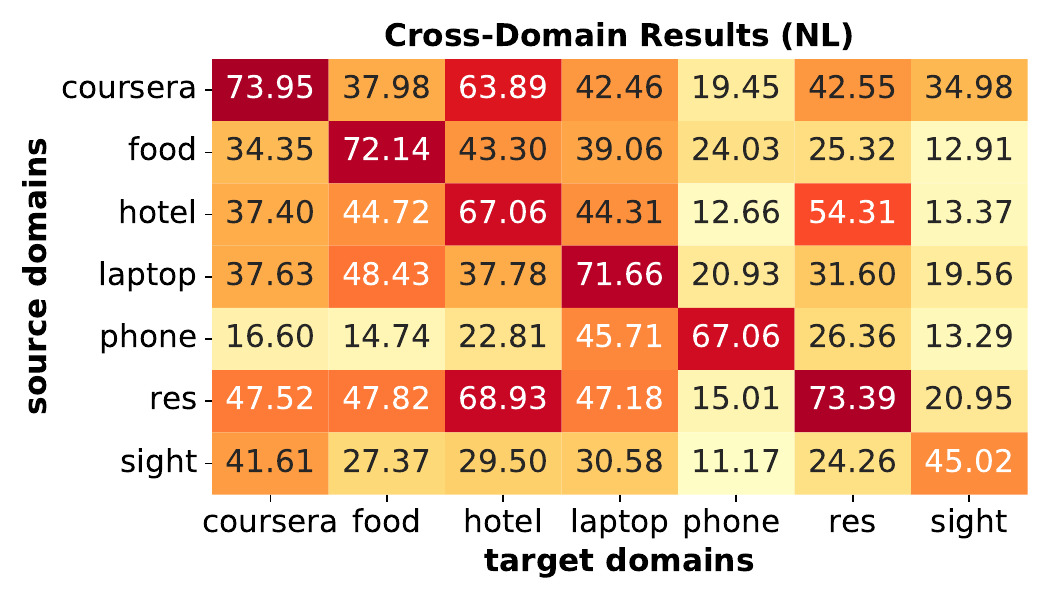}
  \end{minipage}

\vspace{0.5cm}  

  \begin{minipage}{0.3\textwidth}
 \centering
 \includegraphics[width=\linewidth]{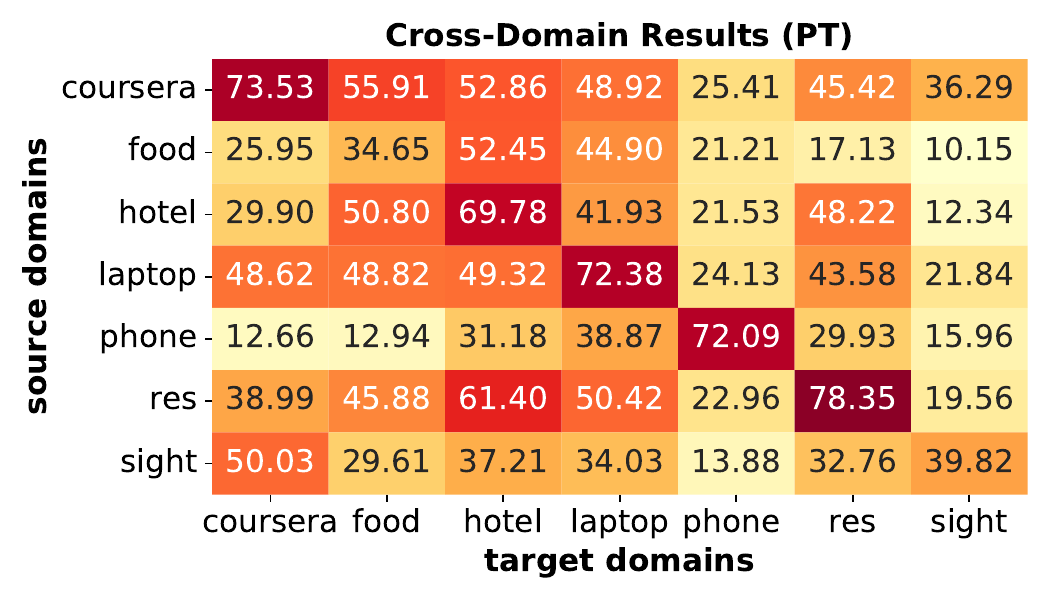}
  \end{minipage}%
  \begin{minipage}{0.3\textwidth}
 \centering
 \includegraphics[width=\linewidth]{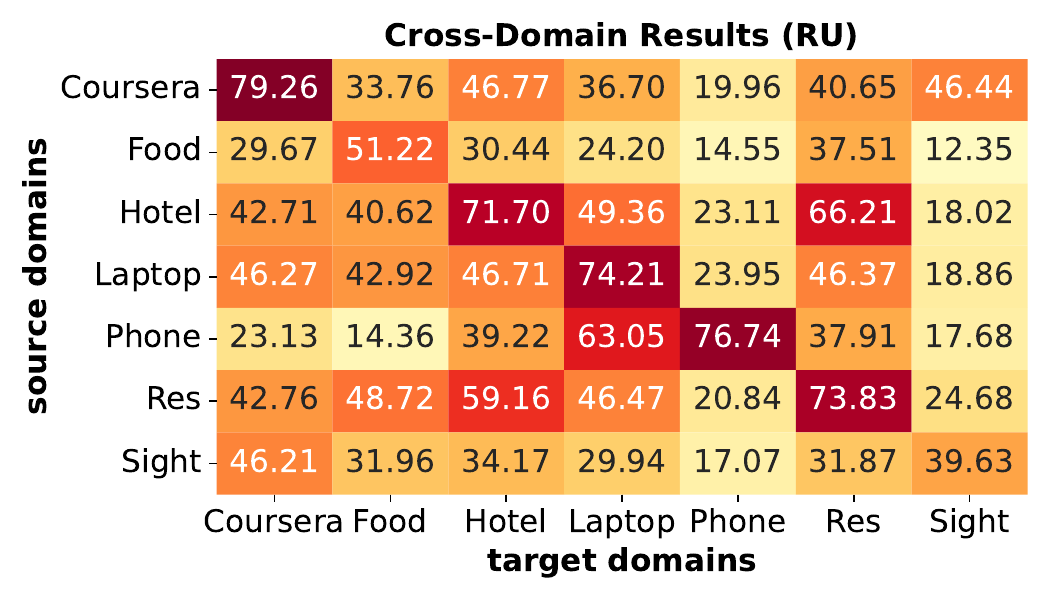}
  \end{minipage}%
  \begin{minipage}{0.3\textwidth}
 \centering
 \includegraphics[width=\linewidth]{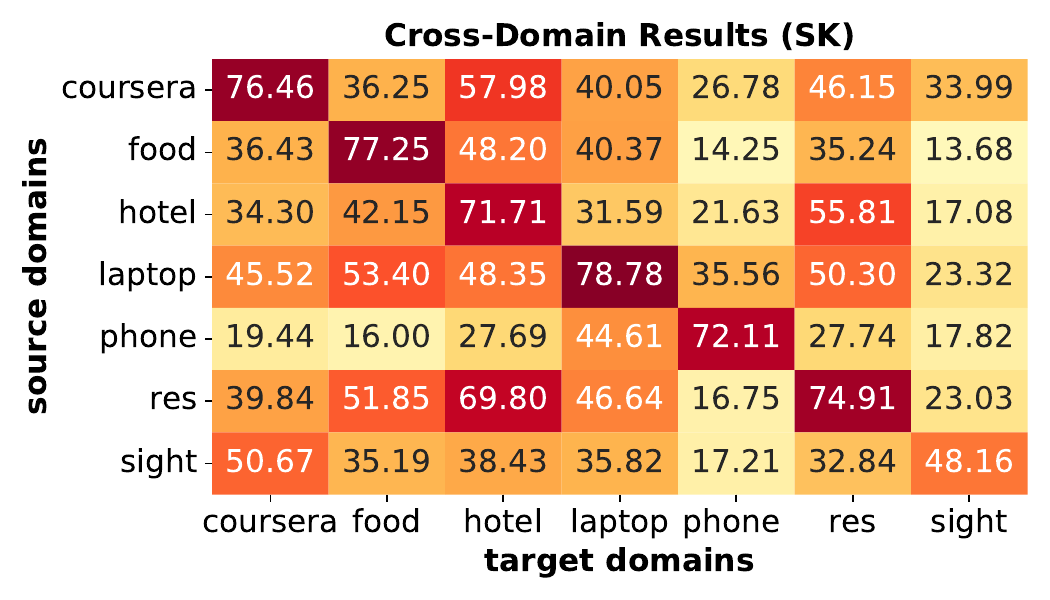}
  \end{minipage}

  \vspace{0.5cm}  

  \begin{minipage}{0.3\textwidth}
 \centering
 \includegraphics[width=\linewidth]{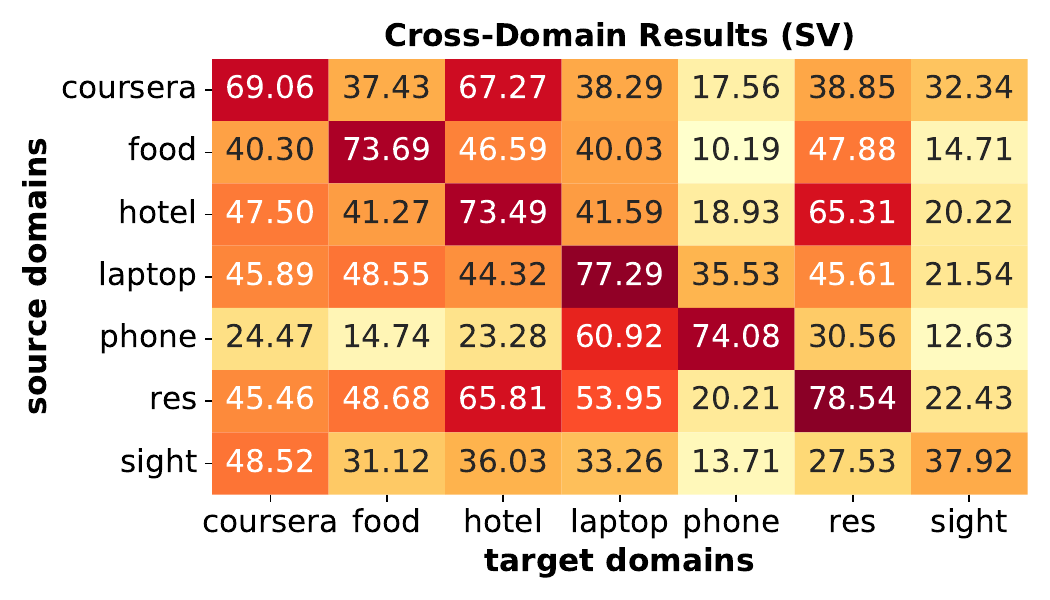}
  \end{minipage}%
  \begin{minipage}{0.3\textwidth}
 \centering
 \includegraphics[width=\linewidth]{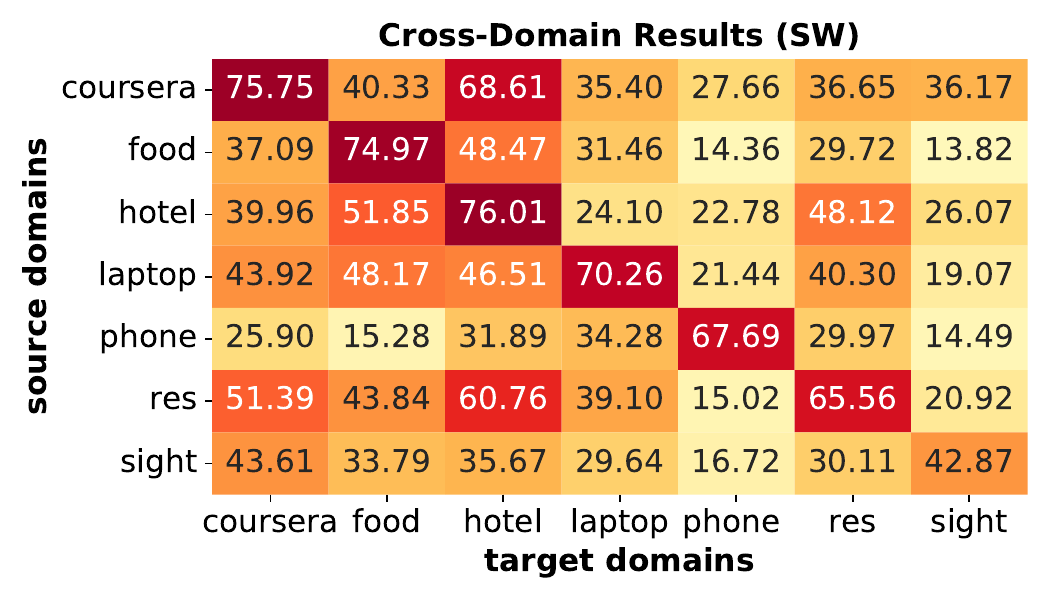}
  \end{minipage}%
  \begin{minipage}{0.3\textwidth}
 \centering
 \includegraphics[width=\linewidth]{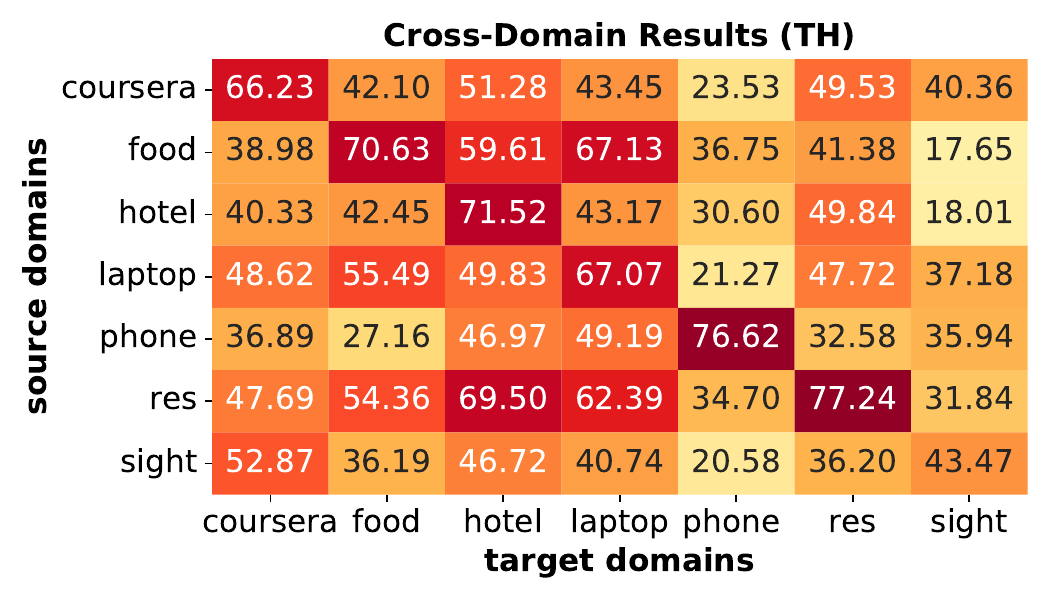}
  \end{minipage}

  \vspace{0.5cm}  

  \begin{minipage}{0.3\textwidth}
 \centering
 \includegraphics[width=\linewidth]{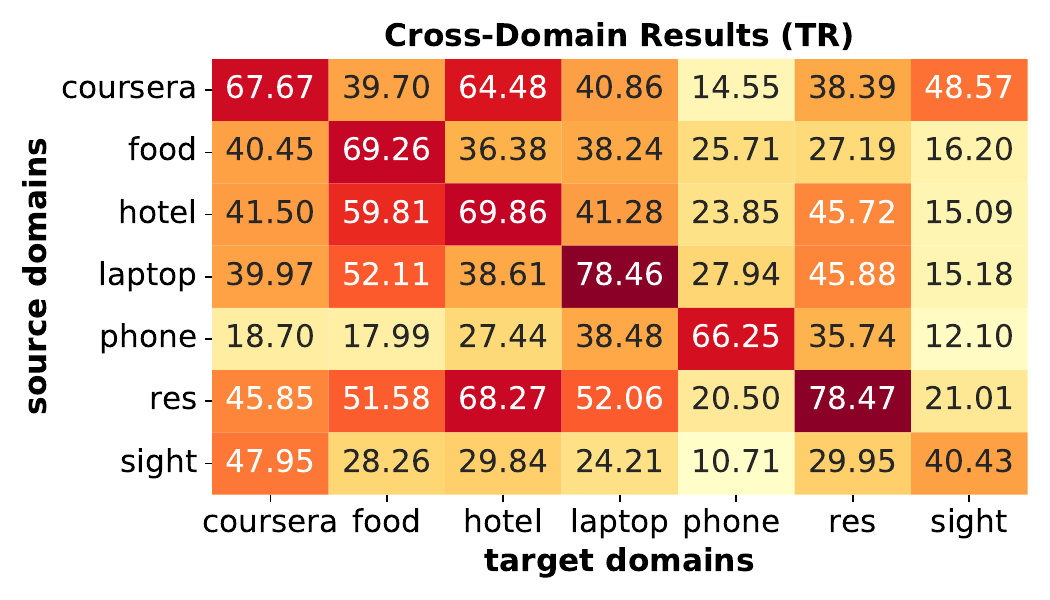}
  \end{minipage}%
  \begin{minipage}{0.3\textwidth}
 \centering
 \includegraphics[width=\linewidth]{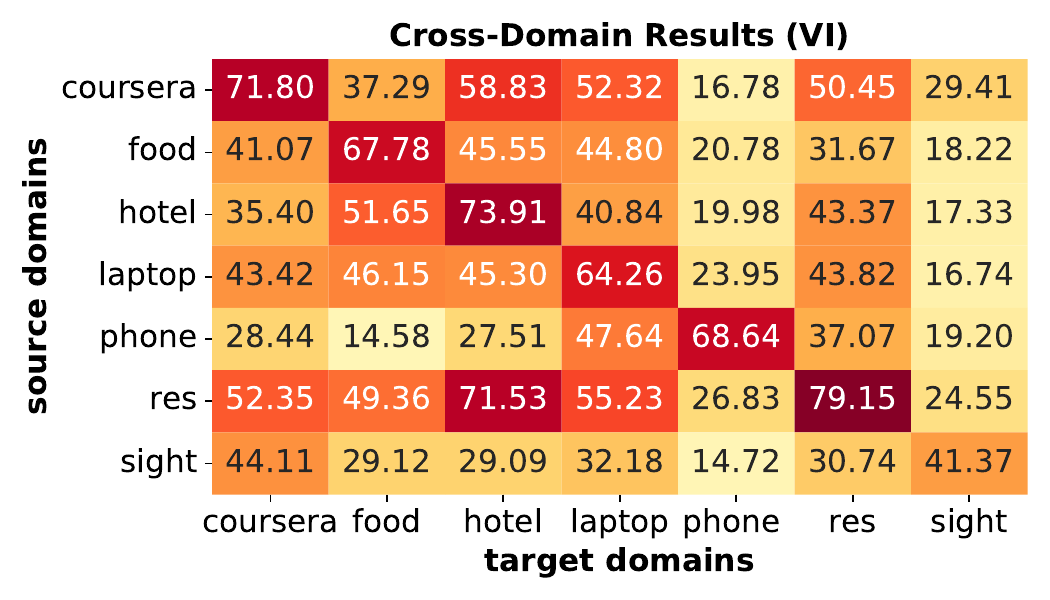}
  \end{minipage}%
  \begin{minipage}{0.3\textwidth}
 \centering
 \includegraphics[width=\linewidth]{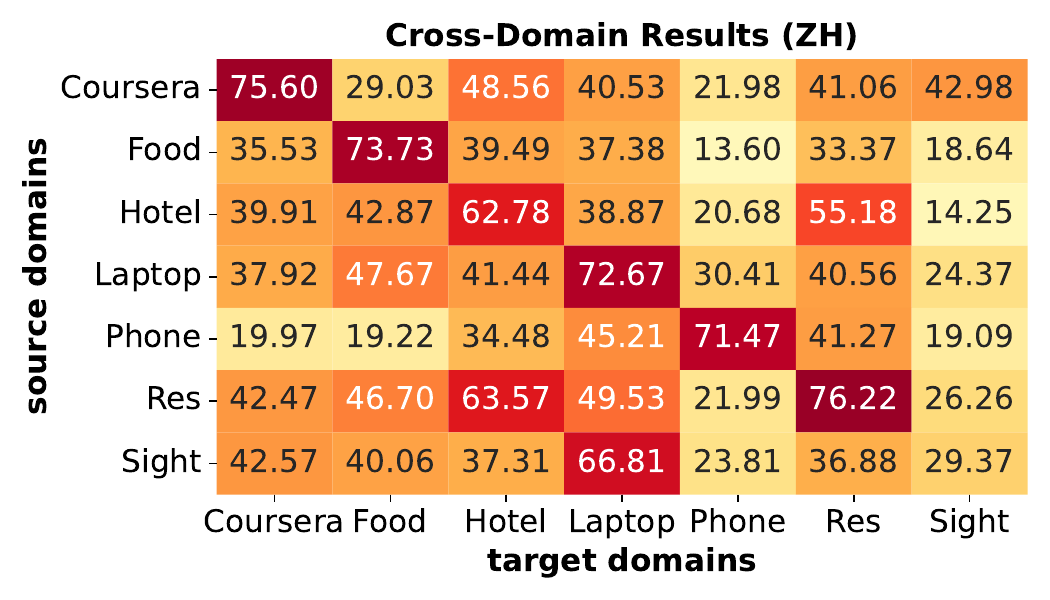}
  \end{minipage}
  
  \caption{Cross-domain results on all 21 languages.}
  \label{fig:cross-domain-all}
\end{figure*}

In Figure \ref{fig:cross-domain-all}, we present the additional cross-domain results of all languages.

\begin{table*}[htbp]
\small
\centering
\begin{tabular}{l|rrrrr|rrrrr}
\toprule
& \multicolumn{5}{c}{Random Baseline} & \multicolumn{5}{c}{Our Dataset} \\
\cmidrule(lr){2-6} \cmidrule(lr){7-11}
Lang. & Acc & chrF++ & BERTScore & SBERT & BLEU & Acc & chrF++ & BERTScore & SBERT & BLEU \\
\midrule
ar & 0.00 & 8.90 & 83.70 & 15.90 & 0.20& 86.00 & 96.20 & 95.50 & 90.80 & 53.30 \\
da & 3.50 & 12.20 & 83.70 & 16.30 & 0.20& 79.00 & 97.70 & 96.80 & 95.60 & 69.00 \\
de & 7.00 & 12.60 & 83.70 & 15.90 & 0.30& 70.20 & 96.10 & 95.80 & 93.10 & 52.80  \\
es & 10.50 & 11.70 & 83.70 & 16.20 & 0.20& 82.50 & 98.30 & 96.10 & 92.80 & 57.90 \\
fr & 5.30 & 11.60 & 83.70 & 16.00 & 0.20& 75.40 & 97.40 & 96.30 & 94.10 & 59.20 \\
hi & 0.00 & 9.10 & 83.70 & 15.70 & 0.20& 75.40 & 95.30 & 95.60 & 92.30 & 50.70 \\
hr & 3.50 & 10.60 & 83.60 & 15.80 & 0.20& 57.90 & 93.60 & 96.10 & 93.30 & 59.30 \\
id & 5.30 & 13.40 & 83.70 & 15.80 & 0.20& 75.40 & 95.20 & 95.40 & 88.40 & 48.10 \\
ja & 0.00 & 4.10 & 83.80 & 15.70 & 0.20& 87.70 & 89.80 & 94.50 & 88.50 & 35.50 \\
ko & 0.00 & 4.00 & 83.80 & 15.70 & 0.20& 79.00 & 91.20 & 93.90 & 85.90 & 33.40  \\
nl & 0.00 & 12.00 & 83.70 & 15.90 & 0.20& 87.70 & 96.60 & 95.70 & 92.00 & 50.70 \\
pt & 0.00 & 11.50 & 83.70 & 15.80 & 0.20& 86.00 & 95.70 & 96.10 & 92.10 & 56.20 \\
ru & 3.50 & 10.20 & 83.70 & 15.80 & 0.20& 70.20 & 95.90 & 95.20 & 89.70 & 45.60 \\
sk & 0.00 & 9.90 & 83.60 & 16.10 & 0.20& 68.40 & 93.80 & 96.10 & 93.90 & 58.80 \\
sv & 1.80 & 11.40 & 83.70 & 16.10 & 0.20& 80.70 & 97.50 & 96.80 & 95.00 & 67.80 \\
sw & 0.00 & 12.50 & 83.70 & 15.20 & 0.20& 61.40 & 93.50 & 95.60 & 92.50 & 57.30 \\
th & 17.50 & 7.50 & 83.60 & 15.70 & 0.20& 79.00 & 91.30 & 94.20 & 86.80 & 34.50 \\
tr & 0.00 & 10.50 & 83.70 & 16.00 & 0.20& 73.70 & 94.50 & 95.00 & 89.30 & 46.10 \\
vi & 1.80 & 10.10 & 83.70 & 16.40 & 0.20& 71.90 & 95.40 & 95.10 & 89.60 & 46.10 \\
zh & 7.00 & 1.700 & 83.70 & 15.80 & 0.30& 84.20 & 87.30 & 94.60 & 87.50 & 40.60 \\
\bottomrule
\end{tabular}
\caption{Evaluation of Translation Quality of random baseline and our dataset on \textbf{Food} domain.}
\label{tab:baseline_evaluation-results_food}
\end{table*}

\begin{table*}[htbp]
\small
\centering
\begin{tabular}{l|rrrrr|rrrrr}
\toprule
& \multicolumn{5}{c}{Random Baseline} & \multicolumn{5}{c}{Our Dataset} \\
\cmidrule(lr){2-6} \cmidrule(lr){7-11}
Lang. & Acc & chrF++ & BERTScore & SBERT & BLEU & Acc & chrF++ & BERTScore & SBERT & BLEU \\
ar & 13.60 & 11.10 & 84.50 & 25.90 & 1.80& 81.60 & 90.90 & 95.70 & 91.00 & 55.00 \\
da & 10.10 & 13.90 & 84.30 & 25.50 & 1.70& 79.10 & 93.70 & 97.00 & 95.70 & 70.70 \\
de & 18.10 & 14.90 & 84.50 & 25.70 & 1.70& 92.10 & 91.40 & 95.80 & 91.50 & 53.00 \\
es & 17.60 & 14.80 & 84.50 & 25.60 & 1.70& 74.80 & 93.10 & 96.30 & 93.50 & 60.50 \\
fr & 18.80 & 14.20 & 84.40 & 25.40 & 1.80& 75.00 & 94.30 & 96.20 & 92.90 & 59.20 \\
hi & 17.60 & 10.80 & 84.50 & 25.50 & 1.50& 91.20 & 89.80 & 95.80 & 92.10 & 53.50 \\
hr & 8.80 & 13.10 & 84.40 & 25.50 & 1.70& 80.80 & 91.80 & 96.10 & 93.20 & 59.50 \\
id & 10.30 & 15.40 & 84.40 & 25.50 & 1.60& 69.80 & 91.60 & 95.90 & 91.50 & 54.60  \\
ja & 14.70 & 4.80 & 84.40 & 25.50 & 0.80& 86.80 & 82.00 & 94.70 & 89.20 & 39.00  \\
ko & 16.80 & 5.90 & 84.50 & 25.10 & 1.60& 88.80 & 84.10 & 94.40 & 87.60 & 35.40 \\
nl & 10.30 & 14.80 & 84.50 & 25.90 & 1.70& 78.60 & 93.40 & 95.90 & 92.70 & 53.70 \\
pt & 12.00 & 13.80 & 84.40 & 25.50 & 1.60& 75.20 & 93.90 & 96.30 & 92.90 & 61.10 \\
ru & 9.60 & 13.30 & 84.40 & 25.20 & 1.40& 73.60 & 91.00 & 95.30 & 90.60 & 48.00 \\
sk & 9.60 & 11.80 & 84.40 & 25.60 & 1.50& 62.40 & 92.40 & 96.30 & 93.70 & 62.40  \\
sv & 12.60 & 12.90 & 84.30 & 25.40 & 1.50& 82.70 & 93.30 & 97.00 & 95.10 & 70.60 \\
sw & 11.10 & 14.70 & 84.40 & 25.60 & 1.60& 69.80 & 90.00 & 95.80 & 91.60 & 57.90 \\
th & 27.60 & 9.40 & 84.50 & 24.70 & 0.80& 91.30 & 84.80 & 94.60 & 87.40 & 38.30 \\
tr & 7.90 & 12.20 & 84.40 & 25.80 & 1.70& 72.20 & 90.20 & 95.30 & 91.10 & 47.90 \\
vi & 12.70 & 12.00 & 84.40 & 25.50 & 1.20& 76.20 & 92.50 & 95.60 & 90.40 & 50.40 \\
zh & 16.80 & 2.20 & 84.40 & 25.10 & 0.20& 89.60 & 77.70 & 94.90 & 89.90 & 43.50 \\
\bottomrule
\end{tabular}
\caption{Evaluation of Translation Quality of random baseline and our dataset on \textbf{Hotel} domain.}
\label{tab:baseline_evaluation-results_hotel}
\end{table*}

\begin{table*}[htbp]
\small
\centering
\begin{tabular}{l|rrrrr|rrrrr}
\toprule
& \multicolumn{5}{c}{Random Baseline} & \multicolumn{5}{c}{Our Dataset} \\
\cmidrule(lr){2-6} \cmidrule(lr){7-11}
Lang. & Acc & chrF++ & BERTScore & SBERT & BLEU & Acc & chrF++ & BERTScore & SBERT & BLEU \\
ar & 3.30 & 10.10 & 84.20 & 20.70 & 0.20& 69.10 & 86.50 & 95.20 & 91.70 & 49.50 \\
da & 2.60 & 13.20 & 84.10 & 21.50 & 0.20& 84.60 & 91.90 & 95.90 & 94.80 & 61.70 \\
de & 2.50 & 13.20 & 84.20 & 21.30 & 0.20& 80.90 & 90.10 & 95.20 & 91.90 & 48.20  \\
es & 4.00 & 13.40 & 84.20 & 21.40 & 0.20& 76.00 & 91.00 & 95.40 & 92.10 & 52.60 \\
fr & 3.40 & 12.70 & 84.10 & 21.40 & 0.20& 77.50 & 91.80 & 95.70 & 93.10 & 56.80 \\
hi & 3.80 & 9.50 & 84.20 & 21.50 & 0.20& 87.90 & 89.00 & 95.80 & 94.90 & 56.40 \\
hr & 2.60 & 12.10 & 84.20 & 21.30 & 0.20& 76.10 & 87.10 & 95.50 & 93.10 & 54.10 \\
id & 1.90 & 13.40 & 84.10 & 21.00 & 0.20& 73.40 & 86.60 & 95.30 & 92.00 & 50.60 \\
ja & 2.50 & 4.50 & 84.20 & 21.50 & 0.20& 82.20 & 73.20 & 94.00 & 89.90 & 35.80 \\
ko & 4.50 & 5.00 & 84.20 & 21.60 & 0.20& 87.90 & 77.10 & 93.60 & 86.40 & 31.90  \\
nl & 3.20 & 13.30 & 84.20 & 21.60 & 0.20& 76.40 & 86.70 & 95.40 & 92.20 & 52.40 \\
pt & 1.30 & 12.40 & 84.10 & 20.80 & 0.20& 76.60 & 91.30 & 95.50 & 93.60 & 55.10 \\
ru & 0.70 & 10.70 & 84.10 & 21.50 & 0.20& 65.80 & 88.80 & 94.70 & 90.30 & 44.10 \\
sk & 1.90 & 10.90 & 84.20 & 21.70 & 0.20& 65.60 & 85.60 & 95.40 & 93.50 & 52.30 \\
sv & 2.50 & 12.50 & 84.10 & 20.90 & 0.20& 81.50 & 90.10 & 95.90 & 94.80 & 59.90 \\
sw & 3.80 & 13.30 & 84.10 & 21.20 & 0.20& 67.50 & 83.50 & 95.30 & 92.10 & 54.30 \\
th & 1.40 & 8.30 & 84.10 & 21.50 & 0.20& 64.60 & 75.10 & 94.00 & 87.90 & 36.10 \\
tr & 3.20 & 11.50 & 84.10 & 21.30 & 0.20& 66.90 & 87.30 & 94.70 & 89.80 & 44.50 \\
vi & 4.30 & 10.40 & 84.10 & 21.40 & 0.20& 69.50 & 87.90 & 94.90 & 89.50 & 45.00 \\
zh & 2.50 & 2.40 & 84.20 & 21.90 & 0.20& 82.20 & 67.30 & 94.20 & 89.60 & 39.40 \\
\bottomrule
\end{tabular}
\caption{Evaluation of Translation Quality of random baseline and our dataset on \textbf{Laptop} domain.}
\label{tab:baseline_evaluation-results_laptop}
\end{table*}

\begin{table*}[htbp]
\small
\centering
\begin{tabular}{l|rrrrr|rrrrr}
\toprule
& \multicolumn{5}{c}{Random Baseline} & \multicolumn{5}{c}{Our Dataset} \\
\cmidrule(lr){2-6} \cmidrule(lr){7-11}
Lang. & Acc & chrF++ & BERTScore & SBERT & BLEU & Acc & chrF++ & BERTScore & SBERT & BLEU \\
\midrule
ar & 2.70 & 8.80 & 83.70 & 18.90 & 0.20& 63.40 & 82.70 & 95.30 & 89.30 & 52.30 \\
da & 2.10 & 11.70 & 83.60 & 19.20 & 0.20& 68.50 & 89.80 & 96.80 & 95.50 & 69.00 \\
de & 3.70 & 12.20 & 83.60 & 19.00 & 0.20& 87.80 & 89.50 & 96.00 & 91.60 & 56.80 \\
es & 3.30 & 11.70 & 83.70 & 19.20 & 0.20& 86.90 & 91.90 & 96.00 & 91.30 & 56.80  \\
fr & 2.70 & 11.60 & 83.60 & 19.00 & 0.20& 78.40 & 88.80 & 95.90 & 91.50 & 60.60 \\
hi & 2.60 & 8.40 & 83.60 & 18.80 & 0.20& 83.10 & 85.90 & 95.50 & 90.30 & 54.80 \\
hr & 2.10 & 11.10 & 83.60 & 19.30 & 0.20& 60.30 & 87.60 & 95.90 & 92.40 & 60.40 \\
id & 3.20 & 13.30 & 83.70 & 19.20 & 0.20& 64.30 & 85.40 & 95.10 & 88.80 & 47.20 \\
ja & 3.70 & 3.90 & 83.70 & 19.10 & 0.20& 79.90 & 67.80 & 94.30 & 86.40 & 36.30  \\
ko & 4.20 & 4.70 & 83.70 & 19.60 & 0.20& 86.20 & 76.00 & 94.00 & 85.50 & 34.70 \\
nl & 2.60 & 12.10 & 83.60 & 19.20 & 0.20& 70.90 & 87.50 & 95.90 & 91.70 & 56.90 \\
pt & 2.60 & 11.50 & 83.70 & 19.20 & 0.20& 82.50 & 90.80 & 95.90 & 92.40 & 56.20 \\
ru & 1.60 & 10.50 & 83.60 & 19.30 & 0.20& 60.30 & 83.30 & 95.00 & 88.80 & 46.50 \\
sk & 2.10 & 9.70 & 83.60 & 19.20 & 0.20& 58.20 & 88.80 & 96.10 & 93.00 & 61.40 \\
sv & 1.60 & 10.70 & 83.60 & 19.10 & 0.20& 63.60 & 90.00 & 96.60 & 94.10 & 67.70 \\
sw & 3.70 & 12.40 & 83.60 & 18.80 & 0.20& 49.20 & 80.00 & 95.50 & 88.90 & 57.40 \\
th & 4.90 & 8.00 & 83.60 & 19.00 & 0.30& 74.50 & 70.90 & 94.10 & 84.70 & 35.40 \\
tr & 2.70 & 10.20 & 83.60 & 19.20 & 0.20& 62.80 & 83.50 & 95.10 & 89.00 & 45.90 \\
vi & 3.30 & 9.50 & 83.60 & 19.60 & 0.20& 60.60 & 85.50 & 94.80 & 87.00 & 42.90 \\
zh & 2.70 & 1.40 & 83.60 & 19.30 & 0.20& 79.60 & 59.70 & 94.80 & 87.20 & 39.60 \\
\bottomrule
\end{tabular}
\caption{Evaluation of Translation Quality of random baseline and our dataset on \textbf{Restaurant} domain.}
\label{tab:baseline_evaluation-results_restaurant}
\end{table*}

\begin{table*}[htbp]
\small
\centering
\begin{tabular}{l|rrrrr|rrrrr}
\toprule
& \multicolumn{5}{c}{Random Baseline} & \multicolumn{5}{c}{Our Dataset} \\
\cmidrule(lr){2-6} \cmidrule(lr){7-11}
Lang. & Acc & chrF++ & BERTScore & SBERT & BLEU & Acc & chrF++ & BERTScore & SBERT & BLEU \\
\midrule
ar & 8.30 & 8.60 & 84.70 & 22.60 & 1.50& 62.50 & 67.00 & 95.80 & 91.90 & 59.90 \\
da & 10.40 & 14.10 & 84.70 & 23.10 & 0.90& 78.10 & 95.60 & 96.30 & 94.60 & 65.10 \\
de & 9.40 & 13.90 & 84.80 & 22.70 & 0.90 & 81.20 & 91.20 & 95.40 & 91.20 & 47.10 \\
es & 16.70 & 13.70 & 84.80 & 23.00 & 1.50& 92.70 & 95.00 & 95.70 & 93.00 & 56.80 \\
fr & 13.50 & 13.60 & 84.70 & 22.20 & 1.20& 84.40 & 94.70 & 95.90 & 91.30 & 55.60 \\
hi & 12.50 & 10.90 & 84.70 & 22.70 & 1.40& 80.20 & 91.00 & 95.60 & 91.70 & 50.00 \\
hr & 12.50 & 12.50 & 84.80 & 23.10 & 1.90& 75.00 & 92.20 & 95.80 & 93.50 & 58.90 \\
id & 5.20 & 15.60 & 84.50 & 22.40 & 2.00& 67.70 & 92.80 & 95.30 & 90.80 & 47.60  \\
ja & 12.50 & 5.70 & 84.80 & 23.40 & 1.10& 88.50 & 82.90 & 94.40 & 88.50 & 37.50 \\
ko & 7.30 & 6.30 & 84.90 & 22.10 & 1.10& 57.30 & 80.90 & 94.00 & 84.00 & 32.00 \\
nl & 9.40 & 13.60 & 84.70 & 22.60 & 1.70& 79.20 & 90.40 & 95.50 & 91.40 & 50.00 \\
pt & 10.40 & 13.30 & 84.80 & 22.60 & 1.80& 85.40 & 94.50 & 96.00 & 93.10 & 57.10 \\
ru & 10.40 & 12.00 & 84.80 & 23.30 & 1.30& 76.00 & 92.80 & 95.20 & 89.90 & 45.90 \\
sk & 12.50 & 11.20 & 84.80 & 23.20 & 1.70& 76.00 & 91.50 & 95.70 & 93.20 & 55.90 \\
sv & 13.50 & 13.20 & 84.80 & 22.90 & 1.30& 85.40 & 93.70 & 96.40 & 93.70 & 64.90 \\
sw & 6.20 & 14.20 & 84.80 & 22.60 & 1.60& 58.30 & 89.70 & 95.20 & 89.40 & 50.90 \\
th & 14.60 & 10.30 & 84.60 & 22.40 & 1.40& 81.20 & 83.10 & 94.40 & 85.90 & 34.70 \\
tr & 9.40 & 12.40 & 84.80 & 22.80 & 1.50& 54.20 & 89.50 & 94.80 & 89.30 & 43.00 \\
vi & 9.40 & 11.80 & 84.80 & 23.10 & 1.40& 75.00 & 92.90 & 95.00 & 89.20 & 40.70 \\
zh & 11.50 & 3.90 & 84.70 & 22.70 & 1.60& 83.30 & 82.90 & 94.80 & 88.80 & 38.40 \\
\bottomrule
\end{tabular}
\caption{Evaluation of Translation Quality of random baseline and our dataset on \textbf{Coursera} domain.}
\label{tab:baseline_evaluation-results_coursera}
\end{table*}

\begin{table*}[htbp]
\small
\centering
\begin{tabular}{l|rrrrr|rrrrr}
\toprule
& \multicolumn{5}{c}{Random Baseline} & \multicolumn{5}{c}{Our Dataset} \\
\cmidrule(lr){2-6} \cmidrule(lr){7-11}
Lang. & Acc & chrF++ & BERTScore & SBERT & BLEU & Acc & chrF++ & BERTScore & SBERT & BLEU \\
\midrule
ar & 5.80 & 10.10 & 82.70 & 14.60 & 1.50& 72.30 & 93.10 & 96.10 & 89.60 & 56.30  \\
da & 2.20 & 12.00 & 82.70 & 15.10 & 2.20& 79.60 & 96.00 & 97.70 & 95.10 & 72.20 \\
de & 7.30 & 12.80 & 82.80 & 15.30 & 1.50& 80.30 & 94.30 & 96.70 & 93.20 & 59.90 \\
es & 6.60 & 12.60 & 82.80 & 14.90 & 1.70& 77.90 & 95.10 & 96.40 & 91.50 & 62.80 \\
fr & 5.80 & 12.30 & 82.70 & 15.00 & 1.70& 76.60 & 95.20 & 96.90 & 92.60 & 64.00 \\
hi & 3.60 & 10.20 & 82.80 & 14.90 & 1.60& 78.10 & 92.70 & 96.30 & 91.90 & 59.70 \\
hr & 2.90 & 11.50 & 82.70 & 14.90 & 1.60& 67.90 & 93.40 & 97.00 & 94.10 & 65.80 \\
id & 5.10 & 13.40 & 82.70 & 15.20 & 1.40& 79.60 & 94.70 & 96.70 & 92.80 & 59.50 \\
ja & 5.20 & 4.90 & 82.80 & 14.20 & 1.20& 84.40 & 85.90 & 95.00 & 89.90 & 45.10 \\
ko & 3.60 & 4.90 & 82.80 & 14.60 & 1.30& 73.70 & 87.80 & 95.00 & 89.40 & 45.60 \\
nl & 5.10 & 11.90 & 82.70 & 14.70 & 1.50& 82.50 & 95.30 & 97.00 & 93.70 & 65.80 \\
pt & 5.80 & 12.20 & 82.70 & 14.80 & 1.70& 89.80 & 96.80 & 97.10 & 93.00 & 67.90 \\
ru & 5.10 & 11.10 & 82.70 & 14.90 & 1.50& 65.70 & 92.50 & 96.30 & 92.70 & 57.00 \\
sk & 2.20 & 10.40 & 82.70 & 15.00 & 1.80& 62.00 & 92.90 & 96.90 & 94.00 & 63.00 \\
sv & 3.60 & 12.00 & 82.60 & 15.10 & 2.30& 81.80 & 95.00 & 97.70 & 94.90 & 71.00 \\
sw & 2.90 & 12.40 & 82.80 & 14.60 & 1.90& 64.70 & 93.20 & 96.70 & 89.80 & 64.60 \\
th & 5.80 & 9.00 & 82.80 & 14.40 & 0.80& 76.60 & 88.10 & 95.00 & 88.00 & 45.20 \\
tr & 6.60 & 11.20 & 82.70 & 14.80 & 1.60& 76.60 & 92.80 & 95.90 & 90.20 & 54.00 \\
vi & 5.10 & 10.70 & 82.70 & 15.10 & 1.90& 78.80 & 94.70 & 96.60 & 91.90 & 59.50 \\
zh & 5.80 & 3.40 & 82.70 & 14.90 & 1.40& 86.10 & 88.20 & 95.80 & 91.40 & 50.60 \\
\bottomrule
\end{tabular}
\caption{Evaluation of Translation Quality of random baseline and our dataset on \textbf{Sight} domain.}
\label{tab:baseline_evaluation-results_sight}
\end{table*}

\begin{table*}[htbp]
\small
\centering
\begin{tabular}{l|rrrrr|rrrrr}
\toprule
& \multicolumn{5}{c}{Random Baseline} & \multicolumn{5}{c}{Our Dataset} \\
\cmidrule(lr){2-6} \cmidrule(lr){7-11}
Lang. & Acc & chrF++ & BERTScore & SBERT & BLEU & Acc & chrF++ & BERTScore & SBERT & BLEU \\
\midrule
ar & 3.40 & 10.30 & 83.20 & 16.40 & 0.30& 63.50 & 83.60 & 96.10 & 91.40 & 49.10 \\
da & 1.70 & 13.70 & 83.20 & 16.60 & 0.30& 78.40 & 91.30 & 97.80 & 95.80 & 67.50 \\
de & 3.40 & 14.40 & 83.20 & 16.80 & 0.30& 77.80 & 87.00 & 96.60 & 93.30 & 52.90 \\
es & 2.00 & 14.40 & 83.30 & 16.40 & 0.30& 67.40 & 87.40 & 96.30 & 92.40 & 52.60 \\
fr & 4.20 & 14.40 & 83.20 & 16.80 & 0.30& 67.40 & 88.50 & 96.70 & 93.10 & 55.10 \\
hi & 3.10 & 10.40 & 83.20 & 16.80 & 0.30& 70.80 & 84.20 & 96.60 & 93.50 & 56.50 \\
hr & 2.00 & 13.20 & 83.20 & 16.90 & 0.30& 58.70 & 85.70 & 96.70 & 93.50 & 54.70 \\
id & 3.40 & 15.20 & 83.20 & 17.00 & 0.30& 61.00 & 85.10 & 96.20 & 92.40 & 49.50 \\
ja & 2.20 & 4.90 & 83.40 & 16.60 & 0.30& 72.20 & 70.60 & 94.80 & 89.90 & 36.40 \\
ko & 2.50 & 5.50 & 83.40 & 16.50 & 0.30& 67.40 & 72.60 & 94.60 & 89.20 & 36.10 \\
nl & 2.00 & 14.20 & 83.20 & 16.80 & 0.30& 70.50 & 87.90 & 97.00 & 94.10 & 59.50 \\
pt & 3.90 & 13.80 & 83.30 & 16.60 & 0.30& 70.70 & 89.60 & 96.80 & 93.30 & 57.30 \\
ru & 2.20 & 12.30 & 83.30 & 16.30 & 0.30& 63.80 & 85.70 & 95.90 & 90.80 & 49.20 \\
sk & 1.40 & 12.10 & 83.30 & 16.50 & 0.30& 58.40 & 84.50 & 96.40 & 93.40 & 54.40 \\
sv & 2.80 & 13.10 & 83.20 & 16.80 & 0.30& 81.50 & 90.90 & 97.50 & 94.90 & 64.50 \\
sw & 2.80 & 14.70 & 83.20 & 17.00 & 0.30& 58.40 & 85.50 & 96.40 & 92.10 & 54.70 \\
th & 7.30 & 9.60 & 83.40 & 16.70 & 0.30& 69.10 & 76.00 & 94.50 & 88.40 & 36.60 \\
tr & 2.50 & 12.10 & 83.20 & 16.30 & 0.30& 69.40 & 86.10 & 95.70 & 90.90 & 47.60 \\
vi & 4.20 & 11.70 & 83.20 & 16.70 & 0.30& 69.90 & 88.00 & 96.00 & 90.90 & 50.30 \\
zh & 4.30 & 2.20 & 83.30 & 16.90 & 0.30& 75.80 & 71.10 & 95.20 & 89.70 & 42.80 \\
\bottomrule
\end{tabular}
\caption{Evaluation of Translation Quality of random baseline and our dataset on \textbf{Phone} domain.}
\label{tab:baseline_evaluation-results_phone}
\end{table*}

\begin{figure*}
\centering

\begin{tcolorbox}[colframe=gray!50!black, colback=gray!5!white, title=Unified ABSA]
\small
\vspace{3pt}
Aspect-Based Sentiment Analysis (ABSA) involves identifying specific entity (such as a person, product, service, or experience) mentioned in a text and determining the sentiment expressed toward each entity.

\vspace{3pt}
Each entity is associated with a sentiment that can be [positive, negative, or neutral].

\vspace{3pt}
Your task is to:

\begin{enumerate}
\item Identify the entity with a sentiment mentioned in the given text.
\item For each identified entity, determine the sentiment in the label set (positive, negative, or neutral).
\item The output should be a list of dictionaries, where each dictionary contains the entity with a sentiment and its corresponding sentiment. If there are no sentiment-bearing entities in the text, the output should be an empty list.
\end{enumerate}

Example Output format:

\begin{itemize}
\item [] [{``entity'': ``<entity>'', ``sentiment'': ``<label>''}]
\end{itemize}

Please return the final output based on the following text in JSON format.

\vspace{3pt}
\end{tcolorbox}

\begin{tcolorbox}[colframe=gray!50!black, colback=gray!5!white, title=Target-Aspect-Sentiment Detection]
\small
\vspace{3pt}
Aspect-Based Sentiment Analysis (ABSA) requires identifying specific entities mentioned in a text and determining the sentiment expressed toward each entity.

\vspace{3pt}
Each entity is associated with:
\begin{itemize}
\item [-] A category from the list: \{str(get\_category(args.type))\}.
\item [-] A sentiment: [positive, negative, neutral].
\end{itemize}

Your task is to:

\begin{enumerate}
\item Identify entities in the text, along with their categories and sentiments.
\item For each identified entity, assign a category from the provided category list.
\item Determine the sentiment for each entity as one of [positive, negative, neutral].
\item Return the results as a list of dictionaries, each containing the entity, category, and sentiment. If no entities are found, return an empty list.
\end{enumerate}

Example Output format:

\begin{itemize}
\item [] [{``entity'': ``<entity>'', ``category'': ``<category>'', ``sentiment'': ``<sentiment>''}]
\end{itemize}

Please return the final output based on the following text in JSON format.

\vspace{3pt}
\end{tcolorbox}

\caption{Prompts for UABSA and TASD tasks.}
\label{fig:absa-prompt}
\end{figure*}

\begin{table*}[htbp]
\scriptsize
\setlength\tabcolsep{2pt}\centering
\begin{tabular}{ccccccccccccccccccccccc}
\toprule
&ar	&da	&de	&en	&es	&fr	&hi	&hr	&id	&ja	&ko	&nl	&pt	&ru	&sk	&sv	&sw	&th	&tr	&vi	&zh &avg \\
\midrule
ar & 38.81& 24.26& 32.37& 17.85& 20.90 & 18.93& 20.46& 27.63& 26.95& 24.89& 16.70 & 21.51& 21.07& 27.22& 24.58& 28.40 & 19.49& 28.85& 22.60 & 20.82& 29.88& 24.48\\
fr & 12.72& 26.24& 31.74& 23.86& 29.72& 50.32& 15.95& 20.66& 28.46& 19.16& 14.49& 30.46& 33.87& 27.44& 23.45& 27.82& 16.80 & 22.07& 27.56& 15.87& 26.61& 25.01\\
ko & 22.16& 26.90 & 31.24& 31.76& 21.97& 21.34& 20.58& 12.64& 30.33& 32.22& 47.51& 23.73& 26.55& 28.23& 30.2 & 26.68& 14.20 & 26.87& 24.34& 24.89& 32.80 & 26.53\\
ru & 23.90 & 26.34& 38.62& 23.50 & 24.23& 19.86& 25.50 & 19.21& 25.33& 28.51& 20.90 & 29.89& 21.99& 60.40 & 27.96& 27.32& 19.96& 28.59& 24.67& 24.12& 36.81& 27.51 \\
zh & 21.00& 31.61& 29.94& 18.42& 30.97& 19.56& 18.09& 18.74& 24.66& 34.98& 20.27& 29.10 & 27.37& 31.71& 25.83& 32.74& 22.17& 26.73& 26.96& 21.94& 62.70 & 27.40 \\
\bottomrule
\end{tabular}
\caption{Cross-lingual TASD results on the Coursera dataset with non-English source languages.}
\label{tab:non-en-7}
\end{table*}

\begin{table*}[htbp]
\scriptsize
\setlength\tabcolsep{2pt}\centering
\begin{tabular}{ccccccccccccccccccccccc}
\toprule
&ar	&da	&de	&en	&es	&fr	&hi	&hr	&id	&ja	&ko	&nl	&pt	&ru	&sk	&sv	&sw	&th	&tr	&vi	&zh &avg \\
\midrule
ar & 81.60 & 43.91& 45.39& 48.13& 54.96& 39.49& 41.04& 55.30 & 50.30 & 45.12& 35.31& 43.35& 51.36& 52.16& 62.23& 49.16& 54.61& 46.21& 52.42& 45.70 & 60.32& 50.38\\
fr & 38.84& 58.48& 69.38& 62.37& 60.80 & 76.55& 41.84& 61.26& 63.07& 41.19& 36.25& 56.70 & 67.06& 64.83& 61.82& 65.41& 39.02& 40.57& 62.26& 48.06& 60.38& 56.01 \\
ko & 30.99& 32.94& 36.52& 35.06& 30.59& 27.72& 30.44& 23.49& 33.65& 36.85& 63.66& 28.96& 30.27& 31.63& 32.99& 33.09& 25.89& 37.59& 33.21& 32.33& 40.74& 33.74\\
ru & 27.17& 34.79& 55.84& 33.43& 44.82& 41.58& 36.36& 36.60& 33.50 & 40.77& 33.86& 38.37& 31.59& 79.26& 43.54& 43.77& 25.30 & 39.91& 41.11& 40.11& 52.69& 40.68\\
zh & 26.46& 30.99& 39.18& 28.61& 27.29& 26.73& 25.88& 25.06& 38.37& 49.18& 33.93& 29.18& 25.15& 32.85& 28.64& 31.18& 25.68& 38.65& 30.17& 35.42& 75.60& 33.53  \\
\bottomrule
\end{tabular}
\caption{Cross-lingual UABSA results on the Coursera dataset with non-English source languages.}
\label{tab:non-en-8}
\end{table*}

\begin{table*}[htbp]
\scriptsize
\setlength\tabcolsep{2pt}\centering
\begin{tabular}{ccccccccccccccccccccccc}
\toprule
&ar	&da	&de	&en	&es	&fr	&hi	&hr	&id	&ja	&ko	&nl	&pt	&ru	&sk	&sv	&sw	&th	&tr	&vi	&zh &avg \\
\midrule
ar & 50.99& 27.53& 29.27& 31.17& 29.72& 31.33& 27.06& 23.42& 28.16& 26.38& 25.32& 28.32& 29.41& 28.96& 30.36& 29.27& 24.37& 30.38& 26.42& 27.22& 30.78& 29.33\\
fr & 25.37& 32.42& 36.41& 38.24& 38.38& 47.38& 27.06& 22.59& 31.67& 30.85& 24.66& 36.79& 37.23& 34.45& 32.15& 33.98& 25.16& 32.44& 25.16& 29.11& 33.10 & 32.12\\
ko & 31.30 & 36.64& 39.46& 27.52& 36.22& 30.83& 33.14& 23.64& 42.14& 41.96& 47.44& 41.18& 36.21& 45.85& 37.07& 35.60 & 25.13& 23.10 & 32.45& 37.51& 48.79& 35.87\\
ru & 23.10 & 23.89& 25.00& 26.11& 25.71& 25.00& 23.26& 23.73& 24.37& 24.05& 23.73& 24.68& 26.27& 34.44& 26.42& 26.11& 22.78& 25.95& 23.73& 24.37& 25.95& 25.17\\
zh & 23.73& 25.22& 33.10 & 30.22& 30.90 & 27.21& 24.05& 17.88& 37.42& 36.24& 23.96& 27.05& 22.02& 28.74& 22.63& 29.55& 18.01& 27.53& 21.19& 28.73& 63.15& 28.50 \\
\bottomrule
\end{tabular}
\caption{Cross-lingual TASD results on the Food dataset with non-English source languages.}
\label{tab:non-en-5}
\end{table*}

\begin{table*}[htbp]
\scriptsize
\setlength\tabcolsep{2pt}\centering
\begin{tabular}{ccccccccccccccccccccccc}
\toprule
&ar	&da	&de	&en	&es	&fr	&hi	&hr	&id	&ja	&ko	&nl	&pt	&ru	&sk	&sv	&sw	&th	&tr	&vi	&zh &avg \\
\midrule
ar & 67.84& 35.66& 33.99& 34.34& 31.96& 36.60 & 38.93& 36.49& 33.07& 32.44& 33.23& 34.70 & 34.12& 35.67& 32.91& 36.72& 34.41& 33.81& 37.73& 35.00& 33.51& 36.34\\
fr & 40.59& 57.07& 56.33& 53.08& 58.65& 75.75& 46.48& 36.55& 55.96& 45.00& 36.65& 50.19& 52.63& 56.59& 54.97& 55.43& 39.56& 46.98& 46.56& 51.61& 46.40 & 50.62\\
ko & 33.70 & 37.26& 36.22& 36.97& 33.23& 34.02& 31.34& 30.85& 34.80 & 39.22& 56.20 & 33.83& 33.44& 34.93& 33.97& 35.49& 28.48& 31.80 & 33.46& 34.81& 43.51& 35.60\\
ru & 34.91& 40.28& 42.90 & 42.45& 38.56& 38.12& 37.50 & 30.01& 39.84& 37.03& 33.83& 39.47& 43.00& 51.22& 38.88& 40.82& 34.25& 42.37& 38.11& 38.45& 41.67& 39.22\\
zh & 33.54& 44.07& 41.65& 43.00& 41.86& 42.32& 38.48& 30.65& 52.63& 46.59& 47.79& 43.07& 42.93& 38.98& 36.06& 45.00& 32.91& 39.08& 35.11& 42.34& 73.73& 42.47 \\
\bottomrule
\end{tabular}
\caption{Cross-lingual UABSA results on the Food dataset with non-English source languages.}
\label{tab:non-en-6}
\end{table*}

\begin{table*}[htbp]
\scriptsize
\setlength\tabcolsep{2pt}\centering
\begin{tabular}{ccccccccccccccccccccccc}
\toprule
&ar	&da	&de	&en	&es	&fr	&hi	&hr	&id	&ja	&ko	&nl	&pt	&ru	&sk	&sv	&sw	&th	&tr	&vi	&zh &avg \\
\midrule
ar & 68.33& 25.03& 31.10 & 18.68& 21.15& 20.59& 30.7 & 26.15& 30.26& 32.05& 21.22& 20.90 & 29.52& 32.98& 32.60 & 24.71& 21.90 & 32.88& 19.89& 20.51& 33.15& 28.30  \\
fr & 10.06& 16.49& 20.88& 21.88& 17.36& 46.78& 10.11& 10.67& 11.29& 13.21& 10.08& 19.62& 15.27& 11.79& 18.55& 15.51& 12.85& 14.62& 18.89& 11.32& 12.91& 16.20\\
ko & 25.70 & 17.51& 35.02& 15.18& 29.07& 24.52& 27.86& 13.22& 21.08& 36.35& 73.15& 24.85& 36.15& 30.54& 24.95& 23.23& 12.55& 20.69& 18.36& 21.16& 37.95& 27.10\\
ru & 39.59& 32.41& 38.88& 24.23& 41.22& 34.31& 33.28& 41.53& 33.04& 40.08& 25.21& 33.19& 44.22& 77.28& 39.01& 26.23& 26.86& 31.16& 22.76& 25.96& 41.44& 35.80\\
zh & 30.97& 20.70 & 40.59& 21.02& 27.83& 32.86& 30.08& 26.43& 23.36& 32.00& 22.70 & 31.30 & 32.43& 29.44& 28.26& 31.34& 15.15& 24.68& 17.79& 22.61& 80.85& 29.64 \\
\bottomrule
\end{tabular}
\caption{Cross-lingual TASD results on the Hotel dataset with non-English source languages.}
\label{tab:non-en-13}
\end{table*}

\begin{table*}[htbp]
\scriptsize
\setlength\tabcolsep{2pt}\centering
\begin{tabular}{ccccccccccccccccccccccc}
\toprule
&ar	&da	&de	&en	&es	&fr	&hi	&hr	&id	&ja	&ko	&nl	&pt	&ru	&sk	&sv	&sw	&th	&tr	&vi	&zh &avg \\
\midrule
ar & 59.67& 19.42& 20.87& 19.44& 17.60 & 17.46& 23.63& 21.23& 20.18& 29.05& 19.63& 17.46& 18.48& 18.82& 20.24& 20.98& 22.17& 25.35& 21.83& 21.37& 19.90 & 22.62\\
fr & 28.62& 43.55& 52.33& 49.45& 48.77& 70.78& 42.03& 34.51& 32.27& 34.77& 37.42& 50.11& 42.40 & 42.54& 42.68& 35.12& 32.11& 34.54& 28.82& 31.54& 41.33& 40.75\\
ko & 31.18& 56.88& 40.33& 42.33& 46.10 & 44.81& 47.33& 49.58& 45.90 & 50.16& 70.76& 48.85& 49.12& 50.50 & 65.74& 50.34& 35.70 & 34.86& 47.09& 40.07& 66.13& 48.27\\
ru & 32.33& 57.16& 48.55& 32.78& 58.30 & 54.53& 55.32& 55.60 & 52.33& 50.97& 53.08& 48.70 & 54.60 & 71.70 & 57.80 & 54.24& 35.13& 39.86& 56.70 & 46.96& 59.10 & 51.23\\
zh & 28.12& 22.90 & 30.42& 27.99& 23.20 & 33.63& 32.44& 28.26& 26.67& 37.71& 29.14& 28.36& 33.49& 32.49& 30.50 & 33.41& 25.11& 31.92& 23.27& 23.66& 62.78& 30.74  \\
\bottomrule
\end{tabular}
\caption{Cross-lingual UABSA results on the Hotel dataset with non-English source languages.}
\label{tab:non-en-14}
\end{table*}

\begin{table*}[htbp]
\scriptsize
\setlength\tabcolsep{2pt}\centering
\begin{tabular}{ccccccccccccccccccccccc}
\toprule
&ar	&da	&de	&en	&es	&fr	&hi	&hr	&id	&ja	&ko	&nl	&pt	&ru	&sk	&sv	&sw	&th	&tr	&vi	&zh &avg \\
\midrule
ar & 38.23& 27.40 & 25.62& 30.57& 24.43& 23.08& 25.49& 15.28& 31.66& 27.42& 21.92& 27.64& 21.08& 22.4 & 30.08& 27.32& 18.95& 29.09& 24.02& 23.23& 27.95& 25.85\\
fr & 23.70 & 27.15& 27.76& 32.91& 30.59& 37.33& 19.68& 16.50 & 27.54& 25.11& 22.49& 29.53& 30.39& 24.53& 28.61& 26.81& 13.88& 28.36& 22.91& 21.25& 27.21& 25.92\\
ko & 23.59& 26.82& 30.26& 28.85& 24.33& 27.44& 20.76& 13.36& 29.24& 28.90 & 42.75& 31.02& 20.15& 20.46& 28.55& 27.47& 17.63& 24.54& 21.80 & 22.92& 29.59& 25.73 \\
ru & 19.52& 24.61& 28.59& 27.07& 21.24& 23.70 & 23.51& 13.61& 23.14& 25.31& 17.42& 23.32& 21.58& 39.07& 22.85& 26.18& 13.73& 26.74& 18.32& 21.29& 25.62& 23.16\\
zh & 22.52& 22.22& 26.53& 27.30 & 27.34& 20.58& 22.66& 12.21& 24.68& 28.08& 21.86& 25.93& 20.78& 19.26& 24.53& 28.61& 11.68& 25.31& 18.23& 17.46& 41.35& 23.29 \\
\bottomrule
\end{tabular}
\caption{Cross-lingual TASD results on the Laptop dataset with non-English source languages.}
\label{tab:non-en-3}
\end{table*}

\begin{table*}[htbp]
\scriptsize
\setlength\tabcolsep{2pt}\centering
\begin{tabular}{ccccccccccccccccccccccc}
\toprule
&ar	&da	&de	&en	&es	&fr	&hi	&hr	&id	&ja	&ko	&nl	&pt	&ru	&sk	&sv	&sw	&th	&tr	&vi	&zh &avg \\
\midrule
ar & 69.01& 47.26& 41.45& 49.80 & 48.71& 45.26& 40.17& 30.87& 52.78& 51.69& 40.00& 50.80 & 40.08& 38.65& 43.56& 52.58& 33.67& 52.60 & 38.29& 35.74& 61.20 & 45.91\\
fr & 39.61& 47.58& 52.59& 59.22& 40.38& 65.09& 35.48& 30.45& 44.77& 44.19& 39.65& 55.33& 53.68& 37.97& 41.54& 50.72& 30.61& 42.19& 40.62& 37.68& 52.63& 44.86\\
ko & 33.48& 41.18& 49.72& 46.20 & 41.34& 40.52& 40.83& 29.72& 41.25& 46.65& 68.45& 46.01& 36.60 & 31.94& 40.06& 45.15& 31.78& 43.69& 37.40 & 40.68& 55.26& 42.28\\
ru & 39.85& 55.03& 50.79& 47.87& 48.22& 47.86& 51.39& 33.5 & 48.96& 48.69& 39.71& 50.81& 50.86& 74.21& 46.62& 54.11& 34.93& 49.7 & 43.13& 43.74& 59.86& 48.56\\
zh & 32.73& 39.04& 47.45& 45.98& 39.97& 41.15& 41.62& 25.88& 42.71& 46.59& 43.54& 44.31& 37.11& 31.88& 40.09& 43.70 & 28.34& 44.59& 36.69& 37.74& 72.67& 41.13 \\
\bottomrule
\end{tabular}
\caption{Cross-lingual UABSA results on the Laptop dataset with non-English source languages.}
\label{tab:non-en-4}
\end{table*}

\begin{table*}[htbp]
\scriptsize
\setlength\tabcolsep{2pt}\centering
\begin{tabular}{ccccccccccccccccccccccc}
\toprule
&ar	&da	&de	&en	&es	&fr	&hi	&hr	&id	&ja	&ko	&nl	&pt	&ru	&sk	&sv	&sw	&th	&tr	&vi	&zh &avg \\
\midrule
ar & 54.39& 11.57& 13.44& 14.34& 14.76& 16.68& 10.12& 13.51& 12.59& 15.17& 10.32& 11.76& 13.01& 13.04& 16.22& 10.05& 12.43& 14.09& 11.78& 11.22& 22.34& 15.37\\
fr & 11.02& 16.57& 16.93& 14.48& 19.82& 43.44& 10.33& 12.32& 11.91& 10.31& 11.01& 16.80 & 17.16& 14.22& 15.66& 15.29& 10.77& 12.03& 12.57& 10.124 & 13.22& 15.05\\
ko & 14.02& 12.48& 15.39& 12.29& 12.29& 15.01& 11.32& 13.23& 14.97& 20.47& 55.51& 13.71& 11.70 & 10.07& 12.9 & 12.31& 10.54& 13.47& 12.14& 16.45& 17.05& 15.59\\
ru & 13.14& 11.15& 16.18& 13.25& 13.46& 19.54& 10.77& 11.02& 14.46& 19.76& 10.08& 11.05& 14.21& 63.16& 11.33& 10.47& 11.86& 19.70 & 11.71& 15.12& 24.10 & 16.45\\
zh & 19.25& 21.57& 24.75& 20.29& 20.26& 24.89& 13.57& 16.00& 20.48& 20.79& 12.69& 22.74& 15.30 & 24.17& 21.73& 20.29& 13.47& 26.42& 17.79& 18.47& 62.78& 21.80\\
\bottomrule
\end{tabular}
\caption{Cross-lingual TASD results on the Phone dataset with non-English source languages.}
\label{tab:non-en-9}
\end{table*}

\begin{table*}[htbp]
\scriptsize
\setlength\tabcolsep{2pt}\centering
\begin{tabular}{ccccccccccccccccccccccc}
\toprule
&ar	&da	&de	&en	&es	&fr	&hi	&hr	&id	&ja	&ko	&nl	&pt	&ru	&sk	&sv	&sw	&th	&tr	&vi	&zh &avg \\
\midrule
ar & 78.30 & 15.14& 20.94& 30.69& 20.75& 19.71& 15.61& 25.24& 20.56& 32.87& 15.87& 16.48& 22.17& 19.54& 20.85& 16.05& 22.90 & 26.85& 15.13& 17.97& 34.19& 24.18\\
fr & 32.13& 35.60 & 41.88& 35.93& 55.22& 73.23& 34.78& 32.39& 35.19& 36.40 & 29.40 & 38.32& 55.18& 36.48& 41.21& 38.30 & 30.28& 42.38& 30.92& 25.58& 51.06& 39.61\\
ko & 17.55& 28.79& 33.15& 42.21& 27.46& 26.01& 17.71& 31.19& 24.93& 36.02& 74.73& 19.45& 25.29& 17.12& 21.42& 25.47& 21.13& 21.48& 20.50 & 26.94& 43.36& 28.66\\
ru & 38.55& 34.36& 47.82& 41.64& 35.98& 41.50 & 31.35& 23.55& 39.11& 40.38& 24.55& 37.27& 37.69& 76.74& 23.95& 36.95& 29.06& 40.78& 31.99& 39.09& 57.95& 39.16\\
zh & 28.26& 27.21& 35.29& 29.08& 21.25& 29.02& 17.56& 24.00& 23.32& 31.38& 15.51& 24.22& 21.08& 23.67& 25.64& 28.85& 21.18& 35.29& 18.12& 26.82& 77.71& 27.83 \\
\bottomrule
\end{tabular}
\caption{Cross-lingual UABSA results on the Phone dataset with non-English source languages.}
\label{tab:non-en-10}
\end{table*}

\begin{table*}[htbp]
\scriptsize
\setlength\tabcolsep{2pt}\centering
\begin{tabular}{ccccccccccccccccccccccc}
\toprule
&ar	&da	&de	&en	&es	&fr	&hi	&hr	&id	&ja	&ko	&nl	&pt	&ru	&sk	&sv	&sw	&th	&tr	&vi	&zh &avg \\
\midrule
ar& 63.60& 30.03& 26.53& 22.39& 27.06& 24.19& 21.87& 23.01& 40.21& 38.29& 21.88& 30.04& 33.67& 26.53& 27.76& 30.37& 23.70& 41.34& 28.93& 22.64& 43.51&30.84\\
fr& 35.77& 51.47& 47.01& 51.77& 49.03& 59.82& 37.41& 29.82& 47.49& 41.95& 34.59& 43.79& 49.70& 40.50& 44.86& 51.52& 32.90& 45.54& 39.70& 34.47& 52.21&43.87\\
ko& 31.15& 34.31& 41.41& 32.28& 43.63& 35.29& 37.63& 21.41& 36.40& 40.86& 66.02& 36.86& 40.62& 39.74& 36.50& 36.25& 24.02& 38.76& 29.10& 30.24& 49.76&37.25\\
ru& 35.97& 40.83& 40.63& 34.13& 35.76& 32.57& 36.49& 24.40& 37.58& 43.14& 32.22& 35.65& 42.34& 63.15& 33.00  & 45.70& 27.83& 48.07& 35.23& 31.79& 51.50&38.48\\
zh& 33.15& 35.42& 41.93& 40.58& 39.09& 34.08& 31.99& 26.24& 36.00  & 46.32& 29.38& 41.12& 43.58& 37.58& 34.93& 39.48& 26.00  & 43.44& 30.77& 34.02& 70.53&37.89 \\
\bottomrule
\end{tabular}
\caption{Cross-lingual TASD results on the Restaurant dataset with non-English source languages.}
\label{tab:non-en-1}
\end{table*}

\begin{table*}[htbp]
\scriptsize
\centering
\setlength\tabcolsep{2pt}
\begin{tabular}{ccccccccccccccccccccccc}
\toprule
&ar	&da	&de	&en	&es	&fr	&hi	&hr	&id	&ja	&ko	&nl	&pt	&ru	&sk	&sv	&sw	&th	&tr	&vi	&zh &avg \\
\midrule
ar & 73.92& 46.25& 40.88& 38.77& 35.49& 36.11& 38.59& 39.23& 46.95& 52.63& 32.11& 40.41& 39.22& 41.65& 41.09& 47.77& 39.53& 57.29& 36.29& 32.86& 52.85& 43.33\\
fr & 34.58& 57.25& 55.91& 54.78& 55.48& 78.20 & 34.21& 39.98& 58.18& 44.63& 29.95& 41.89& 60.85& 34.98& 51.85& 55.11& 45.52& 48.93& 47.51& 39.22& 59.15& 48.96 \\
ko & 40.93& 46.39& 50.73& 39.98& 47.57& 40.52& 40.31& 37.07& 44.76& 49.48& 73.94& 43.66& 48.27& 45.67& 42.29& 48.96& 32.20 & 48.03& 39.54& 35.72& 51.74& 45.13\\
ru & 40.38& 49.74& 53.07& 40.33& 42.33& 36.79& 43.52& 31.48& 45.82& 52.11& 40.02& 38.55& 47.55& 73.83& 42.04& 54.91& 36.22& 53.27& 46.84& 33.85& 59.71& 45.83\\
zh & 35.89& 39.98& 43.69& 42.66& 34.45& 35.12& 31.20 & 28.45& 38.16& 50.46& 34.15& 42.14& 42.11& 36.44& 37.29& 42.37& 33.52& 46.59& 34.86& 32.73& 78.96& 40.06 \\
\bottomrule
\end{tabular}
\caption{Cross-lingual UABSA results on the Restaurant dataset with non-English source languages.}
\label{tab:non-en-2}
\end{table*}

\begin{table*}[htbp]
\scriptsize
\setlength\tabcolsep{2pt}\centering
\begin{tabular}{ccccccccccccccccccccccc}
\toprule
&ar	&da	&de	&en	&es	&fr	&hi	&hr	&id	&ja	&ko	&nl	&pt	&ru	&sk	&sv	&sw	&th	&tr	&vi	&zh &avg \\
\midrule
ar & 32.54& 21.26& 23.61& 15.20 & 21.64& 22.66& 13.80 & 18.68& 21.22& 24.91& 20.18& 20.11& 21.28& 22.27& 20.48& 19.47& 14.26& 22.02& 16.59& 20.99& 23.67& 29.45\\
fr & 18.99& 25.47& 26.13& 15.98& 27.09& 38.14& 18.18& 22.35& 27.65& 24.95& 19.91& 25.23& 23.59& 19.45& 23.03& 24.44& 20.01& 25.74& 20.89& 28.52& 23.84& 23.79\\
ko & 10.27& 17.36& 24.26& 11.56& 19.44& 17.97& 12.72& 14.34& 18.15& 25.24& 38.22& 18.65& 16.76& 18.91& 17.85& 14.22& 14.14& 16.17& 16.01& 18.84& 20.98& 18.19\\
ru & 22.20 & 25.55& 31.54& 16.78& 30.20 & 27.02& 17.14& 23.33& 27.99& 23.57& 17.59& 26.76& 27.78& 40.65& 24.93& 29.96& 17.23& 23.17& 20.40 & 25.96& 25.09& 24.99\\
zh & 11.27& 23.24& 21.53& 14.34& 17.20 & 17.79& 13.39& 10.79& 21.08& 28.00& 18.67& 19.40 & 17.25& 19.94& 18.51& 18.93& 10.39& 21.62& 14.34& 15.46& 38.79& 18.66 \\
\bottomrule
\end{tabular}
\caption{Cross-lingual TASD results on the Sight dataset with non-English source languages.}
\label{tab:non-en-11}
\end{table*}

\begin{table*}[htbp]
\scriptsize
\setlength\tabcolsep{2pt}\centering
\begin{tabular}{ccccccccccccccccccccccc}
\toprule
&ar	&da	&de	&en	&es	&fr	&hi	&hr	&id	&ja	&ko	&nl	&pt	&ru	&sk	&sv	&sw	&th	&tr	&vi	&zh &avg \\
\midrule
ar & 39.16& 29.89& 31.25& 25.78& 28.36& 31.79& 21.82& 23.49& 27.91& 35.59& 30.05& 30.50 & 28.23& 31.51& 28.43& 27.24& 26.54& 32.16& 24.82& 29.59& 31.33& 29.31\\
fr & 24.53& 26.78& 30.36& 25.00& 28.04& 41.9 & 22.40 & 21.68& 27.82& 29.84& 22.60 & 28.47& 26.47& 26.91& 25.56& 23.54& 19.46& 30.77& 22.04& 28.45& 28.65& 26.73\\
ko & 14.40 & 17.87& 21.87& 16.74& 23.22& 19.64& 14.23& 15.33& 18.17& 19.50 & 40.43& 20.67& 19.61& 18.06& 18.38& 17.22& 14.98& 17.84& 16.80 & 19.89& 22.44& 19.39 \\
ru & 23.70 & 26.80 & 32.85& 28.65& 29.52& 28.99& 19.70 & 20.29& 24.66& 27.13& 19.54& 25.90 & 26.07& 39.63& 19.67& 27.69& 23.18& 29.94& 22.85& 22.27& 33.75& 26.32\\
zh & 20.86& 25.66& 29.86& 25.72& 24.75& 24.98& 22.14& 18.75& 25.90 & 37.25& 27.25& 29.70 & 25.55& 26.15& 25.55& 23.09& 18.82& 31.94& 19.72& 21.78& 42.42& 26.09  \\
\bottomrule
\end{tabular}
\caption{Cross-lingual UABSA results on the Sight dataset with non-English source languages.}
\label{tab:non-en-12}
\end{table*}


\begin{table*}[htbp]
	\centering  
	\subfigure[\textbf{TASD} Results]{  
		\begin{minipage}{.485\linewidth}
			\centering 
\setlength\tabcolsep{3pt}
\scriptsize
\begin{tabular}{c|ccccccc|c}
\toprule 
Lang. & Coursera & Food & Hotel & Laptop & Phone & Res. & Sight & Avg. \\
\midrule
ar       & 11.03   & 13.30  & 29.65 & 18.31 & 25.33 & 37.32 & 18.11& 21.87 \\
da       & 19.67  & 13.89 & 31.70  & 21.90  & 24.93 & 40.26 & 17.88& 24.32 \\
de       & 17.42  & 15.25 & 29.71 & 23.85 & 29.17 & 38.14 & 18.66& 24.60       \\
en       &19.86&15.56&29.57&23.43&25.68&44.88 &18.39& 25.34 \\
es       & 12.03  & 14.99 & 26.32 & 19.50  & 25.56 & 32.08 & 17.47& 21.14 \\
fr       & 13.88  & 12.91 & 23.24 & 19.36 & 21.51 & 31.17 & 17.57& 19.95 \\
hi       & 22.24  & 15.31 & 30.94 & 20.08 & 26.04 & 39.45 & 16.44& 24.36 \\
hr       & 18.94  & 11.75 & 29.31 & 18.27 & 21.98 & 35.86 & 17.46& 21.94 \\
id       & 20.94  & 15.79 & 29.04 & 21.10  & 25.30  & 42.18 & 16.96& 24.47 \\
ja       & 17.48  & 15.25 & 31.74 & 22.93 & 28.48 & 42.01 & 18.55& 25.21 \\
ko       & 16.98  & 11.52 & 27.89 & 19.05 & 25.67 & 36.59 & 17.87& 22.22 \\
nl       & 17.11  & 16.27 & 29.89 & 22.22 & 24.77 & 38.56 & 18.53& 23.91 \\
pt       & 21.90   & 15.14 & 32.92 & 23.78 & 26.22 & 38.71 & 18.81& 25.35 \\
ru       & 19.85  & 14.77 & 31.19 & 21.60  & 23.16 & 39.27 & 17.17& 23.86 \\
sk       & 22.69  & 14.07 & 32.66 & 22.85 & 23.31 & 39.90  & 18.55& 24.86 \\
sv       & 21.47  & 15.56 & 27.62 & 19.85 & 23.93 & 39.34 & 17.24& 23.57 \\
sw       & 18.38  & 13.07 & 22.98 & 15.32 & 22.59 & 32.95 & 15.04& 20.05 \\
th       & 22.07  & 14.30  & 28.72 & 20.99 & 26.05 & 35.27 & 17.33& 23.53 \\
tr       & 20.17  & 13.45 & 29.48 & 18.76 & 24.24 & 36.81 & 16.55& 22.78      \\
vi       & 21.11  & 11.93 & 29.05 & 20.40  & 22.52 & 36.93 & 17.35& 22.76 \\
zh       & 17.42  & 14.22 & 29.95 & 24.49 & 25.20  & 37.78 & 16.98& 23.72  \\ 
\bottomrule
\end{tabular}

		\end{minipage}
	}
  \subfigure[\textbf{UABSA} Results]{ 
		\begin{minipage}{.485\linewidth}
			\centering 
\setlength\tabcolsep{3pt}
\scriptsize
\begin{tabular}{c|ccccccc|c}
\toprule 
Lang. & Coursera & Food & Hotel & Laptop & Phone & Res. & Sight & Avg. \\
\midrule
ar       & 31.71  & 22.49 & 45.23 & 48.23 & 32.88 & 48.98  & 31.25& 37.25 \\
da       & 39.20  & 25.83 & 49.55 & 54.08 & 35.85 & 52.42 & 33.78 & 41.53      \\
de       & 33.04 & 26.41 & 46.59 & 54.60  & 39.80  & 52.29 & 36.21 & 41.28 \\
en       & 31.19 & 21.37 & 37.38 & 47.71 & 32.32 & 45.88 & 31.62&35.35 \\
es       & 24.87 & 24.68 & 44.85 & 46.95 & 34.77 & 50.23 & 33.46 & 37.12 \\
fr       & 28.11  & 23.98 & 43.38 & 46.15 & 31.15 & 46.06 & 31.40 & 35.75 \\
hi       & 38.63 & 26.58 & 50.30  & 54.49 & 36.11 & 53.19 & 34.18 & 41.93 \\
hr       & 36.31 & 22.22 & 45.66 & 44.36 & 33.63 & 49.18 & 33.46 & 37.83 \\
id       & 34.77 & 26.79 & 47.22 & 53.54 & 35.45 & 50.23 & 33.12 & 40.16      \\
ja       & 31.78 & 27.31 & 48.79 & 55.94 & 37.66 & 54.67 & 34.92 & 41.58 \\
ko       & 30.29  & 23.72 & 43.31 & 52.15 & 36.93 & 49.84 & 32.30 & 38.36 \\
nl       & 35.22 & 27.49 & 48.26 & 52.53 & 35.55 & 52.35 & 35.72 & 41.02 \\
pt       & 41.62 & 26.48 & 50.27 & 54.29 & 37.27 & 54.98 & 35.37 & 42.90 \\
ru       & 32.42 & 22.77 & 44.81 & 51.37 & 31.79 & 49.18 & 31.95 & 37.76 \\
sk       & 37.72 & 24.34 & 50.76 & 53.22 & 34.46 & 52.92 & 36.61 & 41.43 \\
sv       & 39.06 & 26.32 & 46.61 & 51.99 & 38.00    & 51.91 & 35.38 & 41.32 \\
sw       & 30.12 & 22.94 & 38.41 & 44.29 & 31.88 & 41.80 & 31.45  & 34.41 \\
th       & 37.76  & 23.67 & 46.97 & 48.44 & 34.91 & 50.23 & 34.20 & 39.45 \\
tr       & 34.94 & 21.99 & 45.17 & 48.22 & 34.66 & 47.58 & 32.43 & 37.86 \\
vi       & 35.43 & 23.79 & 43.92 & 45.02 & 33.17 & 47.93 & 33.16 & 37.49 \\
zh       & 33.28 & 24.59 & 45.50  & 53.58 & 33.56 & 51.30 & 32.55  & 39.19  \\ 
\bottomrule
\end{tabular}
  \end{minipage}
	}
\caption{Zero-shot results on the M-ABSA dataset with Gemma-2 model.} 
\label{tab:main_result_gemma}
\end{table*}


\begin{table*}[htbp]
	\centering  
	\subfigure[\textbf{TASD} Results]{  
		\begin{minipage}{.485\linewidth}
			\centering 
\setlength\tabcolsep{3pt}
\scriptsize
\begin{tabular}{c|ccccccc|c}
\toprule 
Lang. & Coursera & Food & Hotel & Laptop & Phone & Res. & Sight & Avg. \\
\midrule
ar       & 9.56     & 6.42  & 12.88 & 9.53  & 9.79  & 6.63  & 7.70& 8.93 \\
da       & 12.62  & 9.62  & 23.93 & 15.65 & 12.46 & 25.54 & 10.08& 15.7       \\
de       & 14.68  & 12.60  & 28.28 & 16.07 & 13.92 & 25.54 & 14.97& 18.01 \\
en       &11.82  & 14.33 & 29.12 & 19.59 & 13.62 & 38.22& 14.05&20.11 \\
es       & 12.20   & 9.96  & 24.29 & 14.50  & 11.50  & 28.02 & 13.41& 16.27 \\
fr       & 10.95  & 8.58  & 21.01 & 14.44 & 11.83 & 22.42 & 10.19& 14.20 \\
hi       & 11.02  & 7.40   & 20.90  & 16.62 & 11.80  & 13.37 & 11.43& 13.22      \\
hr       & 8.55  & 7.92  & 25.98 & 12.82 & 11.78 & 26.93  & 11.32& 15.04 \\
id       & 10.73  & 11.47 & 25.37 & 14.00    & 11.45 & 34.27 & 11.88& 17.02 \\
ja       & 6.84    & 6.36  & 10.13 & 13.91 & 9.49  & 4.56 & 3.35 & 7.81 \\
ko       & 6.13    & 4.61  & 12.90  & 11.76 & 11.39 & 4.10  & 6.30  & 8.17       \\
nl       & 10.62  & 12.05 & 26.81 & 15.51 & 12.49 & 30.57 & 13.94& 17.43 \\
pt       & 14.78  & 12.19 & 27.03 & 14.59 & 12.22 & 29.27 & 12.77& 17.55      \\
ru       & 13.33  & 7.92  & 24.21 & 16.02 & 11.14 & 13.37 & 11.42& 13.92 \\
sk       & 10.59  & 7.17  & 19.91 & 13.6  & 12.68 & 24.02 & 11.35& 14.19 \\
sv       & 11.47  & 9.58  & 23.85 & 12.84 & 12.84 & 26.48 & 11.95& 15.57 \\
sw       & 7.61    & 8.38  & 12.66 & 6.91  & 10.54 & 22.22 & 9.98& 11.19 \\
th       & 8.01    & 5.77  & 13.30  & 12.14 & 12.34 & 8.56 & 9.54 & 9.95 \\
tr       & 9.94   & 8.27  & 23.63 & 13.83 & 11.59 & 24.04 & 11.19& 14.64 \\
vi       & 10.35  & 7.13  & 24.77 & 15.94 & 13.45 & 19.30 & 11.88 & 14.69 \\
zh       & 9.88   & 9.15  & 7.05  & 8.98  & 9.34  & 10.63  & 8.27& 9.04  \\
\bottomrule
\end{tabular}
		\end{minipage}
	}
  \subfigure[\textbf{UABSA} Results]{ 
		\begin{minipage}{.485\linewidth}
			\centering 
\setlength\tabcolsep{3pt}
\scriptsize
\begin{tabular}{c|ccccccc|c}
\toprule 
Lang. & Coursera & Food & Hotel & Laptop & Phone & Res. & Sight & Avg. \\
\midrule
ar       & 21.32   & 16.63 & 31.86 & 40.67 & 26.07 & 38.16 & 23.33& 28.29 \\
da       & 29.70  & 19.40  & 37.01 & 39.56 & 29.67 & 37.35  & 27.55& 31.46 \\
de       & 28.64  & 18.93 & 37.81 & 46.43 & 30.11 & 39.32 & 31.94& 33.31 \\
en       & 31.19 & 21.37 & 37.38 & 47.71 & 32.32 & 45.88 & 31.62&35.35\\
es       & 21.74 & 15.38 & 32.61 & 34.02 & 25.29 & 34.73  & 28.88& 27.52 \\
fr       & 22.49  & 15.12 & 25.32 & 32.40  & 25.18 & 28.07 & 22.70 & 24.47 \\
hi       & 28.05 & 16.45 & 36.71 & 45.97 & 29.01 & 39.71 & 29.47 & 32.20 \\
hr       & 25.49   & 13.95 & 35.88 & 34.42 & 24.37 & 37.71& 26.70 & 28.36      \\
id       & 27.09  & 18.26 & 36.21 & 41.85 & 26.96 & 39.74 & 28.89& 31.29 \\
ja       & 18.73  & 15.59 & 25.53 & 39.44 & 24.92 & 28.80 & 21.56 & 24.94 \\
ko       & 20.37  & 16.05 & 29.18 & 42.05 & 28.19 & 32.01& 24.03 & 27.41 \\
nl       & 26.86 & 17.68 & 33.75 & 41.24 & 28.85 & 38.98  & 29.53& 30.98 \\
pt       & 33.94 & 20.70  & 40.72 & 43.33 & 29.49 & 43.04  & 30.35& 34.51      \\
ru       & 29.60  & 18.22 & 36.32 & 45.38 & 27.53 & 42.77  & 29.47& 32.76 \\
sk       & 24.92 & 14.90  & 35.40  & 37.42 & 27.18 & 36.00   & 27.55  & 29.05 \\
sv       & 25.67  & 17.81 & 35.12 & 36.40  & 30.58 & 37.99& 29.17 & 30.39 \\
sw       & 19.13  & 13.76 & 21.78 & 29.33 & 24.29 & 32.31& 23.78 & 23.48 \\
th       & 26.13   & 17.28 & 32.01 & 39.74 & 31.26 & 39.84& 30.30 & 30.94 \\
tr       & 25.24 & 15.35 & 32.99 & 37.11 & 26.71 & 35.47 & 25.26 & 28.30 \\
vi       & 27.95  & 17.79 & 36.99 & 40.31 & 27.17 & 38.34 & 30.43& 31.28 \\
zh       & 18.38  & 10.53 & 17.30  & 30.00    & 23.42 & 17.40  & 21.40 & 19.78  \\
\bottomrule
\end{tabular}
  \end{minipage}
	}
\caption{Zero-shot results on the M-ABSA dataset with Llama-3.1 model.} 
\label{tab:main_result_llama}
\end{table*}


\begin{table*}[htbp]
	\centering  
	\subfigure[\textbf{TASD} Results]{  
		\begin{minipage}{.485\linewidth}
			\centering 
\setlength\tabcolsep{3pt}
\scriptsize
\begin{tabular}{c|ccccccc|c}
\toprule 
Lang. & Coursera & Food & Hotel & Laptop & Phone & Res. & Sight & Avg. \\
\midrule
ar       & 9.37   & 1.79 & 4.94  & 2.55 & 2.58 & 11.55 & 6.32& 5.59 \\
da       & 11.28  & 3.58 & 11.26 & 3.56 & 2.61 & 17.39 & 6.46& 8.02       \\
de       & 11.63  & 4.12 & 10.94 & 8.00    & 6.12 & 16.76 & 9.68& 9.61 \\
en       & 16.13  & 6.77 & 19.11 & 3.45 & 2.71 & 27.03& 8.64& 11.98\\
es       & 6.64    & 5.60  & 14.24 & 3.96 & 2.02 & 19.92 & 7.50& 8.55 \\
fr       & 4.44   & 4.72 & 9.77  & 3.75 & 1.05 & 15.95& 6.31 & 6.57       \\
hi       & 5.46   & 3.33 & 6.41  & 1.73 & 2.95 & 10.53 & 6.65& 5.29 \\
hr       & 6.53   & 3.29 & 10.73 & 3.94 & 2.61 & 17.45 & 6.48& 7.29       \\
id       & 9.19   & 5.68 & 11.23 & 4.47 & 2.31 & 18.02& 8.28 & 8.45 \\
ja       & 8.15   & 3.97 & 11.22 & 11.80 & 4.55 & 18.91 & 9.69& 9.76 \\
ko       & 5.33   & 3.63 & 7.05  & 4.89 & 1.76 & 14.81 & 9.67& 6.73 \\
nl       & 8.14   & 5.44 & 11.77 & 6.10  & 2.89 & 16.44 & 7.81& 8.37       \\
pt       & 8.20    & 4.68 & 16.40  & 5.12 & 2.72 & 22.01 & 7.73& 9.50 \\
ru       & 3.96   & 4.89 & 6.32  & 3.29 & 1.07 & 18.59& 8.26 & 6.63 \\
sk       & 7.33   & 2.92 & 9.85  & 4.03 & 3.09 & 15.92& 6.77 & 7.13       \\
sv       & 9.74   & 3.25 & 9.19  & 3.65 & 2.74 & 17.66 & 6.75& 7.57 \\
sw       & 2.79   & 1.16 & 2.99  & 1.00    & 1.15 & 3.83 & 4.53 & 2.49 \\
th       & 6.63   & 4.41 & 8.44  & 2.60  & 1.30  & 13.96 & 9.03& 6.62 \\
tr       & 6.71  & 3.30  & 7.52  & 4.07 & 3.74 & 13.91  & 7.63& 6.70 \\
vi       & 5.58  & 3.43 & 8.67  & 3.24 & 1.70  & 14.22  & 7.83& 6.38 \\
zh       & 5.10   & 2.78 & 9.27  & 3.32 & 2.33 & 17.01  & 8.78& 6.94\\
\bottomrule
\end{tabular}
		\end{minipage}
	}
  \subfigure[\textbf{UABSA} Results]{ 
		\begin{minipage}{.485\linewidth}
			\centering 
\setlength\tabcolsep{3pt}
\scriptsize
\begin{tabular}{c|ccccccc|c}
\toprule 
Lang. & Coursera & Food & Hotel & Laptop & Phone & Res. & Sight & Avg. \\
\midrule
ar       & 21.37  & 9.42  & 16.70  & 18.31 & 9.94  & 26.68 & 18.74 & 17.31 \\
da       & 29.25 & 14.97 & 36.13 & 32.56 & 24.51 & 34.27 & 21.99 & 27.67 \\
de       & 29.68 & 19.73 & 38.02 & 44.75 & 29.46 & 42.28 & 26.47 & 32.91 \\
en       & 29.44 & 23.43 & 37.85 & 46.75 & 28.93 & 47.51 & 28.24&34.60 \\
es       & 19.00   & 18.26 & 37.81 & 35.14 & 24.73 & 41.77  & 24.95 & 28.81 \\
fr       & 23.40  & 18.57 & 29.26 & 32.19 & 20.98 & 35.91  & 18.82& 25.59      \\
hi       & 24.96  & 14.89 & 30.80  & 30.39 & 18.75 & 27.76& 16.63 & 23.45 \\
hr       & 23.61 & 15.01 & 33.94 & 31.46 & 22.32 & 35.68 & 22.17 & 26.31 \\
id       & 29.04 & 20.62 & 35.05 & 38.57 & 24.14 & 40.52 & 23.85 & 30.26 \\
ja       & 28.47 & 21.05 & 42.55 & 44.31 & 30.36 & 47.00  & 29.59   & 34.76 \\
ko       & 25.87  & 16.23 & 30.88 & 35.11 & 24.90  & 38.85 & 24.44& 28.04      \\
nl       & 31.03  & 18.88 & 36.43 & 42.40  & 22.97 & 39.14& 24.29 & 30.73 \\
pt       & 33.33    & 19.02 & 41.20  & 37.63 & 27.03 & 44.92 & 27.00 & 32.88 \\
ru       & 24.56  & 16.67 & 31.62 & 36.80  & 22.85 & 36.96& 23.87 & 27.62 \\
sk       & 22.55  & 13.18 & 32.42 & 32.45 & 19.16 & 33.93& 20.77 & 24.92 \\
sv       & 26.06 & 15.52 & 35.89 & 31.75 & 22.09 & 34.52 & 20.09 & 26.56      \\
sw       & 10.73 & 8.66  & 16.62 & 16.67 & 11.69 & 17.37 & 13.08 & 13.55 \\
th       & 29.15  & 19.85 & 33.91 & 30.79 & 21.75 & 34.85 & 23.80 & 27.73 \\
tr       & 27.75 & 16.11 & 31.53 & 31.16 & 21.23 & 32.62 & 19.18 & 25.65 \\
vi       & 27.66 & 14.08 & 30.62 & 31.27 & 17.62 & 31.33 & 25.79 & 25.48 \\
zh       & 26.62 & 16.93 & 38.47 & 42.31 & 24.94 & 41.97 & 24.97 & 30.89\\
\bottomrule
\end{tabular}
  \end{minipage}
	}
\caption{Zero-shot results on the M-ABSA dataset with Mistral model.} 
\label{tab:main_result_mistral}
\end{table*}


\begin{table*}[htbp]
	\centering  
	\subfigure[\textbf{TASD} Results]{  
		\begin{minipage}{.485\linewidth}
			\centering 
\setlength\tabcolsep{3pt}
\scriptsize
\begin{tabular}{c|ccccccc|c}
\toprule 
Lang. & Coursera & Food & Hotel & Laptop & Phone & Res. & Sight & Avg. \\
\midrule
ar       & 12.41  & 4.08 & 11.53 & 7.79  & 6.34  & 25.20 & 12.24  & 11.37 \\
da       & 17.70  & 7.85 & 27.49 & 15.08 & 14.87 & 39.14 & 13.01 & 19.31 \\
de       & 15.48 & 9.49 & 27.10  & 18.29 & 17.16 & 41.29 & 18.32 & 21.02 \\
en       & 17.48 & 9.90 & 28.47 & 21.61 & 18.21 & 47.66 & 15.99&22.76  \\
es       & 13.36 & 7.41 & 24.35 & 15.19 & 11.75 & 39.95 & 15.03 & 18.15 \\
fr       & 14.21 & 8.45 & 24.43 & 18.04 & 11.31 & 35.15 & 14.97 & 18.08      \\
hi       & 2.93  & 8.07 & 9.72  & 10.42 & 4.49  & 36.21 & 11.55 & 11.91 \\
hr       & 14.76 & 6.55 & 26.07 & 15.22 & 13.45 & 34.74 & 12.57 & 17.62 \\
id       & 17.80  & 9.27 & 24.63 & 17.91 & 12.50  & 36.80  & 15.29 & 19.17 \\
ja       & 11.19 & 7.15 & 25.54 & 19.11 & 17.46 & 45.21 & 17.95 & 20.52 \\
ko       & 7.87 & 6.23 & 22.03 & 14.94 & 14.15 & 39.15  & 16.22 & 17.23 \\
nl       & 19.13 & 8.53 & 25.95 & 18.29 & 14.88 & 41.73 & 14.82 & 20.48 \\
pt       & 20.24 & 7.87 & 26.97 & 16.01 & 15.21 & 41.64 & 13.91 & 20.26 \\
ru       & 4.95 & 7.45 & 18.19 & 12.88 & 5.43  & 37.02  & 11.37 & 13.90 \\
sk       & 17.59 & 7.57 & 23.61 & 15.68 & 10.75 & 35.85 & 14.49 & 17.93 \\
sv       & 16.48 & 9.52 & 28.87 & 15.08 & 15.30  & 41.04 & 15.52 & 20.26 \\
sw       & 4.21 & 3.29 & 6.02  & 5.75  & 3.77  & 19.44  & 10.98 & 7.64 \\
th       & 6.48 & 9.89 & 23.04 & 14.91 & 11.78 & 37.04  & 13.67 & 16.69 \\
tr       & 14.48& 7.70  & 23.38 & 14.71 & 10.39 & 37.38 & 16.45  & 17.78 \\
vi       & 7.70  & 7.73 & 19.91 & 12.28 & 9.17  & 36.72  & 16.22 & 15.68 \\
zh       & 11.02 & 9.30  & 26.86 & 18.44 & 18.66 & 43.45 & 15.79 & 20.50 \\
\bottomrule
\end{tabular}
		\end{minipage}
	}
  \subfigure[\textbf{UABSA} Results]{ 
		\begin{minipage}{.485\linewidth}
			\centering 
\setlength\tabcolsep{3pt}
\scriptsize
\begin{tabular}{c|ccccccc|c}
\toprule 
Lang. & Coursera & Food & Hotel & Laptop & Phone & Res. & Sight & Avg. \\
\midrule
ar       & 21.59  & 17.48 & 27.81 & 39.94 & 24.07 & 38.46 & 19.26 & 26.94 \\
da       & 29.51 & 17.83 & 44.61 & 41.69 & 30.18 & 45.37 & 26.27 & 33.64 \\
de       & 33.60 & 21.42 & 43.74 & 48.38 & 33.08 & 45.92  & 30.74 & 36.70 \\
en       & 33.50& 21.97 & 41.44 & 51.42 & 34.09 & 54.01& 32.13&38.37 \\
es       & 19.61 & 18.94 & 41.79 & 42.72 & 30.67 & 45.98 & 28.04 & 32.54 \\
fr       & 24.53 & 18.98 & 39.64 & 39.94 & 27.83 & 39.87 & 23.15 & 30.56 \\
hi       & 31.68 & 20.45 & 38.61 & 45.90  & 27.48 & 43.16 & 23.29 & 32.94 \\
hr       & 27.88 & 16.88 & 39.41 & 43.67 & 27.45 & 42.15 & 25.13 & 31.80 \\
id       & 31.60  & 22.33 & 41.39 & 46.74 & 28.13 & 46.47   & 26.70& 34.77 \\
ja       & 27.97 & 23.41 & 43.86 & 51.76 & 37.90  & 49.22  & 28.65& 37.54 \\
ko       & 25.53  & 17.55 & 41.38 & 46.56 & 31.65 & 45.87 & 26.29& 33.55 \\
nl       & 31.98  & 19.20  & 42.41 & 46.37 & 28.49 & 46.78 & 28.63& 34.84 \\
pt       & 32.67   & 20.97 & 46.60  & 48.13 & 33.19 & 51.04& 30.20 & 37.54 \\
ru       & 32.41  & 17.43 & 40.87 & 45.97 & 30.62 & 48.62 & 24.98& 34.41 \\
sk       & 31.37 & 19.22 & 43.05 & 43.48 & 26.36 & 42.13  & 27.11& 33.25 \\
sv       & 31.33 & 18.96 & 45.97 & 43.66 & 28.17 & 47.92 & 27.71 & 34.82 \\
sw       & 13.67  & 11.13 & 19.04 & 23.49 & 15.50  & 22.84 & 16.52& 17.46 \\
th       & 37.11  & 22.02 & 43.28 & 47.80  & 35.38 & 48.82 & 26.39& 37.26 \\
tr       & 25.31  & 17.67 & 39.37 & 41.08 & 25.81 & 41.00  & 21.80   & 30.29 \\
vi       & 32.51  & 20.44 & 40.95 & 43.29 & 28.09 & 44.13& 28.34 & 33.96 \\
zh       & 27.30  & 18.87 & 40.34 & 45.29 & 33.52 & 47.02  & 28.72& 34.44 \\
\bottomrule
\end{tabular}
  \end{minipage}
	}
\caption{Zero-shot results on the M-ABSA dataset with Qwen-2.5 model.} 
\label{tab:main_result_qwen}
\end{table*}

\end{document}